\documentclass{article}

\usepackage{aaai2027_conference}
\usepackage[utf8]{inputenc} %
\usepackage[T1]{fontenc}    %
\usepackage{hyperref}       %
\usepackage{xcolor}
\hypersetup{
    colorlinks=true,      %
    linkcolor=blue,      %
    urlcolor=blue,       %
    citecolor=blue,      %
    linkbordercolor=blue, %
    urlbordercolor=blue,
    citebordercolor=blue,
    pdfborderstyle={/S/U/W 1}, %
}
\usepackage{float}
\usepackage{url}            %
\usepackage{booktabs}       %
\usepackage{amsfonts}       %
\usepackage{nicefrac}       %
\usepackage{microtype}      %
\usepackage{graphicx}
\usepackage{wrapfig}
\usepackage{natbib}
\usepackage{amsmath}
\usepackage{amssymb} %
\usepackage{amsthm}
\usepackage{xspace}
\usepackage{enumitem}
\usepackage{multirow}
\usepackage{subcaption} 
\usepackage{caption}
\DeclareCaptionLabelSeparator{pipe}{ | }
\usepackage{tcolorbox}
\newtcolorbox{coloredquote}[1][]{
    colback=green!5!white,  
    colframe=green!70!black, 
    boxrule=2pt,
    arc=7pt,
    left=6pt,
    right=6pt,
    top=4pt,
    bottom=4pt,
    title=#1
}

\setlist[itemize]{leftmargin=*}
\setlist[enumerate]{leftmargin=*}
\setlist[description]{leftmargin=*}

\newcommand{\nl}{\ensuremath{\hookleftarrow}}

\title{SAF-OPD: Stable Advantage Fusion for On-Policy Distillation}

\author{
Yifan Ding\textsuperscript{1,2}\thanks{Equal contribution.}, Xincheng Wei\textsuperscript{2,3}$^{*}$, Yoshua Y. Li\textsuperscript{2}$^{*}$\thanks{Corresponding author.} \\
\textbf{Ziheng Li\textsuperscript{2}, Yuquan Lu\textsuperscript{2}, Siyu Zhang\textsuperscript{2}} \\
\textbf{Dongsheng Ma\textsuperscript{2,4}, Rongxiang Weng\textsuperscript{2}, Xunliang Cai\textsuperscript{2}, Yun Chen\textsuperscript{1}\footnotemark[2]} \\
\textsuperscript{1}Shanghai University of Finance and Economics \quad
\textsuperscript{2}Meituan, LongCat Team \\
\textsuperscript{3}The Chinese University of Hong Kong, Shenzhen \quad
\textsuperscript{4}Peking University \\
	\texttt{dingyii2002@gmail.com, yoshua\_li@meituan.com, yunchen@sufe.edu.cn} \\
}

\renewcommand{\headeright}{}
\renewcommand{\shorttitle}{SAF-OPD}

\begin{document}
\maketitle
\setcounter{footnote}{0}

\begin{abstract}
Reinforcement learning with verifiable rewards (RLVR) broadcasts a single response-level reward to every token, while on-policy distillation (OPD) scores each token against a stronger teacher for a dense advantage but caps performance at teacher quality and discourages exploration beyond it. Their complementarity makes combining RLVR and OPD promising, but we find that fusing the two advantages with a fixed coefficient triggers entropy collapse from two miscalibrations: a \textbf{magnitude mismatch}, where token-level OPD advantages can spike far beyond the bounded RLVR advantage and erase its signal, and a \textbf{temporal mismatch}, where sustained full-strength OPD keeps pulling the student toward the teacher and limits exploration needed to surpass it. We propose SAF, a Stable Advantage Fusion framework that resolves both issues via a lightweight, four-stage pipeline applied only to the OPD advantage: a sparsify-then-compress mechanism for magnitude control paired with a warm-up-then-anneal mechanism for temporal control, with each stage independently switchable and adding negligible overhead. Instantiating RLVR with GRPO, we evaluate SAF across seven mathematical reasoning and code generation benchmarks with Qwen3-1.7B/4B/8B: SAF avoids entropy collapse and consistently outperforms fixed-coefficient GRPO+OPD fusion, improving the aggregate score by 0.51--2.70\% across all six model--domain settings while achieving more stable training.
\end{abstract}

\section{Introduction}
\label{sec:introduction}

\begin{wrapfigure}{r}{0.5\linewidth}
\centering
\IfFileExists{Figures/intro.pdf}{%
\includegraphics[width=\linewidth]{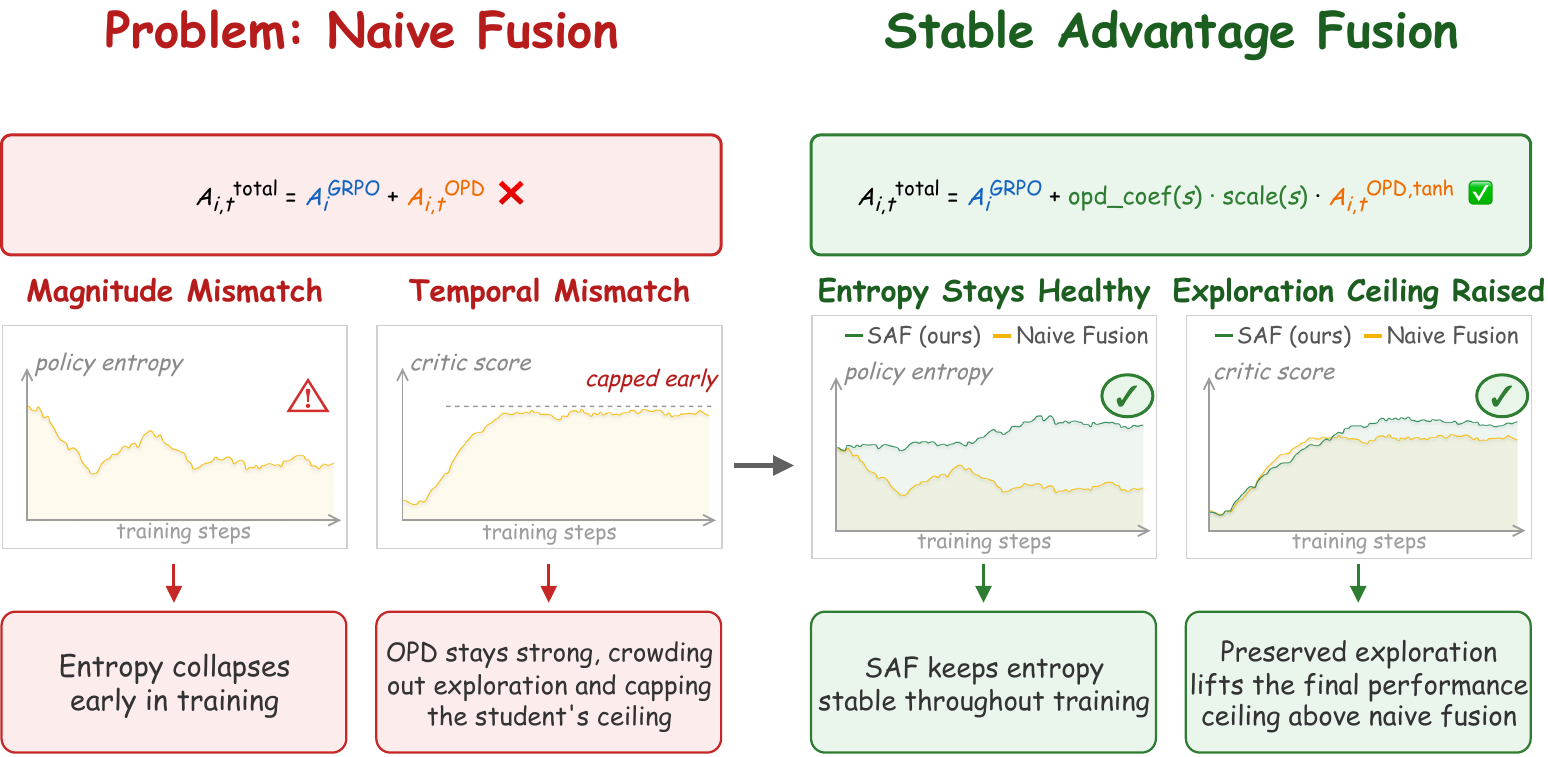}%
}{%
\fbox{\parbox[c][1.25in][c]{\linewidth}{\centering Missing figure file:\\
\texttt{Figures/intro.pdf}}}%
}
\captionsetup{font=small, skip=6pt}
\caption{Fixed-coefficient fusion versus SAF. SAF pairs the OPD advantage's magnitude and temporal mismatches with dedicated control mechanisms, avoiding entropy collapse, preserving exploration, and improving final performance.}
\label{fig:intro}
\end{wrapfigure}
With the development of reinforcement learning with verifiable rewards (RLVR), large language models (LLMs) have demonstrated remarkable capabilities in complex reasoning tasks such as mathematics and code generation~\cite{zhang2025surveyreinforcementlearninglarge}. Among RLVR algorithms, group-relative policy optimization (GRPO)~\cite{shao2024deepseekmathpushinglimitsmathematical, Guo_2025} is a widely adopted and representative instantiation: it scores each rollout with a rule-based verifier and normalizes the score within its sampling group into a single response-level advantage $A_i^{\mathrm{GRPO}}$, broadcast identically to every token in that rollout; we adopt GRPO as our RLVR instantiation. This broadcast is what makes GRPO simple but also coarse: $A_i^{\mathrm{GRPO}}$ treats every token as equally responsible for the outcome, so a long correct derivation and its single decisive step receive the same credit. On-policy distillation (OPD) instead scores each student-generated token under a stronger teacher, providing a vector of token-level advantages $A_{i,t}^{\mathrm{OPD}}= \log \pi_T(y_{i,t}\mid s_{i,t})-\log \pi_{\theta}(y_{i,t}\mid s_{i,t})$, recomputed at every update, giving dense, teacher-relative feedback on problems the student could not yet solve independently~\cite{agarwal2024onpolicydistillationlanguagemodels,gu2024minillm,liu2026selfdistilledpolicygradient}. Yet OPD alone has a complementary weakness: because $A_{i,t}^{\mathrm{OPD}}$ is defined purely relative to the teacher's own token-level likelihood, it rewards matching the teacher's distribution regardless of whether the teacher itself is correct, so its performance ceiling is implicitly capped by teacher quality unless the reward is explicitly reshaped to extrapolate past it~\cite{yang2026learning}; without a ground-truth response-level verifier reward, sustained teacher-matching also offers no pressure to explore beyond the teacher, an outcome an independently verified RLVR reward can directly supply~\cite{cui2025entropymechanismreinforcementlearning}. Since RLVR supplies a sparse but verified signal and OPD a dense but teacher-relative one, the two appear complementary by construction, motivating a growing line of work that fuses them by simply adding the two advantages with a fixed mixing coefficient~\cite{wang2026distilledreinforcementlearningllm,coreteam2026mimov2flashtechnicalreport}.

However, treating this fusion as a simple additive combination glosses over a basic asymmetry between the two terms. $A_i^{\mathrm{GRPO}}$ is a single scalar per rollout, produced once by normalizing a bounded response-level verifier reward against its own sampling group, so its magnitude is self-calibrating~\cite{shao2024deepseekmathpushinglimitsmathematical}. $A_{i,t}^{\mathrm{OPD}}$, by contrast, is a token-level quantity recomputed from a log-probability gap at every update. Its magnitude is not bounded by group-relative normalization and can become large for individual tokens, a known source of training pathologies such as reward spikes and length inflation in OPD-style systems~\cite{wang2026demystifyingonpolicydistillationroles,xing2026trustregiononpolicydistillation,zhao2026poweropdstabilizingonpolicydistillation,luo2026demystifyingopdlengthinflation}. While this gap itself faithfully tracks the local student-teacher discrepancy~\cite{liu2026selfdistilledpolicygradient}, a fixed fusion coefficient does not adapt the contribution of this signal to the student's evolving distance from the teacher: it implicitly assumes the two advantages remain commensurate at every token and step, even as one is normalized once per rollout and the other is unnormalized and re-estimated online. This paper asks whether that assumption holds, and if not, how to fuse the two advantages once it is dropped.

As a preliminary study, we instantiate this recipe with the simplest possible choice, a fixed fusion of $A_i^{\mathrm{GRPO}}$ and $A_{i,t}^{\mathrm{OPD}}$, and find that the asymmetry above is not merely a theoretical concern: it destabilizes training in practice. Policy entropy collapses early in training and stays low thereafter, a failure mode widely reported to stall exploration and cap final accuracy in RLVR~\cite{cui2025entropymechanismreinforcementlearning}; the student is also pulled toward the teacher more aggressively than intended, and later-stage accuracy plateaus below what a better-controlled fusion can reach (Section~\ref{sec:motivation-and-dynamics}). Diagnosing this failure, we trace it to two distinct, coexisting forms of miscalibration that a single fixed coefficient cannot simultaneously address. The first is a \textbf{magnitude mismatch}: unlike the bounded, group-normalized $A_i^{\mathrm{GRPO}}$, the token-level $A_{i,t}^{\mathrm{OPD}}$ is unbounded and heavy-tailed, consistent with reports that unconstrained teacher--student log-ratio signals can spike to extreme magnitudes on individual tokens~\cite{zhao2026poweropdstabilizingonpolicydistillation}, so a small fraction of tokens can carry advantages far larger than any RLVR term and numerically dominate the fused update. The second is a \textbf{temporal mismatch}: the value of full-strength teacher guidance is not constant over training, since once the student has internalized much of what the teacher offers, continuing to apply full-strength OPD guidance over-constrains the policy and caps its performance ceiling. Because both mismatches act on different axes, one on the token-level scale and one on the training-time schedule, neither can be resolved by simply retuning that coefficient; Section~\ref{sec:motivation-and-dynamics} reports the full empirical evidence underlying this diagnosis.

Motivated by this diagnosis, we propose SAF (Stable Advantage Fusion), which pairs the \textbf{magnitude mismatch} and the \textbf{temporal mismatch} with one dedicated mechanism each, rather than tuning a single global coefficient. For the magnitude mismatch, a token-level magnitude controller sparsifies and bounds the empirical OPD advantage distribution within each response, so heavy-tailed tokens can no longer dominate the update. For the temporal mismatch, a training-stage controller tracks the online student--teacher KL divergence to detect when the OPD signal has served its purpose, then anneals its global coefficient accordingly, so full-strength guidance is applied only while it remains useful. Relative to the RLVR+OPD fixed-coefficient baseline, SAF introduces no additional model or auxiliary loss and can be applied as a lightweight transformation of the sampled OPD advantage; Section~\ref{sec:method} details the four constituent stages and their composition into the complete fusion rule.

We summarize our contributions as follows.
\begin{itemize}
    \item We identify a basic asymmetry between a bounded, once-normalized RLVR advantage and an unbounded, online-recomputed OPD advantage, and show that fusing them with a fixed coefficient manifests this asymmetry as two related but distinct forms of miscalibration, one on token-level magnitude and one on training-time scheduling, neither resolved by coefficient tuning alone.
    \item We propose SAF, a lightweight, four-stage advantage fusion pipeline that pairs a sparsify-then-compress mechanism for magnitude control with a warm-up-then-anneal mechanism for temporal control. Every stage is independently switchable, and the annealing horizon is determined relative to the observed warm-up endpoint rather than fixed in advance.
    \item Using GRPO as our RLVR instantiation, extensive experiments across seven mathematical-reasoning and code-generation benchmarks with Qwen3-8B, Qwen3-4B, and Qwen3-1.7B show that SAF: 1) improves the aggregate score over fixed-coefficient GRPO+OPD in every one of the six model--domain settings, with gains of 0.51--2.70\%; and 2) as shown by our training-dynamics analysis, effectively avoids the entropy collapse induced by fixed-coefficient fusion, preserving the policy's exploration capacity throughout training and thereby raising the ceiling on final task performance.
\end{itemize}


\section{Related Work}

\subsection{RLVR and coarse credit assignment}

Reinforcement learning with verifiable rewards (RLVR) trains LLM reasoning with a critic-free, group-normalized response-level advantage broadcast to every token, most commonly instantiated via GRPO~\cite{shao2024deepseekmathpushinglimitsmathematical,Guo_2025,schulman2017proximalpolicyoptimizationalgorithms}; follow-ups such as Dr.GRPO, DAPO, and GSPO refine this scalar's normalization, clipping, or importance weighting but leave its coarse granularity unchanged~\cite{liu2025understandingr1zeroliketrainingcritical,yu2025dapoopensourcellmreinforcement,zheng2025groupsequencepolicyoptimization}. This granularity is most limiting exactly where rollouts are least informative: on hard problems where the sampling group rarely contains a correct response, every token receives the same near-zero advantage, so GRPO provides little signal precisely where the student most needs to improve. Process reward models and entropy- or uncertainty-based reweighting target this granularity directly but rely on signals internal to the student, so they cannot guide problems it cannot yet solve~\cite{lightman2023letsverifystepstep,wang2024mathshepherdverifyreinforcellms,cui2025entropymechanismreinforcementlearning}.

\subsection{On-policy distillation and its combination with RLVR}

Unlike offline distillation on fixed teacher targets~\cite{hinton2015distillingknowledgeneuralnetwork,kim2016sequencelevelknowledgedistillation}, on-policy distillation (OPD) scores student-generated trajectories with teacher token probabilities, reducing train--generation mismatch and enabling dense, token-level feedback~\cite{agarwal2024onpolicydistillationlanguagemodels,gu2024minillm}; it is now a common post-training stage~\cite{song2026surveyonpolicydistillationlarge,zhang2026formuladrivensurveyresearchagenda}, and we adopt its sampled-token variant for a cheaper single-token estimate~\cite{ko2024distillmstreamlineddistillationlarge}. One line regulates this signal in isolation under teacher--student mismatch—restricting distillation to reliable states, adapting the divergence to local uncertainty, relaxing strict imitation, shifting supervision to intermediate representations, or reducing estimator variance, while G-OPD instead reinterprets OPD as KL-constrained RL~\cite{zhong2026sod,jin2026entropyawareonpolicydistillationlanguage,ko2026scalingreasoningefficientlyrelaxed,xing2026trustregiononpolicydistillation,wang2026demystifyingonpolicydistillationroles,yang2026learning,yang2026oprdonpolicyrepresentationdistillation,oh2026klklonpolicydistillation}—but none combines OPD with an RLVR signal. A second line fuses OPD with RLVR but leaves the resulting miscalibration unaddressed: KDRL sums the two losses~\cite{xu2025kdrlposttrainingreasoningllms}, and MiMo-V2-Flash adds a token-level teacher--student advantage to the GRPO advantage at a fixed strength~\cite{coreteam2026mimov2flashtechnicalreport}—the fixed-coefficient fusion whose entropy collapse motivates this paper (Section~\ref{sec:introduction}). RLSD, SDAR, and others move beyond a fixed coefficient but only gate or schedule the whole OPD term via a coarse, response-level statistic~\cite{yang2026selfdistilledrlvr, liu2026selfdistilledpolicygradient, lu2026sdar, tan2026atodannealedturnawareonpolicy, pan2026rlcsdreinforcementlearningcontrastive, wang2026tracedistillingmatterstokenrouted}; none bounds an individual token's OPD advantage relative to $A_i^{\mathrm{GRPO}}$, so heavy-tailed magnitudes can still dominate after coarse gating. SAF instead pairs each mismatch with a matched-granularity mechanism, bounding each token's influence and adapting OPD strength from the online student--teacher KL divergence rather than a response-level statistic.

\section{Method}
\label{sec:method}

\begin{figure*}[!t]
    \centering
\IfFileExists{Figures/method.pdf}{%
\includegraphics[width=\textwidth]{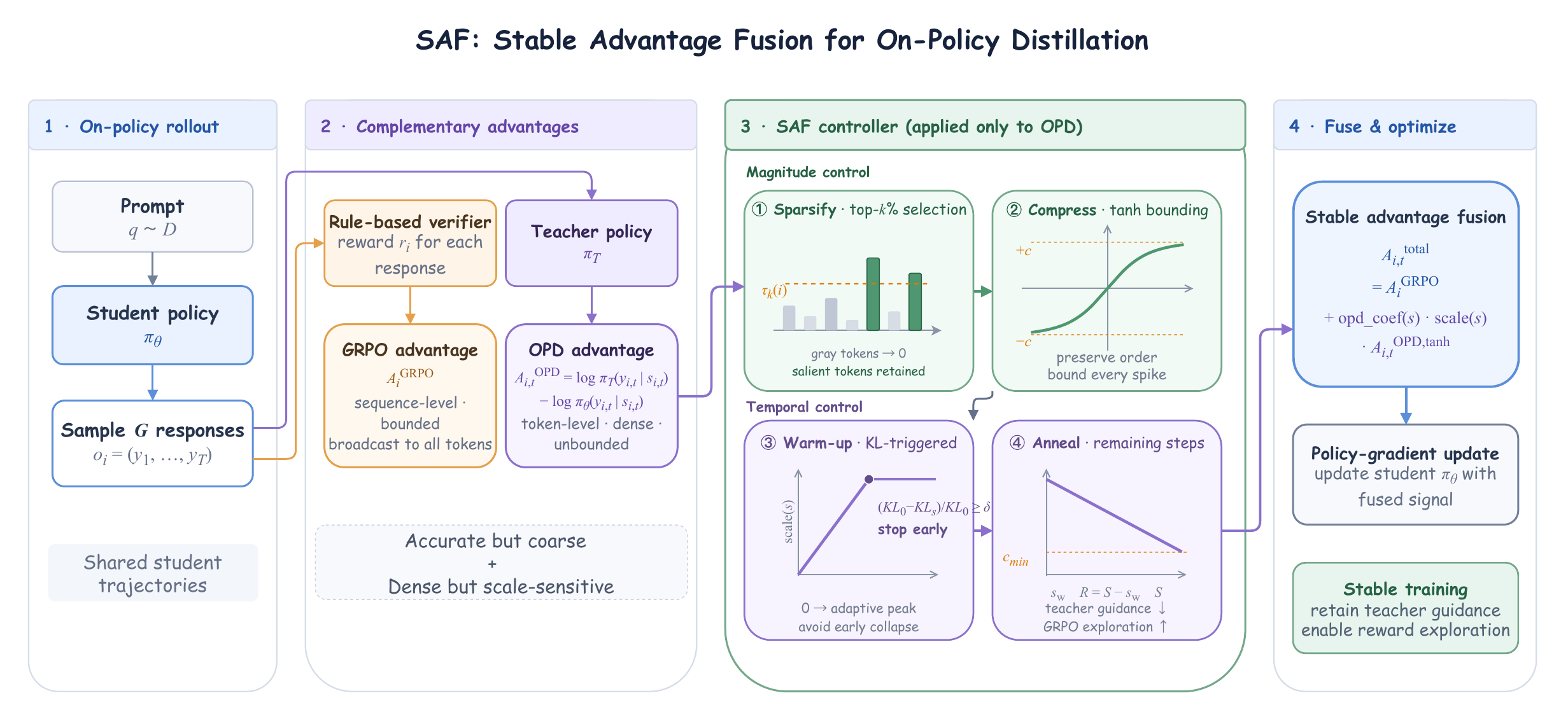}%
}{%
\fbox{\parbox[c][1.25in][c]{\textwidth}{\centering Missing figure file:\\
\texttt{Figures/method.pdf}}}%
}
    \caption{Overview of SAF. SAF modifies only the OPD branch through four stages before it is added to the unchanged GRPO advantage for the policy-gradient update.}
    \label{fig:method}
\end{figure*}

\subsection{Preliminary: Advantage Fusion Problem}
\label{sec:preliminary}

Given a prompt $q$, GRPO samples a group of $G$ responses $\{o_1, \dots, o_G\}$ from the rollout policy and assigns each response a single scalar reward from a rule-based verifier, normalized against the group's own statistics into a response-level advantage $A_i^{\mathrm{GRPO}}$ that is broadcast identically to every token in response $o_i$. $A_i^{\mathrm{GRPO}}$ is therefore bounded and verified, but carries no information about which token within $o_i$ was responsible for the outcome. In parallel, OPD minimizes the reverse KL divergence from the student policy $\pi_{\theta}$ to the teacher policy $\pi_T$ on student-generated trajectories, where $s_{i,t}=(\boldsymbol{x}_i,\boldsymbol{y}_{i,<t})$ denotes the token context. Under the next-token approximation used in recent OPD systems~\cite{coreteam2026mimov2flashtechnicalreport,yang2026learning}, comparison with the policy-gradient form identifies the negative token-level log-probability gap as the OPD advantage:
\begin{equation}
\label{eq:opd-advantage}
A_{i,t}^{\mathrm{OPD}}
= \log \pi_T(y_{i,t}\mid s_{i,t})
   -\log \pi_{\theta}(y_{i,t}\mid s_{i,t}),
\end{equation}
so a token receives a positive advantage when the teacher assigns it a higher probability than the student and a negative advantage otherwise. Unlike $A_i^{\mathrm{GRPO}}$, $A_{i,t}^{\mathrm{OPD}}$ is recomputed independently at every token $t$, so its magnitude is not bounded by any group-relative normalization.

Notably, G-OPD~\cite{yang2026learning} shows that OPD is theoretically a special case of dense KL-constrained RL, with $A_{i,t}^{\mathrm{OPD}}$ itself an implicit reward under this view, grounding $A_i^{\mathrm{GRPO}}$ and $A_{i,t}^{\mathrm{OPD}}$ as two reward signals within the same RL formulation and justifying their summation into a single fused advantage.

Our goal is to construct a fused signal $A_{i,t}^{\mathrm{total}}$ for the policy-gradient update that (i) bounds the absolute influence of any single token's $A_{i,t}^{\mathrm{OPD}}$ and prevents it from numerically overwhelming the sequence-level $A_i^{\mathrm{GRPO}}$, and (ii) adapts the overall strength of the OPD contribution over training. SAF addresses these two requirements with four independently switchable stages applied only to $A_{i,t}^{\mathrm{OPD}}$, leaving $A_i^{\mathrm{GRPO}}$ untouched: Section~\ref{sec:magnitude-control} presents a magnitude controller (top-$k\%$ sparsification followed by bounded $\tanh$ compression), Section~\ref{sec:temporal-control} presents a temporal controller (KL-triggered warm-up followed by linear annealing), and Section~\ref{sec:overall-formula} composes the four stages into the complete fusion rule.

\subsection{Magnitude Control: Sparsify-then-Compress}
\label{sec:magnitude-control}

A remaining problem after fixing $A_{i,t}^{\mathrm{OPD}}$ as in Section~\ref{sec:preliminary} is that its token-level magnitude is unconstrained: because it is recomputed independently at every token from a log-probability difference, a small number of tokens can produce values an order of magnitude larger than $A_i^{\mathrm{GRPO}}$, letting a handful of tokens dominate the gradient for the entire sequence once incorporated into $A_{i,t}^{\mathrm{total}}$. We address this with a sparsification stage that removes low-salience tokens and a compression stage that bounds the magnitude of the tokens that remain.

\paragraph{Stage 1: Top-$k\%$ sparsification.} As shown in Figure~\ref{fig:opd_bin_dist} (Section~\ref{sec:motivation-and-dynamics}), the empirical distribution of $|A_{i,t}^{\mathrm{OPD}}|$ is highly concentrated near zero, with substantially larger magnitudes occurring at only a small number of tokens. For each sequence $i$, we therefore retain only the tokens whose magnitude exceeds the $(1-k\%)$ quantile within that sequence, zeroing out the rest:
$$
M_{i,t} = \mathbf{1}\!\left[|A_{i,t}^{\mathrm{OPD}}| \ge \tau_k(i)\right],
$$
where $\tau_k(i)$ is the $(1-k\%)$ quantile of $\{|A_{i,t}^{\mathrm{OPD}}|\}$ over sequence $i$ and $k$ is a hyperparameter (\texttt{topk\_percent}). This yields the filtered signal $A_{i,t}^{\mathrm{OPD},\,\text{top-}k\%} = M_{i,t} \cdot A_{i,t}^{\mathrm{OPD}}$. Because $\tau_k(i)$ is recomputed per sequence, the filter adapts to each sequence's own OPD magnitude distribution rather than applying a single global cutoff; setting $k=100$ disables filtering entirely, so this stage strictly generalizes the unfiltered signal.

\paragraph{Stage 2: Bounded $\tanh$ compression.} Even after Stage 1 removes the least salient tokens, the magnitudes of the surviving tokens can still differ substantially, with occasional extreme outliers. We optionally pass the filtered signal through a bounded, tunable-scale transform:
\begin{equation}
A_{i,t}^{\mathrm{OPD,tanh}}
= c\,\tanh\!\left(A_{i,t}^{\mathrm{OPD},\,\text{top-}k\%}\right),
\end{equation}
where $c$ is a tunable coefficient. Since $\tanh$ maps $\mathbb{R}$ into $(-1, 1)$, the compressed signal is confined to $(-c, c)$ regardless of the pre-compression magnitude, while remaining close to linear near the origin so relative ordering of small-to-moderate signals is preserved. This stage is elementwise, adds negligible overhead, and can be switched off independently. Together, the two stages bound the \textbf{magnitude mismatch}: Stage 1 decides \emph{which} tokens contribute based on salience relative to their own sequence, and Stage 2 decides \emph{how large} a contribution any surviving token can make.

\subsection{Temporal Control: Warm-up-then-Anneal}
\label{sec:temporal-control}

A separate problem is that the appropriate strength of $A_{i,t}^{\mathrm{OPD}}$ changes over training. Injecting it at full strength from the first update can prematurely collapse policy entropy~\cite{cui2025entropymechanismreinforcementlearning}, whereas maintaining that strength after the student has absorbed most of the teacher's guidance can pin the policy to teacher behavior, crowd out GRPO exploration, and even destabilize generation altogether~\cite{luo2026demystifyingopdlengthinflation}. We therefore first ramp the OPD strength linearly from zero, allowing the observed student--teacher KL drop to terminate this warm-up early, and then reduce the OPD coefficient linearly over the remaining steps so GRPO progressively regains control.

\paragraph{Stage 3: KL-triggered warm-up.} We scale the output of Stage 2 by a factor that increases linearly with the global step $s$:
$$
\text{scale}(s) = \min\!\left(\frac{s}{S_{\text{warmup}}}, 1.0\right),
$$
where $S_{\text{warmup}}$ is the maximum warm-up duration. Rather than relying solely on a pre-specified step budget, we monitor a sampled-token estimate of the reverse KL divergence from student to teacher, computed on the same student-generated responses used for policy optimization (Appendix~\ref{app:student-teacher-kl}). Letting $KL_0$ denote this estimate at the onset of warm-up and $KL_s$ its value at step $s$, we terminate warm-up early once
$$
\frac{KL_0-KL_s}{KL_0} \ge \delta, \qquad \delta>0,
$$
freezing $\text{scale}(s)$ at its current value. This early-stop criterion uses the observed reduction in sampled student-to-teacher reverse KL, rather than a fixed step count, to determine when the ramp should stop, since the required reduction varies substantially across models and tasks. It also guards against over-imitation: the teacher's own competence is not unbounded and its token-level scores can be miscalibrated~\cite{yang2026learning}, so once the student has closed most of the reachable gap, prolonging full-strength guidance mainly risks fitting the teacher's errors rather than yielding further benefit.

\paragraph{Stage 4: Linear annealing over the remaining steps.} Let $S$ denote the total training budget and let $s_{\mathrm{w}}$ be the step at which warm-up ends, either naturally or via the KL-triggered condition. We start a local counter $t_{\mathrm{a}}=0$ at $s_{\mathrm{w}}$, increment it after each subsequent update, and define the OPD coefficient over the remaining budget $R=S-s_{\mathrm{w}}$ as
\begin{equation}
\text{opd\_coef}(s)=
\begin{cases}
1, & s<s_{\mathrm{w}}, \\[3pt]
c_{\min}+(1-c_{\min})
\left(1-\dfrac{t_{\mathrm{a}}}{R}\right),
& s\ge s_{\mathrm{w}},
\end{cases}
\end{equation}
where $0\le t_{\mathrm{a}}\le R$ and $c_{\min}$ is a small floor preserving a residual OPD contribution. Annealing thus begins at coefficient $1$ when warm-up ends and reaches $c_{\min}$ at the final step; an earlier KL-triggered termination enlarges $R$ and makes the decay more gradual. When warm-up is disabled, we set $s_{\mathrm{w}}=0$ and anneal over the full training budget.

\subsection{Overall Fusion and Implementation}
\label{sec:overall-formula}

Composing the four stages in order (Stage 1 $\to$ Stage 2 $\to$ Stage 3 $\to$ Stage 4) gives the final fused advantage
\begin{equation}
\label{eq:complete-fusion}
A_{i,t}^{\mathrm{total}}
= A_i^{\mathrm{GRPO}}
+ \text{opd\_coef}(s)\cdot \text{scale}(s)\,
A_{i,t}^{\mathrm{OPD,tanh}}.
\end{equation}
Each disabled stage reduces to the identity; disabling all four recovers the fixed fusion $A_i^{\mathrm{GRPO}}+A_{i,t}^{\mathrm{OPD}}$. SAF therefore modifies only the advantage-fusion step and introduces no additional model, loss, or forward pass: Stage 1 computes one quantile per sequence, Stage 3 stores $KL_0$ to evaluate $(KL_0-KL_s)/KL_0$, and the temporal controller maintains only $s_{\mathrm{w}}$ and $t_{\mathrm{a}}$, so SAF can be inserted into an existing GRPO+OPD training loop as a drop-in replacement.

\begin{table*}[t]
  \centering
  \resizebox{\textwidth}{!}{%
  \begin{tabular}{lccccccccc}
    \toprule
    & \multicolumn{5}{c}{Mathematical Reasoning} & \multicolumn{4}{c}{Code Generation} \\
    \cmidrule(lr){2-6}\cmidrule(lr){7-10}
    Method & AIME-24 & AIME-25 & HMMT25-Feb & HMMT25-Nov & Avg. & HumanEval+ & MBPP+ & LiveCodeBench & Avg. \\
    \midrule
    \textit{Teacher: Qwen3-30B-A3B-Instruct-2507} & 73.65 & 61.98 & 43.85 & 57.81 & 59.32 & 82.93 & 78.31 & 45.00 & 68.75 \\
    \midrule
    \multicolumn{10}{l}{\textit{Student: Qwen3-8B}} \\
    Base                 & 26.56 & 21.25 & 11.35 & 9.79 & 17.24 & 80.49 & 72.49 & 23.71 & 58.90 \\
    GRPO-only            & \underline{61.15} & 49.06 & \underline{28.65} & 37.50 & 44.09 & \underline{81.10} & \textbf{72.22} & 28.85 & 60.72 \\
    OPD-only             & 59.58 & \underline{50.52} & 27.60 & 40.73 & 44.61 & \textbf{84.15} & \underline{71.96} & 33.29 & \underline{63.13} \\
    GRPO+OPD (fixed)     & 60.83 & \textbf{51.25} & 28.44 & \textbf{43.33} & \underline{45.96} & 78.66 & 71.69 & \underline{34.86} & 61.74 \\
    \textbf{SAF (ours)} & \textbf{64.79} & \textbf{51.25} & \textbf{30.00} & \underline{41.67} & \textbf{46.93} & \underline{81.10} & 71.69 & \textbf{37.43} & \textbf{63.41} \\
    \midrule
    \multicolumn{10}{l}{\textit{Student: Qwen3-4B}} \\
    Base                 & 23.02 & 21.88 & 11.67 & 9.17 & 16.44 & 79.27 & 63.49 & 24.57 & 55.78 \\
    GRPO-only            & \underline{58.96} & 50.62 & \underline{30.10} & 37.60 & 44.32 & \underline{80.49} & 68.78 & 32.14 & \underline{60.47} \\
    OPD-only             & 57.81 & 51.56 & 29.38 & 37.71 & 44.12 & 78.66 & \underline{69.31} & 30.71 & 59.56 \\
    GRPO+OPD (fixed)     & 57.19 & \underline{51.98} & 29.90 & \textbf{38.44} & \underline{44.38} & 79.88 & 66.14 & \underline{33.86} & 59.96 \\
    \textbf{SAF (ours)} & \textbf{60.21} & \textbf{53.96} & \textbf{31.15} & \underline{38.23} & \textbf{45.89} & \textbf{82.93} & \textbf{70.63} & \textbf{34.43} & \textbf{62.66} \\
    \midrule
    \multicolumn{10}{l}{\textit{Student: Qwen3-1.7B}} \\
    Base                 & 12.81 & 10.83 & 5.94 & 3.54 & 8.28 & 60.98 & 54.23 & 15.14 & 43.45 \\
    GRPO-only            & \underline{36.25} & \underline{31.35} & 17.19 & \underline{16.88} & \underline{25.42} & 65.24 & 53.70 & 17.57 & 45.50 \\
    OPD-only             & 35.10 & 28.54 & 15.83 & 16.35 & 23.96 & \textbf{70.73} & \textbf{58.73} & 25.86 & \textbf{51.77} \\
    GRPO+OPD (fixed)     & 34.79 & 29.69 & \underline{17.50} & 16.56 & 24.64 & \underline{70.12} & 56.08 & \textbf{26.43} & 50.88 \\
    \textbf{SAF (ours)} & \textbf{36.67} & \textbf{31.98} & \textbf{18.02} & \textbf{19.27} & \textbf{26.49} & \textbf{70.73} & \underline{57.14} & \underline{26.29} & \underline{51.39} \\
    \bottomrule
  \end{tabular}%
  }
  \caption{Results on mathematical reasoning and code generation. All entries are accuracies (\%). Within each model scale and column, the best result is bold and the second best is underlined.}
  \label{tab:main-results}
\end{table*}

\section{Experiments}

\subsection{Experimental Setup}

\paragraph{Models and training.} We implement SAF in a GRPO+OPD pipeline based on verl~\cite{Sheng_2025}, initializing the student from Qwen3-8B, Qwen3-4B, or Qwen3-1.7B with Qwen3-30B-A3B-Instruct-2507 as the teacher~\cite{yang2025qwen3technicalreport} that supplies the token-level log probabilities for $A_{i,t}^{\mathrm{OPD}}$. All methods at a given scale share the same initialization, teacher, and training data: 57K DeepMath~\cite{deepmath} problems with difficulty $\geq 6$ for mathematics and the 25K-problem Eurus-RL-Code dataset~\cite{prime} for code, with on-policy student responses, verifiable rewards driving the GRPO signal, and teacher scores driving the OPD signal. OPD-only uses a dedicated distillation configuration, while the other methods share a separate configuration (Appendix~\ref{app:training-settings}). We compare SAF against four controls: the untrained \textbf{Base} model; \textbf{GRPO-only} (removes $A_{i,t}^{\mathrm{OPD}}$); \textbf{OPD-only} (removes $A_i^{\mathrm{GRPO}}$); and \textbf{GRPO+OPD (fixed)}, which fixes the OPD coefficient at 1 and disables SAF's controls. Unless stated otherwise, SAF uses $k=20$, $c=0.1$, $S_{\text{warmup}}=100$, $\delta=0.2$, and $c_{\min}=0$.

\paragraph{Evaluation.} We evaluate mathematical reasoning on AIME24~\cite{aime2024}, AIME25~\cite{aime2025}, and HMMT25 (February and November)~\cite{hmmt25}, and code generation on HumanEval+, MBPP+~\cite{evalplus}, and LiveCodeBench (v6, February$\sim$May 2025)~\cite{lcb}. Math answers are validated with \texttt{Math-Verify}\footnote{\url{https://github.com/huggingface/Math-Verify}}, and code is scored with the benchmark-provided unit tests; we report the unweighted mean per domain. Full decoding and sampling settings are in Appendix~\ref{app:training-settings}.

\subsection{Main Results}

Table~\ref{tab:main-results} shows that SAF consistently improves the aggregate score over fixed-coefficient fusion across both domains and all three model scales. On mathematical reasoning, SAF improves over GRPO+OPD (fixed) by 0.97\% for Qwen3-8B, 1.51\% for Qwen3-4B, and 1.85\% for Qwen3-1.7B. On code generation, the corresponding gains are 1.67\%, 2.70\%, and 0.51\%. Averaged over the six model--domain settings, SAF reaches 49.46\%, outperforming GRPO+OPD (fixed), GRPO-only, and OPD-only by 1.54\%, 2.71\%, and 1.60\%, respectively. These consistent aggregate gains indicate that controlling the OPD signal is more effective than either discarding it or injecting it at a full strength.

The improvements are broad but not uniform. Qwen3-8B SAF is best (or tied for best) on four of seven benchmarks, improving the domain average by 2.84\% over GRPO-only on mathematics and 2.69\% on code; the exception is MBPP+, where SAF trails GRPO-only and OPD-only by no more than 0.53\% while still improving over fixed fusion on average. Qwen3-4B SAF is best on six of seven benchmarks, improving over GRPO-only by 1.57\% on mathematics and 2.19\% on code. For Qwen3-1.7B, SAF is best or tied for best on five of seven benchmarks and improves over GRPO-only by 1.07\% on mathematics and 5.89\% on code; on 1.7B code, OPD-only attains the highest average (51.77\%) followed by SAF (51.39\%), which still exceeds fixed fusion by 0.51\%. These exceptions make clear that SAF consistently improves over fixed fusion at the aggregate level without dominating every baseline at every scale.

\subsection{Ablation Study}

We next examine representative Qwen3-4B mathematical-reasoning runs under the same 300-step budget. Table~\ref{tab:ablation} shows that magnitude control alone performs similarly to fixed fusion (44.35\% versus 44.38\%), and adding warm-up without the complete temporal controller is also insufficient (44.07\%). This does not negate the need for top-$k$ filtering and $\tanh$ compression: Figure~\ref{fig:opd_bin_dist} and Appendix~\ref{app:raw-signal} directly expose the near-zero mass and extreme token-level magnitudes these operations regulate. Rather, correcting this magnitude mismatch alone is insufficient for final accuracy. When annealing is enabled, the average rises to 45.23\%. With the full SAF configuration, the selected threshold $\delta=0.2$ reaches 45.89\%, whereas increasing it to $\delta=0.3$ yields 44.51\%. Since a larger $\delta$ requires a bigger KL drop before warm-up ends, raising $\delta$ from 0.2 to 0.3 prolongs full-strength OPD guidance yet lowers accuracy by 1.38\%, consistent with the student approaching the ceiling of what a fallible teacher can offer, so continuing to chase it mainly reinforces the teacher's own errors. This supports selecting $\delta=0.2$. The evidence thus supports complementary roles: top-$k$ filtering and $\tanh$ compression make the OPD signal suitable for fusion, while the temporal controller determines when and how strongly it should influence optimization. The full SAF configuration improves over fixed fusion and the magnitude-only variant by 1.51\% and 1.54\%, respectively.

\begin{table}[htbp]
  \centering
  \resizebox{\columnwidth}{!}{%
  \begin{tabular}{lccccc}
    \toprule
    Configuration & AIME-24 & AIME-25 & HMMT25-Feb & HMMT25-Nov & Avg. \\
    \midrule
    GRPO+OPD (fixed)                 & 57.19 & 51.98 & 29.90 & \textbf{38.44} & 44.38 \\
    + top-$k$ and $\tanh$ (fixed weight) & 58.75 & 50.94 & 30.10 & 37.60 & 44.35 \\
    + warm-up (no annealing)        & 58.96 & 51.46 & 29.69 & 36.15 & 44.07 \\
    + annealing                     & 59.27 & 52.60 & \textbf{31.25} & 37.81 & 45.23 \\
    \textbf{SAF ($\delta=0.2$, selected)} & \textbf{60.21} & \textbf{53.96} & 31.15 & 38.23 & \textbf{45.89} \\
    SAF ($\delta=0.3$)              & 58.75 & 52.81 & 29.38 & 37.08 & 44.51 \\
    \bottomrule
  \end{tabular}%
  }
  \caption{Representative SAF ablations on Qwen3-4B mathematical reasoning after 300 training steps.}
  \label{tab:ablation}
\end{table}

\subsection{Analysis of Training Dynamics}
\label{sec:motivation-and-dynamics}

This subsection reports the preliminary study that motivates SAF's design (Section~\ref{sec:introduction}): we instantiate the fixed fusion $A_i^{\mathrm{GRPO}}+A_{i,t}^{\mathrm{OPD}}$ on Qwen3-4B mathematical reasoning, diagnose why it destabilizes training, and trace the resulting \textbf{magnitude mismatch} and \textbf{temporal mismatch} through the raw OPD advantage distribution and the optimization trajectories it produces, then analyze \emph{why} SAF improves over this fusion once both mismatches are controlled.

\begin{wrapfigure}{r}{0.5\linewidth}
\centering
\IfFileExists{Figures/raw_opd_dist_10steps.pdf}{%
  \includegraphics[width=\linewidth]{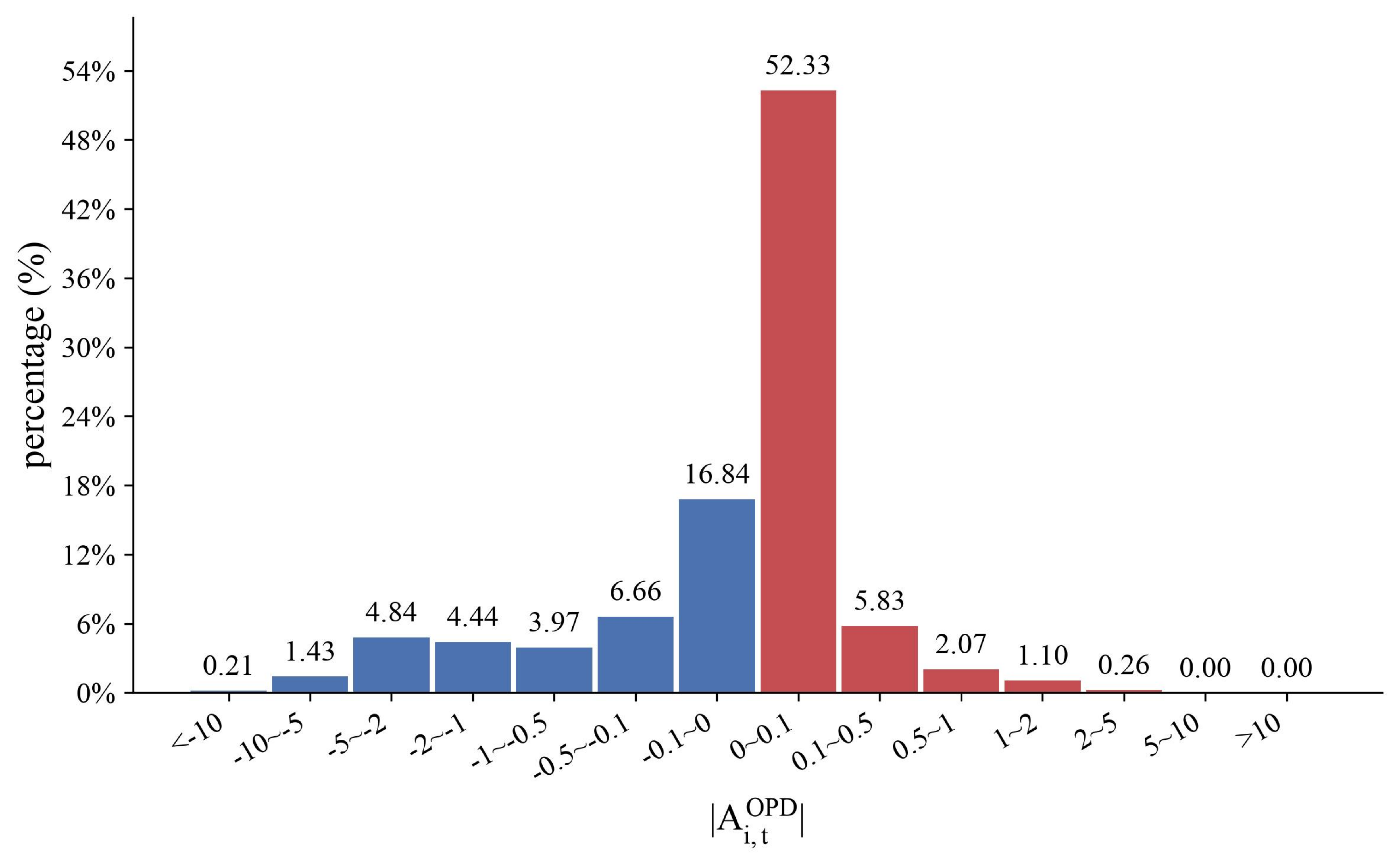}%
}{%
  \fbox{\parbox[c][1.25in][c]{\linewidth}{\centering Missing figure file:\\
  \texttt{Figures/raw\_opd\_dist\_10steps.pdf}}}%
}
\captionsetup{font=small, skip=6pt}
\caption{Empirical distribution of the absolute token-level OPD advantages, $|A_{i,t}^{\mathrm{OPD}}|$, collected during the first 10 training steps of Qwen3-4B on mathematical reasoning tasks.}
\label{fig:opd_bin_dist}
\end{wrapfigure}

\paragraph{OPD advantage distribution.} We first quantify the magnitude mismatch between $A_{i,t}^{\mathrm{OPD}}$ and $A_i^{\mathrm{GRPO}}$ that Stages~1--2 are designed to regulate. Figure~\ref{fig:opd_bin_dist} bins $|A_{i,t}^{\mathrm{OPD}}|$ collected during the first ten training steps of Qwen3-4B, and Appendix~\ref{app:raw-signal} reports representative token-level values over five additional steps. Both views show the same pattern: the overwhelming majority of tokens carry near-zero advantage, while a small subset is one to two orders of magnitude larger. Concretely, every one of the 180 largest-magnitude OPD tokens inspected in Appendix~\ref{app:raw-signal} exceeds the GRPO magnitude of its own sequence, with OPD extremes reaching $20.3585$ against a largest co-occurring GRPO magnitude of only $2.4749$. Under fixed fusion, such tokens numerically dominate the update for their entire sequence despite the two advantages being nominally weighted 1:1. This distribution is the basis for the per-sequence top-$k\%$ filter and bounded $\tanh$ compression introduced in Section~\ref{sec:magnitude-control}.

\begin{figure*}[!t]
    \centering
    \begin{minipage}[t]{0.32\textwidth}
        \centering
        \includegraphics[width=\linewidth]{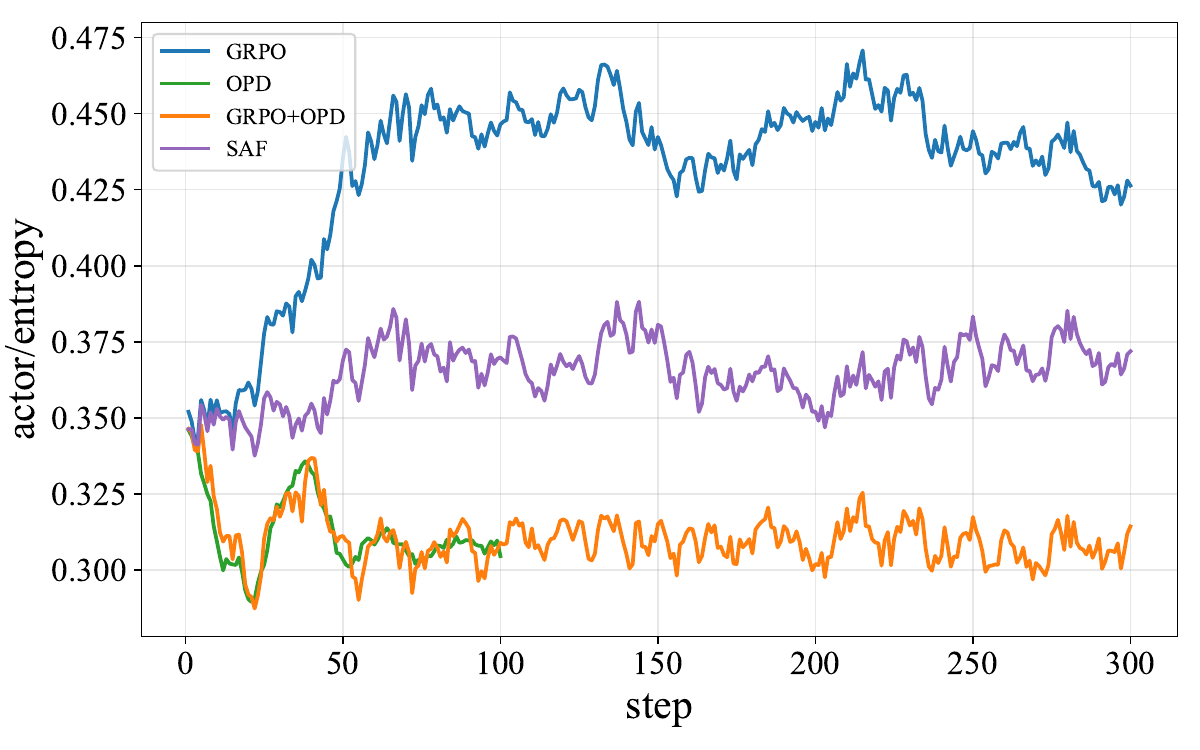}\\
        \small (a) Actor entropy
    \end{minipage}\hfill
    \begin{minipage}[t]{0.32\textwidth}
        \centering
        \includegraphics[width=\linewidth]{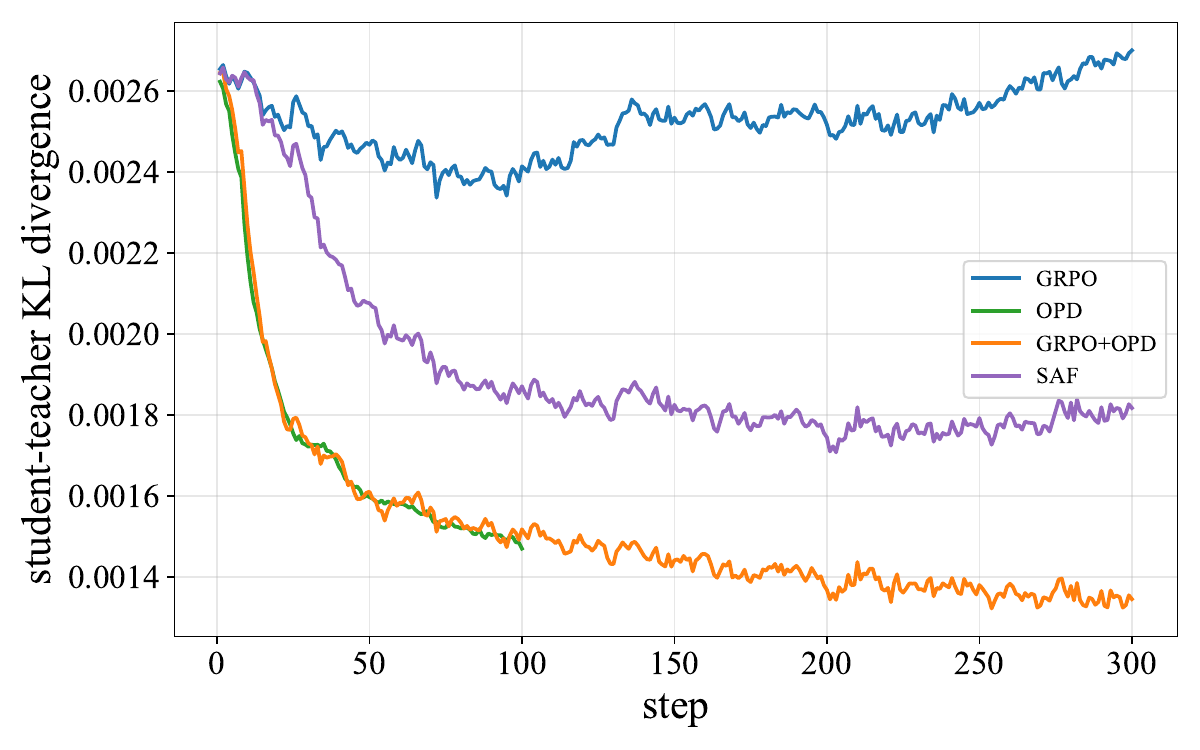}\\
        \small (b) Student--teacher KL divergence
    \end{minipage}\hfill
    \begin{minipage}[t]{0.32\textwidth}
        \centering
        \includegraphics[width=\linewidth]{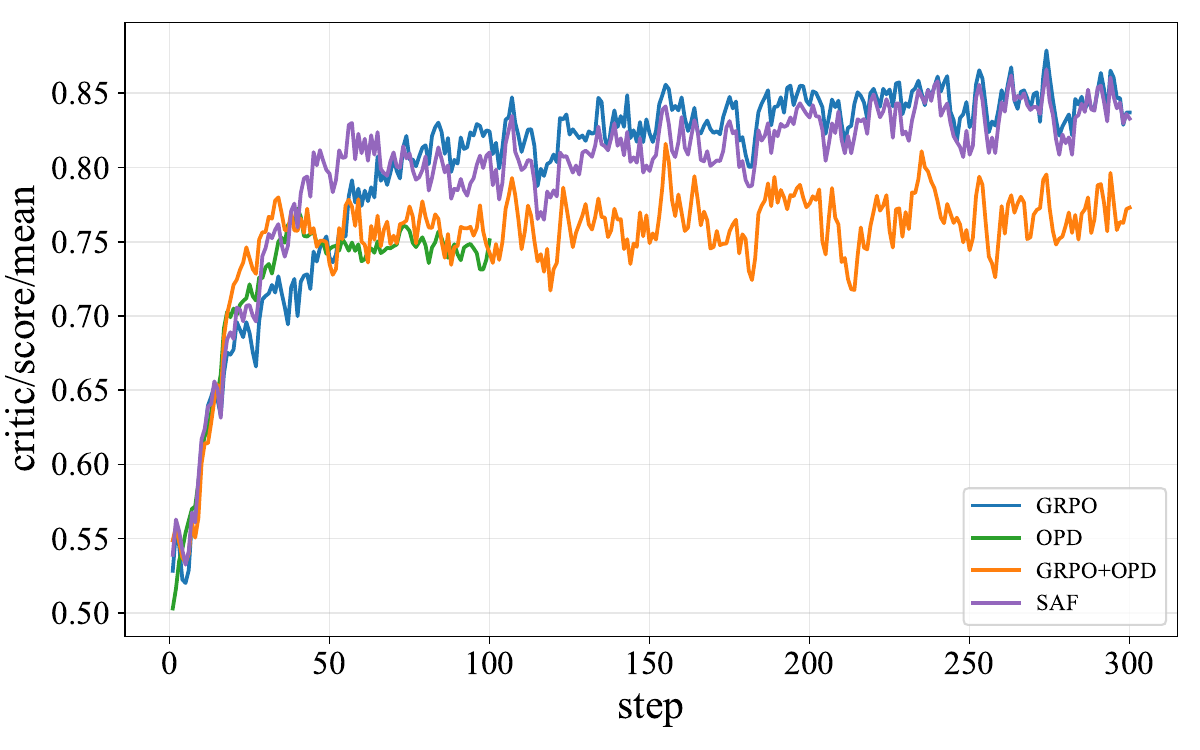}\\
        \small (c) Critic score
    \end{minipage}

    \medskip

    \begin{minipage}[t]{0.32\textwidth}
        \centering
        \includegraphics[width=\linewidth]{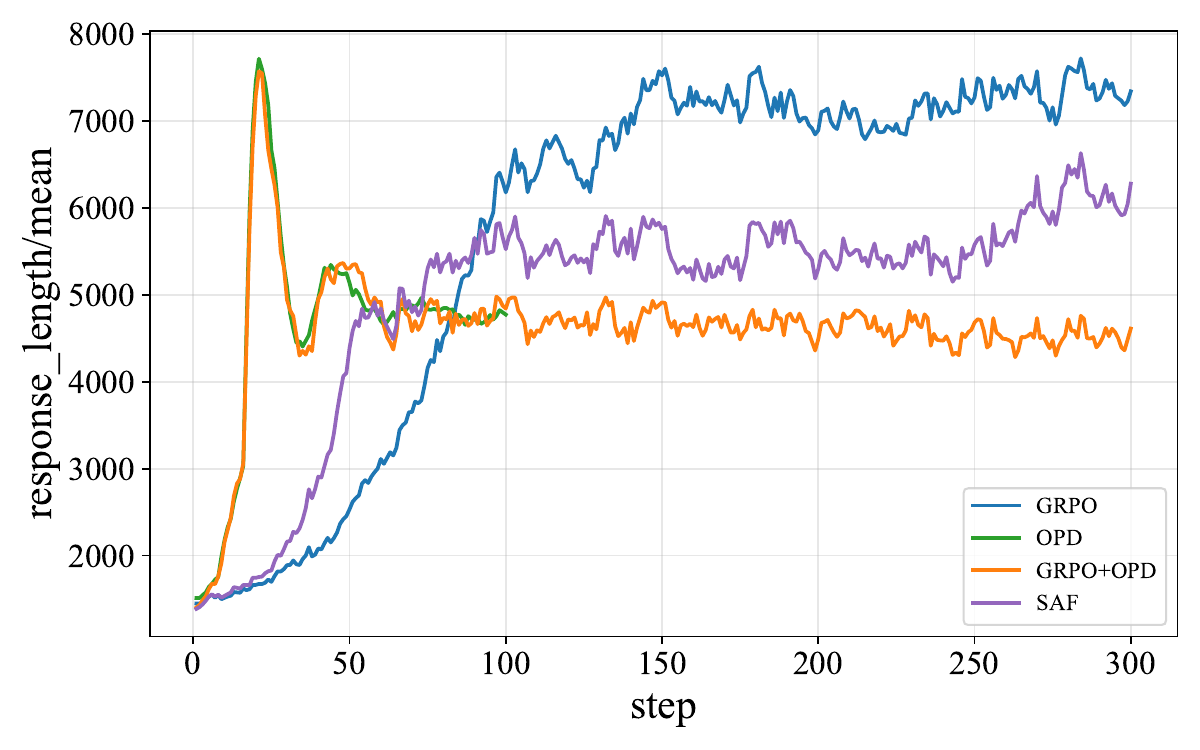}\\
        \small (d) Response length
    \end{minipage}\hfill
    \begin{minipage}[t]{0.32\textwidth}
        \centering
        \includegraphics[width=\linewidth]{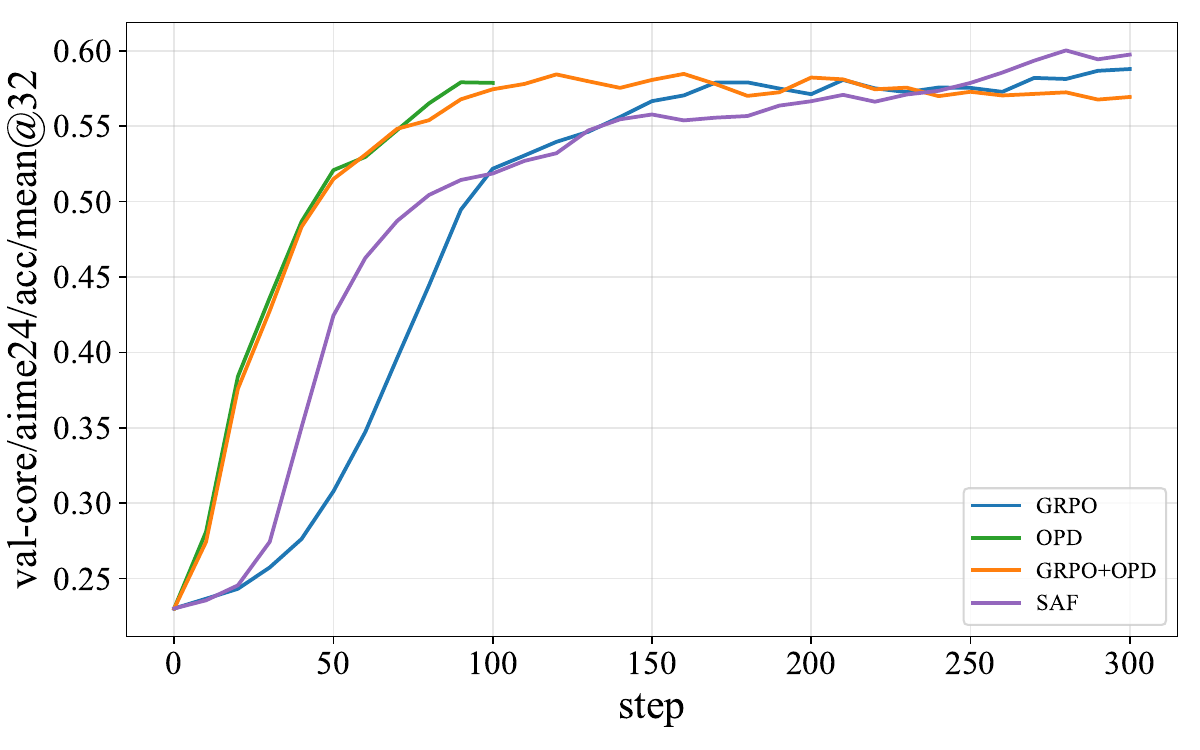}\\
        \small (e) AIME-24 accuracy
    \end{minipage}\hfill
    \begin{minipage}[t]{0.32\textwidth}
        \centering
        \includegraphics[width=\linewidth]{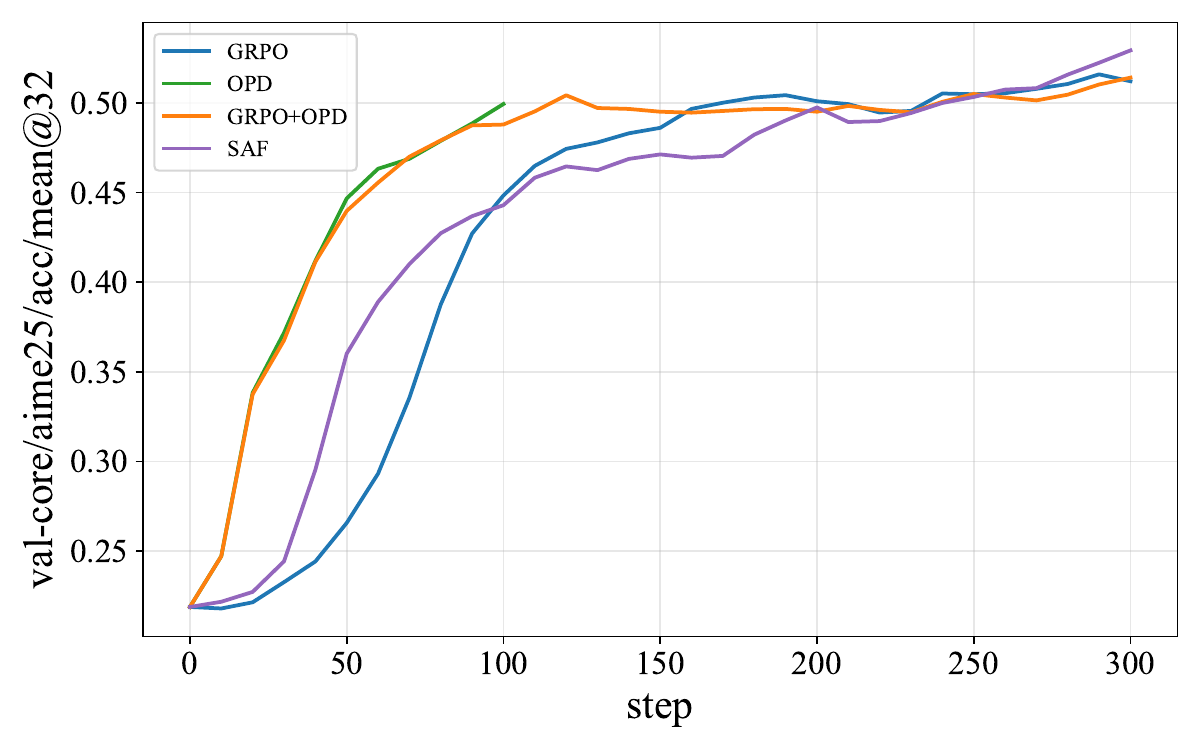}\\
        \small (f) AIME-25 accuracy
    \end{minipage}
    \caption{Training dynamics of Qwen3-4B on mathematical reasoning tasks over 300 optimization steps: (a) actor entropy, (b) student--teacher KL divergence, (c) mean critic score, (d) mean response length, (e) AIME-24 accuracy, and (f) AIME-25 accuracy. Curves compare GRPO, OPD, fixed-coefficient GRPO+OPD, and SAF under the same task setting.}
    \label{fig:training-dynamics}
\end{figure*}

\paragraph{Training dynamics across the four regimes.} We next compare how this magnitude mismatch propagates into optimization. Figure~\ref{fig:training-dynamics} tracks Qwen3-4B on mathematical reasoning under GRPO, OPD, fixed-coefficient GRPO+OPD, and SAF over 300 optimization steps, along five diagnostics: actor entropy, student--teacher KL divergence, critic score, response length, and validation accuracy. Fixed-coefficient fusion rapidly reduces actor entropy from approximately $0.35$ to $0.30$ and keeps it near that level, and it also drives the student--teacher KL to the lowest value among the four regimes, consistent with unregulated, heavy-tailed OPD tokens repeatedly pulling the student toward the teacher. GRPO-only, receiving no teacher signal, sits at the opposite end: its entropy rises above $0.42$ and its KL remains the highest. SAF sits between these extremes, maintaining an intermediate entropy of roughly $0.35$--$0.38$ and gradually reducing KL without tracking the teacher as tightly as fixed fusion, matching the intended effect of pairing magnitude control with the warm-up-then-anneal schedule from Section~\ref{sec:temporal-control}.

\paragraph{Connecting dynamics to task performance.} This intermediate regime translates into measurable downstream benefits rather than merely more randomness. SAF's critic score recovers rapidly and ends around $0.84$, close to the strongest GRPO trajectory, whereas fixed fusion plateaus lower, around $0.77$. Response length shows a matching pattern: after a common early transient, fixed fusion settles near $4.5$k tokens, while SAF continues producing responses of roughly $5.3$k--$6.2$k tokens, avoiding the short-response regime of the low-entropy fixed-fusion policy without expanding to the approximately $7$k-token responses of GRPO. Validation accuracy reflects the same ordering: SAF keeps improving late in training and finishes above fixed fusion on both AIME-24 (approximately $0.59$ versus $0.57$) and AIME-25 (approximately $0.53$ versus $0.51$). Notably, fixed fusion tracks the teacher most closely (lowest KL) yet ends with the lowest accuracy and earliest plateau, so closer imitation does not translate into better performance; this is consistent with the teacher being an imperfect proxy, with prolonged full-strength imitation overfitting to that proxy instead of the verifiable reward. Taken together, these diagnostics support a single account: unregulated, full-strength distillation drives KL down at the cost of entropy, response length, and an earlier accuracy plateau, whereas SAF's controls preserve enough teacher guidance to stabilize training while leaving room for continued reward-driven exploration. We treat this analysis as diagnostic; Table~\ref{tab:ablation} isolates the causal contribution of each component, and Appendix~\ref{app:training-dynamics} reports additional parameter-space evidence consistent with this diagnosis.

\subsection{Analysis Across Scales and Domains}

SAF's advantage over fixed fusion persists in all six model--domain settings spanning Qwen3-8B, Qwen3-4B, and Qwen3-1.7B, ranging from 0.51\% to 2.70\%. The gain is not tied to a single capacity regime: on mathematical reasoning, the margin increases monotonically as the student shrinks (0.97\% at 8B, 1.51\% at 4B, 1.85\% at 1.7B), consistent with smaller students relying more on the denser OPD signal once controlled. On code generation the pattern is not monotonic---largest at 4B (2.70\%), still substantial at 8B (1.67\%), and smallest at 1.7B (0.51\%, where OPD-only is already the strongest baseline)---indicating scale interacts with domain rather than acting as a single axis of difficulty. Together with the ablation, these results suggest that temporally controlling a bounded token-level distillation signal is useful across model scales and task domains, while stopping short of claiming robustness to untested teachers, hyperparameters, or scales beyond the 1.7B--8B range studied here.

\section{Conclusion}

This paper studies how to fuse a response-level GRPO advantage with a token-level OPD advantage without either one destabilizing the other. Naive fixed-coefficient summation suffers from a magnitude mismatch, where token-level OPD advantages spike far beyond the bounded GRPO advantage, and a temporal mismatch, where the value of full-strength guidance drifts as the student converges toward the teacher. SAF addresses both via a four-stage, switchable pipeline pairing top-$k$ sparsify-then-$\tanh$-compress magnitude control with warm-up-then-anneal temporal control, adding no auxiliary model, teacher query, or loss term. Across mathematical reasoning and code generation with Qwen3-8B, Qwen3-4B, and Qwen3-1.7B, SAF improves over fixed-coefficient GRPO+OPD in all six settings by 0.51--2.70\%, avoiding the early entropy collapse and lowest accuracy plateau of the fixed-coefficient baseline despite the latter tracking the teacher most tightly. Fusing a sparse, verified reward with a dense, proxy-based signal should thus not reduce to a single mixing coefficient, but calls for separately controlling its magnitude and trust over training.

\bibliographystyle{unsrtnat}
\bibliography{aaai2027_conference}

\begin{thebibliography}{46}
\providecommand{\natexlab}[1]{#1}
\providecommand{\url}[1]{\texttt{#1}}
\expandafter\ifx\csname urlstyle\endcsname\relax
  \providecommand{\doi}[1]{doi: #1}\else
  \providecommand{\doi}{doi: \begingroup \urlstyle{rm}\Url}\fi

\bibitem[Zhang et~al.(2025)Zhang, Zuo, He, Sun, Liu, Jiang, Fan, Tian, Jia, Li, Fu, Lv, Zhang, Zeng, Qu, Li, Wang, Wang, Long, Liu, Xu, Ma, Zhu, Hua, Liu, Li, Chen, Qu, Li, Chen, Yuan, Gao, Li, Ma, Cui, Liu, Qi, Ding, and Zhou]{zhang2025surveyreinforcementlearninglarge}
Kaiyan Zhang, Yuxin Zuo, Bingxiang He, Youbang Sun, Runze Liu, Che Jiang, Yuchen Fan, Kai Tian, Guoli Jia, Pengfei Li, Yu~Fu, Xingtai Lv, Yuchen Zhang, Sihang Zeng, Shang Qu, Haozhan Li, Shijie Wang, Yuru Wang, Xinwei Long, Fangfu Liu, Xiang Xu, Jiaze Ma, Xuekai Zhu, Ermo Hua, Yihao Liu, Zonglin Li, Huayu Chen, Xiaoye Qu, Yafu Li, Weize Chen, Zhenzhao Yuan, Junqi Gao, Dong Li, Zhiyuan Ma, Ganqu Cui, Zhiyuan Liu, Biqing Qi, Ning Ding, and Bowen Zhou.
\newblock A survey of reinforcement learning for large reasoning models, 2025.
\newblock URL \url{https://arxiv.org/abs/2509.08827}.

\bibitem[Shao et~al.(2024)Shao, Wang, Zhu, Xu, Song, Bi, Zhang, Zhang, Li, Wu, and Guo]{shao2024deepseekmathpushinglimitsmathematical}
Zhihong Shao, Peiyi Wang, Qihao Zhu, Runxin Xu, Junxiao Song, Xiao Bi, Haowei Zhang, Mingchuan Zhang, Y.~K. Li, Y.~Wu, and Daya Guo.
\newblock Deepseekmath: Pushing the limits of mathematical reasoning in open language models, 2024.
\newblock URL \url{https://arxiv.org/abs/2402.03300}.

\bibitem[Guo et~al.(2025)Guo, Yang, Zhang, Song, Wang, Zhu, Xu, Zhang, Ma, Bi, Zhang, Yu, Wu, Wu, Gou, Shao, Li, Gao, Liu, Xue, Wang, Wu, Feng, Lu, Zhao, Deng, Ruan, Dai, Chen, Ji, Li, Lin, Dai, Luo, Hao, Chen, Li, Zhang, Xu, Ding, Gao, Qu, Li, Guo, Li, Chen, Yuan, Tu, Qiu, Li, Cai, Ni, Liang, Chen, Dong, Hu, You, Gao, Guan, Huang, Yu, Wang, Zhang, Zhao, Wang, Zhang, Xu, Xia, Zhang, Zhang, Tang, Zhou, Li, Wang, Li, Tian, Huang, Zhang, Wang, Chen, Du, Ge, Zhang, Pan, Wang, Chen, Jin, Chen, Lu, Zhou, Chen, Ye, Wang, Yu, Zhou, Pan, Li, Zhou, Wu, Yun, Pei, Sun, Wang, Zeng, Liu, Liang, Gao, Yu, Zhang, Xiao, An, Liu, Wang, Chen, Nie, Cheng, Liu, Xie, Liu, Yang, Li, Su, Lin, Li, Jin, Shen, Chen, Sun, Wang, Song, Zhou, Wang, Shan, Li, Wang, Wei, Zhang, Xu, Li, Zhao, Sun, Wang, Yu, Zhang, Shi, Xiong, He, Piao, Wang, Tan, Ma, Liu, Guo, Ou, Wang, Gong, Zou, He, Xiong, Luo, You, Liu, Zhou, Zhu, Huang, Li, Zheng, Zhu, Ma, Tang, Zha, Yan, Ren, Ren, Sha, Fu, Xu, Xie, Zhang, Hao, Ma, Yan, Wu, Gu, Zhu, Liu, Li, Xie, Song, Pan, Huang, Xu, Zhang, and Zhang]{Guo_2025}
Daya Guo, Dejian Yang, Haowei Zhang, Junxiao Song, Peiyi Wang, Qihao Zhu, Runxin Xu, Ruoyu Zhang, Shirong Ma, Xiao Bi, Xiaokang Zhang, Xingkai Yu, Yu~Wu, Z.~F. Wu, Zhibin Gou, Zhihong Shao, Zhuoshu Li, Ziyi Gao, Aixin Liu, Bing Xue, Bingxuan Wang, Bochao Wu, Bei Feng, Chengda Lu, Chenggang Zhao, Chengqi Deng, Chong Ruan, Damai Dai, Deli Chen, Dongjie Ji, Erhang Li, Fangyun Lin, Fucong Dai, Fuli Luo, Guangbo Hao, Guanting Chen, Guowei Li, H.~Zhang, Hanwei Xu, Honghui Ding, Huazuo Gao, Hui Qu, Hui Li, Jianzhong Guo, Jiashi Li, Jingchang Chen, Jingyang Yuan, Jinhao Tu, Junjie Qiu, Junlong Li, J.~L. Cai, Jiaqi Ni, Jian Liang, Jin Chen, Kai Dong, Kai Hu, Kaichao You, Kaige Gao, Kang Guan, Kexin Huang, Kuai Yu, Lean Wang, Lecong Zhang, Liang Zhao, Litong Wang, Liyue Zhang, Lei Xu, Leyi Xia, Mingchuan Zhang, Minghua Zhang, Minghui Tang, Mingxu Zhou, Meng Li, Miaojun Wang, Mingming Li, Ning Tian, Panpan Huang, Peng Zhang, Qiancheng Wang, Qinyu Chen, Qiushi Du, Ruiqi Ge, Ruisong Zhang, Ruizhe Pan, Runji Wang, R.~J. Chen, R.~L. Jin, Ruyi Chen, Shanghao Lu, Shangyan Zhou, Shanhuang Chen, Shengfeng Ye, Shiyu Wang, Shuiping Yu, Shunfeng Zhou, Shuting Pan, S.~S. Li, Shuang Zhou, Shaoqing Wu, Tao Yun, Tian Pei, Tianyu Sun, T.~Wang, Wangding Zeng, Wen Liu, Wenfeng Liang, Wenjun Gao, Wenqin Yu, Wentao Zhang, W.~L. Xiao, Wei An, Xiaodong Liu, Xiaohan Wang, Xiaokang Chen, Xiaotao Nie, Xin Cheng, Xin Liu, Xin Xie, Xingchao Liu, Xinyu Yang, Xinyuan Li, Xuecheng Su, Xuheng Lin, X.~Q. Li, Xiangyue Jin, Xiaojin Shen, Xiaosha Chen, Xiaowen Sun, Xiaoxiang Wang, Xinnan Song, Xinyi Zhou, Xianzu Wang, Xinxia Shan, Y.~K. Li, Y.~Q. Wang, Y.~X. Wei, Yang Zhang, Yanhong Xu, Yao Li, Yao Zhao, Yaofeng Sun, Yaohui Wang, Yi~Yu, Yichao Zhang, Yifan Shi, Yiliang Xiong, Ying He, Yishi Piao, Yisong Wang, Yixuan Tan, Yiyang Ma, Yiyuan Liu, Yongqiang Guo, Yuan Ou, Yuduan Wang, Yue Gong, Yuheng Zou, Yujia He, Yunfan Xiong, Yuxiang Luo, Yuxiang You, Yuxuan Liu, Yuyang Zhou, Y.~X. Zhu, Yanping Huang, Yaohui Li, Yi~Zheng, Yuchen Zhu, Yunxian Ma, Ying Tang, Yukun Zha, Yuting Yan, Z.~Z. Ren, Zehui Ren, Zhangli Sha, Zhe Fu, Zhean Xu, Zhenda Xie, Zhengyan Zhang, Zhewen Hao, Zhicheng Ma, Zhigang Yan, Zhiyu Wu, Zihui Gu, Zijia Zhu, Zijun Liu, Zilin Li, Ziwei Xie, Ziyang Song, Zizheng Pan, Zhen Huang, Zhipeng Xu, Zhongyu Zhang, and Zhen Zhang.
\newblock Deepseek-r1 incentivizes reasoning in llms through reinforcement learning.
\newblock \emph{Nature}, 645\penalty0 (8081):\penalty0 633–638, 2025.
\newblock ISSN 1476-4687.
\newblock \doi{10.1038/s41586-025-09422-z}.
\newblock URL \url{http://dx.doi.org/10.1038/s41586-025-09422-z}.

\bibitem[Agarwal et~al.(2024)Agarwal, Vieillard, Zhou, Stanczyk, Ramos, Geist, and Bachem]{agarwal2024onpolicydistillationlanguagemodels}
Rishabh Agarwal, Nino Vieillard, Yongchao Zhou, Piotr Stanczyk, Sabela Ramos, Matthieu Geist, and Olivier Bachem.
\newblock On-policy distillation of language models: Learning from self-generated mistakes, 2024.
\newblock URL \url{https://arxiv.org/abs/2306.13649}.

\bibitem[Gu et~al.(2024)Gu, Dong, Wei, and Huang]{gu2024minillm}
Yuxian Gu, Li~Dong, Furu Wei, and Minlie Huang.
\newblock Mini{LLM}: Knowledge distillation of large language models.
\newblock In \emph{The Twelfth International Conference on Learning Representations}, 2024.
\newblock URL \url{https://openreview.net/forum?id=5h0qf7IBZZ}.

\bibitem[Liu et~al.(2026)Liu, Zhang, Zhang, and Gu]{liu2026selfdistilledpolicygradient}
Yifeng Liu, Shiyuan Zhang, Yifan Zhang, and Quanquan Gu.
\newblock Self-distilled policy gradient, 2026.
\newblock URL \url{https://arxiv.org/abs/2606.04036}.

\bibitem[Yang et~al.(2026{\natexlab{a}})Yang, Liu, Xie, Yang, Yang, and Lin]{yang2026learning}
Wenkai Yang, Weijie Liu, Ruobing Xie, Kai Yang, Saiyong Yang, and Yankai Lin.
\newblock Learning beyond teacher: Generalized on-policy distillation with reward extrapolation.
\newblock \emph{arXiv preprint arXiv:2602.12125}, 2026{\natexlab{a}}.

\bibitem[Cui et~al.(2025{\natexlab{a}})Cui, Zhang, Chen, Yuan, Wang, Zuo, Li, Fan, Chen, Chen, et~al.]{cui2025entropymechanismreinforcementlearning}
Ganqu Cui, Yuchen Zhang, Jiacheng Chen, Lifan Yuan, Zhi Wang, Yuxin Zuo, Haozhan Li, Yuchen Fan, Huayu Chen, Weize Chen, et~al.
\newblock The entropy mechanism of reinforcement learning for reasoning language models, 2025{\natexlab{a}}.
\newblock URL \url{https://arxiv.org/abs/2505.22617}.

\bibitem[Wang et~al.(2026{\natexlab{a}})Wang, Li, Bai, Zhang, Deng, Lan, and Wang]{wang2026distilledreinforcementlearningllm}
Chen Wang, Zhaochun Li, Jionghao Bai, Yining Zhang, Hexuan Deng, Ge~Lan, and Yue Wang.
\newblock Distilled reinforcement learning for llm post-training, 2026{\natexlab{a}}.
\newblock URL \url{https://arxiv.org/abs/2607.17247}.

\bibitem[{MiMo Core Team}(2026)]{coreteam2026mimov2flashtechnicalreport}
{MiMo Core Team}.
\newblock Mimo-v2-flash technical report, 2026.
\newblock URL \url{https://arxiv.org/abs/2601.02780}.

\bibitem[Wang et~al.(2026{\natexlab{b}})Wang, Wang, Chen, Xue, Fang, Yu, and Wong]{wang2026demystifyingonpolicydistillationroles}
Rui Wang, Hongru Wang, Yi~Chen, Boyang Xue, Tianqing Fang, Wenhao Yu, and Kam-Fai Wong.
\newblock Demystifying on-policy distillation: Roles, pathologies, and regulations, 2026{\natexlab{b}}.
\newblock URL \url{https://arxiv.org/abs/2607.13399}.

\bibitem[Xing et~al.(2026)Xing, Wang, Gao, Li, and Tang]{xing2026trustregiononpolicydistillation}
Xingrun Xing, Haoqing Wang, Boyan Gao, Ziheng Li, and Yehui Tang.
\newblock Trust region on-policy distillation, 2026.
\newblock URL \url{https://arxiv.org/abs/2606.01249}.

\bibitem[Zhao et~al.(2026)Zhao, Tong, Fan, Nie, Li, and Shen]{zhao2026poweropdstabilizingonpolicydistillation}
Anhao Zhao, Junlong Tong, Yingqi Fan, Ping Nie, Wenjie Li, and Xiaoyu Shen.
\newblock Poweropd: Stabilizing on-policy distillation with bounded power transformation, 2026.
\newblock URL \url{https://arxiv.org/abs/2606.17199}.

\bibitem[Luo et~al.(2026)Luo, Chuang, Wang, Xu, Han, Zhang, and Braverman]{luo2026demystifyingopdlengthinflation}
Feng Luo, Yu-Neng Chuang, Guanchu Wang, Zicheng Xu, Xiaotian Han, Tianyi Zhang, and Vladimir Braverman.
\newblock Demystifying opd: Length inflation and stabilization strategies for large language models, 2026.
\newblock URL \url{https://arxiv.org/abs/2604.08527}.

\bibitem[Schulman et~al.(2017)Schulman, Wolski, Dhariwal, Radford, and Klimov]{schulman2017proximalpolicyoptimizationalgorithms}
John Schulman, Filip Wolski, Prafulla Dhariwal, Alec Radford, and Oleg Klimov.
\newblock Proximal policy optimization algorithms, 2017.
\newblock URL \url{https://arxiv.org/abs/1707.06347}.

\bibitem[Liu et~al.(2025)Liu, Chen, Li, Qi, Pang, Du, Lee, and Lin]{liu2025understandingr1zeroliketrainingcritical}
Zichen Liu, Changyu Chen, Wenjun Li, Penghui Qi, Tianyu Pang, Chao Du, Wee~Sun Lee, and Min Lin.
\newblock Understanding r1-zero-like training: A critical perspective, 2025.
\newblock URL \url{https://arxiv.org/abs/2503.20783}.

\bibitem[Yu et~al.(2025)Yu, Zhang, Zhu, Yuan, Zuo, Yue, Dai, Fan, Liu, Liu, et~al.]{yu2025dapoopensourcellmreinforcement}
Qiying Yu, Zheng Zhang, Ruofei Zhu, Yufeng Yuan, Xiaochen Zuo, Yu~Yue, Weinan Dai, Tiantian Fan, Gaohong Liu, Lingjun Liu, et~al.
\newblock Dapo: An open-source llm reinforcement learning system at scale, 2025.
\newblock URL \url{https://arxiv.org/abs/2503.14476}.

\bibitem[Zheng et~al.(2025)Zheng, Liu, Li, Chen, Yu, Gao, Dang, Liu, Men, Yang, et~al.]{zheng2025groupsequencepolicyoptimization}
Chujie Zheng, Shixuan Liu, Mingze Li, Xiong-Hui Chen, Bowen Yu, Chang Gao, Kai Dang, Yuqiong Liu, Rui Men, An~Yang, et~al.
\newblock Group sequence policy optimization, 2025.
\newblock URL \url{https://arxiv.org/abs/2507.18071}.

\bibitem[Lightman et~al.(2023)Lightman, Kosaraju, Burda, Edwards, Baker, Lee, Leike, Schulman, Sutskever, and Cobbe]{lightman2023letsverifystepstep}
Hunter Lightman, Vineet Kosaraju, Yura Burda, Harri Edwards, Bowen Baker, Teddy Lee, Jan Leike, John Schulman, Ilya Sutskever, and Karl Cobbe.
\newblock Let's verify step by step, 2023.
\newblock URL \url{https://arxiv.org/abs/2305.20050}.

\bibitem[Wang et~al.(2024)Wang, Li, Shao, Xu, Dai, Li, Chen, Wu, and Sui]{wang2024mathshepherdverifyreinforcellms}
Peiyi Wang, Lei Li, Zhihong Shao, R.~X. Xu, Damai Dai, Yifei Li, Deli Chen, Y.~Wu, and Zhifang Sui.
\newblock Math-shepherd: Verify and reinforce llms step-by-step without human annotations, 2024.
\newblock URL \url{https://arxiv.org/abs/2312.08935}.

\bibitem[Hinton et~al.(2015)Hinton, Vinyals, and Dean]{hinton2015distillingknowledgeneuralnetwork}
Geoffrey Hinton, Oriol Vinyals, and Jeff Dean.
\newblock Distilling the knowledge in a neural network, 2015.
\newblock URL \url{https://arxiv.org/abs/1503.02531}.

\bibitem[Kim and Rush(2016)]{kim2016sequencelevelknowledgedistillation}
Yoon Kim and Alexander~M. Rush.
\newblock Sequence-level knowledge distillation, 2016.
\newblock URL \url{https://arxiv.org/abs/1606.07947}.

\bibitem[Song and Zheng(2026)]{song2026surveyonpolicydistillationlarge}
Mingyang Song and Mao Zheng.
\newblock A survey of on-policy distillation for large language models, 2026.
\newblock URL \url{https://arxiv.org/abs/2604.00626}.

\bibitem[Zhang(2026)]{zhang2026formuladrivensurveyresearchagenda}
Bowen Zhang.
\newblock A formula-driven survey and research agenda for on-policy distillation, 2026.
\newblock URL \url{https://arxiv.org/abs/2606.22793}.

\bibitem[Ko et~al.(2024)Ko, Kim, Chen, and Yun]{ko2024distillmstreamlineddistillationlarge}
Jongwoo Ko, Sungnyun Kim, Tianyi Chen, and Se-Young Yun.
\newblock Distillm: Towards streamlined distillation for large language models, 2024.
\newblock URL \url{https://arxiv.org/abs/2402.03898}.

\bibitem[Zhong et~al.(2026)Zhong, Zheng, Song, Lin, Sun, Jiang, Wang, and Fang]{zhong2026sod}
Qiyong Zhong, Mao Zheng, Mingyang Song, Xin Lin, Jie Sun, Houcheng Jiang, Xiang Wang, and Junfeng Fang.
\newblock Sod: Step-wise on-policy distillation for small language model agents.
\newblock \emph{arXiv preprint arXiv:2605.07725}, 2026.

\bibitem[Jin et~al.(2026)Jin, Min, Yang, Wei, Zhou, Kadhe, Baracaldo, and Lee]{jin2026entropyawareonpolicydistillationlanguage}
Woogyeol Jin, Taywon Min, Yongjin Yang, Dennis Wei, Yi~Zhou, Swanand~Ravindra Kadhe, Nathalie Baracaldo, and Kimin Lee.
\newblock Entropy-aware on-policy distillation of language models, 2026.
\newblock URL \url{https://arxiv.org/abs/2603.07079}.

\bibitem[Ko et~al.(2026)Ko, Abdali, Kim, Chen, and Cameron]{ko2026scalingreasoningefficientlyrelaxed}
Jongwoo Ko, Sara Abdali, Young~Jin Kim, Tianyi Chen, and Pashmina Cameron.
\newblock Scaling reasoning efficiently via relaxed on-policy distillation, 2026.
\newblock URL \url{https://arxiv.org/abs/2603.11137}.

\bibitem[Yang et~al.(2026{\natexlab{b}})Yang, Zhu, Song, Wang, Xia, Zheng, Ma, Chen, Wang, Zhao, and Chen]{yang2026oprdonpolicyrepresentationdistillation}
Shenzhi Yang, Guangcheng Zhu, Bowen Song, Haobo Wang, Mingxuan Xia, Xing Zheng, Yingfan Ma, Zhongqi Chen, Weiqiang Wang, Junbo Zhao, and Gang Chen.
\newblock Oprd: On-policy representation distillation, 2026{\natexlab{b}}.
\newblock URL \url{https://arxiv.org/abs/2606.06021}.

\bibitem[Oh et~al.(2026)Oh, Song, Choi, Choi, and Jo]{oh2026klklonpolicydistillation}
Minjae Oh, Sangjun Song, Gyubin Choi, Yunho Choi, and Yohan Jo.
\newblock Kl for a kl: On-policy distillation with control variate baseline, 2026.
\newblock URL \url{https://arxiv.org/abs/2605.07865}.

\bibitem[Xu et~al.(2025)Xu, Zhu, Deng, Li, Hou, Wang, Shang, Xu, and Mi]{xu2025kdrlposttrainingreasoningllms}
Hongling Xu, Qi~Zhu, Heyuan Deng, Jinpeng Li, Lu~Hou, Yasheng Wang, Lifeng Shang, Ruifeng Xu, and Fei Mi.
\newblock Kdrl: Post-training reasoning llms via unified knowledge distillation and reinforcement learning, 2025.
\newblock URL \url{https://arxiv.org/abs/2506.02208}.

\bibitem[Yang et~al.(2026{\natexlab{c}})Yang, Qin, Si, Chen, Gu, Yao, Lin, Wang, Wang, and Duan]{yang2026selfdistilledrlvr}
Chenxu Yang, Chuanyu Qin, Qingyi Si, Minghui Chen, Naibin Gu, Dingyu Yao, Zheng Lin, Weiping Wang, Jiaqi Wang, and Nan Duan.
\newblock Self-distilled rlvr, 2026{\natexlab{c}}.
\newblock URL \url{https://arxiv.org/abs/2604.03128}.

\bibitem[Lu et~al.(2026)Lu, Yao, Han, Wang, Wu, Gu, Cai, Lu, Xiao, Zhuang, and Shen]{lu2026sdar}
Zhengxi Lu, Zhiyuan Yao, Zhuowen Han, Zi-Han Wang, Jinyang Wu, Qi~Gu, Xunliang Cai, Weiming Lu, Jun Xiao, Yueting Zhuang, and Yongliang Shen.
\newblock Self-distilled agentic reinforcement learning, 2026.
\newblock URL \url{https://arxiv.org/abs/2605.15155}.

\bibitem[Tan et~al.(2026)Tan, Zong, Li, and Chen]{tan2026atodannealedturnawareonpolicy}
Qitai Tan, Zefang Zong, Yang Li, and Peng Chen.
\newblock Atod: Annealed turn-aware on-policy distillation for multi-turn autonomous agents, 2026.
\newblock URL \url{https://arxiv.org/abs/2606.27814}.

\bibitem[Pan et~al.(2026)Pan, Tao, Zhai, Zhang, Liu, Ding, Liu, and Wen]{pan2026rlcsdreinforcementlearningcontrastive}
Leyi Pan, Shuchang Tao, Yunpeng Zhai, Lingzhe Zhang, Zhaoyang Liu, Bolin Ding, Aiwei Liu, and Lijie Wen.
\newblock Rlcsd: Reinforcement learning with contrastive on-policy self-distillation, 2026.
\newblock URL \url{https://arxiv.org/abs/2606.11709}.

\bibitem[Wang et~al.(2026{\natexlab{c}})Wang, Ouyang, Chen, Hu, Pan, Li, and Guo]{wang2026tracedistillingmatterstokenrouted}
Jiaxuan Wang, Xuan Ouyang, Zhiyu Chen, Yulan Hu, Zheng Pan, Xin Li, and Lan-Zhe Guo.
\newblock Trace: Distilling where it matters via token-routed self on-policy alignment, 2026{\natexlab{c}}.
\newblock URL \url{https://arxiv.org/abs/2605.10194}.

\bibitem[Sheng et~al.(2025)Sheng, Zhang, Ye, Wu, Zhang, Zhang, Peng, Lin, and Wu]{Sheng_2025}
Guangming Sheng, Chi Zhang, Zilingfeng Ye, Xibin Wu, Wang Zhang, Ru~Zhang, Yanghua Peng, Haibin Lin, and Chuan Wu.
\newblock Hybridflow: A flexible and efficient rlhf framework.
\newblock In \emph{Proceedings of the Twentieth European Conference on Computer Systems}, pages 1279--1297. ACM, 2025.
\newblock \doi{10.1145/3689031.3696075}.

\bibitem[Yang et~al.(2025)Yang, Li, Yang, Zhang, Hui, Zheng, Yu, Gao, Huang, Lv, et~al.]{yang2025qwen3technicalreport}
An~Yang, Anfeng Li, Baosong Yang, Beichen Zhang, Binyuan Hui, Bo~Zheng, Bowen Yu, Chang Gao, Chengen Huang, Chenxu Lv, et~al.
\newblock Qwen3 technical report, 2025.
\newblock URL \url{https://arxiv.org/abs/2505.09388}.

\bibitem[He et~al.(2025)He, Liang, Xu, Liu, Chen, Wang, Song, Yu, Liang, Wang, et~al.]{deepmath}
Zhiwei He, Tian Liang, Jiahao Xu, Qiuzhi Liu, Xingyu Chen, Yue Wang, Linfeng Song, Dian Yu, Zhenwen Liang, Wenxuan Wang, et~al.
\newblock Deepmath-103k: A large-scale, challenging, decontaminated, and verifiable mathematical dataset for advancing reasoning.
\newblock \emph{arXiv preprint arXiv:2504.11456}, 2025.

\bibitem[Cui et~al.(2025{\natexlab{b}})Cui, Yuan, Wang, Wang, Li, He, Fan, Yu, Xu, Chen, et~al.]{prime}
Ganqu Cui, Lifan Yuan, Zefan Wang, Hanbin Wang, Wendi Li, Bingxiang He, Yuchen Fan, Tianyu Yu, Qixin Xu, Weize Chen, et~al.
\newblock Process reinforcement through implicit rewards.
\newblock \emph{arXiv preprint arXiv:2502.01456}, 2025{\natexlab{b}}.

\bibitem[AI-MO(2024)]{aime2024}
AI-MO.
\newblock Aime 2024.
\newblock \url{https://huggingface.co/datasets/AI-MO/aimo-validation-aime}, 2024.

\bibitem[OpenCompass(2025)]{aime2025}
OpenCompass.
\newblock Aime 2025.
\newblock \url{https://huggingface.co/datasets/opencompass/AIME2025}, 2025.

\bibitem[Balunovi\'c et~al.(2025)Balunovi\'c, Dekoninck, Petrov, Jovanovi\'c, and Vechev]{hmmt25}
Mislav Balunovi\'c, Jasper Dekoninck, Ivo Petrov, Nikola Jovanovi\'c, and Martin Vechev.
\newblock Matharena: Evaluating llms on uncontaminated math competitions, 2025.
\newblock URL \url{https://matharena.ai/}.

\bibitem[Liu et~al.(2023)Liu, Xia, Wang, and Zhang]{evalplus}
Jiawei Liu, Chunqiu~Steven Xia, Yuyao Wang, and Lingming Zhang.
\newblock Is your code generated by chat{GPT} really correct? rigorous evaluation of large language models for code generation.
\newblock In \emph{Thirty-seventh Conference on Neural Information Processing Systems}, 2023.
\newblock URL \url{https://openreview.net/forum?id=1qvx610Cu7}.

\bibitem[Jain et~al.(2024)Jain, Han, Gu, Li, Yan, Solar-Lezama, Sen, and Stoica]{lcb}
Naman Jain, King Han, Alex Gu, Wen-Ding Li, Fanjia Yan, Armando Solar-Lezama, Koushik Sen, and Ion Stoica.
\newblock Livecodebench: Holistic and contamination free evaluation of large language models for code.
\newblock \emph{arXiv preprint arXiv:2403.07974}, 2024.

\bibitem[Cai et~al.(2026)Cai, Cao, Lin, Luo, Xu, Yang, Liu, Yang, Zhao, Sun, Liu, and Fang]{cai2026learningforeseeunveilingunlocking}
Yuchen Cai, Ding Cao, Liang Lin, Chunxi Luo, Xin Xu, Kai Yang, Weijie Liu, Saiyong Yang, Tianxiang Zhao, Guangzhong Sun, Guiquan Liu, and Junfeng Fang.
\newblock Learning to foresee: Unveiling the unlocking efficiency of on-policy distillation, 2026.
\newblock URL \url{https://arxiv.org/abs/2605.11739}.

\end{thebibliography}

\clearpage

\appendix

\section{Detailed Experiment Settings}
\label{app:training-settings}

We use Qwen3-30B-A3B-Instruct-2507 as the teacher and initialize the student from Qwen3-8B, Qwen3-4B, or Qwen3-1.7B. Math training uses the filtered DeepMath split containing 57K problems of difficulty level at least 6, while code training uses the 25K-problem Eurus-RL-Code dataset. A rule-based verifier assigns reward 1 when the final mathematical answer is correct or all code unit tests pass, and 0 otherwise. Student responses are sampled on policy, and the teacher scores those same responses token by token rather than generating separate trajectories.

\paragraph{GRPO-based methods.} GRPO-only, GRPO+OPD (fixed), and SAF use the same optimization and rollout configuration at a given student scale and domain; only their advantage construction differs. We optimize these methods for 300 steps on mathematics and 200 steps on code. Table~\ref{tab:grpo-training-settings} lists their shared hyperparameters.

\begin{table}[htbp]
\centering
\begin{tabular}{lcc}
\toprule
Hyperparameter & Mathematics & Code \\
\midrule
Train batch size & 128 & 128 \\
Micro batch size & 128 & 128 \\
Responses per prompt ($G$) & 8 & 8 \\
Maximum prompt length & 2,048 & 2,048 \\
Maximum response length & 16,384 & 8,192 \\
Rollout temperature & 1.0 & 1.0 \\
Rollout top-$p$ & 1.0 & 1.0 \\
Actor learning rate & $1\times10^{-6}$ & $1\times10^{-6}$ \\
Optimization steps & 300 & 200 \\
Actor KL-loss coefficient & 0.0 & 0.0 \\
\bottomrule
\end{tabular}
\caption{Training hyperparameters of the GRPO-based methods: GRPO-only, GRPO+OPD (fixed), and SAF.}
\label{tab:grpo-training-settings}
\end{table}

\paragraph{OPD-only.} OPD-only uses separate configurations for mathematical reasoning and code generation, as summarized in Table~\ref{tab:opd-only-training-settings}. Both domains use a batch size of 1,024, one student rollout per prompt, and a learning rate of $1\times10^{-6}$. We optimize OPD-only for 100 steps on mathematics and 50 steps on code; the corresponding maximum response lengths are 16,384 and 8,192 tokens.

\begin{table}[htbp]
\centering
\begin{tabular}{lcc}
\toprule
Hyperparameter & Mathematics & Code \\
\midrule
Batch size & 1,024 & 1,024 \\
Responses per prompt & 1 & 1 \\
Maximum prompt length & 2,048 & 2,048 \\
Maximum response length & 16,384 & 8,192 \\
Rollout temperature & 1.0 & 1.0 \\
Rollout top-$p$ & 1.0 & 1.0 \\
Learning rate & $1\times10^{-6}$ & $1\times10^{-6}$ \\
Optimization steps & 100 & 50 \\
\bottomrule
\end{tabular}
\caption{Training hyperparameters of OPD-only for mathematical reasoning and code generation.}
\label{tab:opd-only-training-settings}
\end{table}

\paragraph{Advantage configurations.} For GRPO-only, we set the OPD term to zero. For OPD-only, we set the verifier-derived GRPO advantage to zero. GRPO+OPD (fixed) uses a unit OPD coefficient, corresponding to the 1:1 fusion $A_i^{\mathrm{GRPO}}+A_{i,t}^{\mathrm{OPD}}$, with all four SAF controls disabled. Full SAF starts from the same unit coefficient and applies the settings in Table~\ref{tab:saf-training-settings}. The top-$k$ threshold is recomputed independently for every response. Warm-up lasts at most 100 optimization steps and may terminate earlier when the relative student--teacher KL decrease reaches 0.2. After warm-up, the local annealing counter spans all remaining steps, so the OPD coefficient reaches zero at the end of the domain-specific training budget rather than using a separately tuned decay duration.

\begin{table}[htbp]
\centering
\begin{tabular}{ll}
\toprule
Hyperparameter & Value \\
\midrule
Per-response retention ratio $k$ & 20\% \\
$\tanh$ compression coefficient $c$ & 0.1 \\
Maximum warm-up steps $S_{\text{warmup}}$ & 100 \\
Relative KL-drop threshold $\delta$ & 0.2 \\
Initial OPD coefficient & 1.0 \\
Annealing floor $c_{\min}$ & 0.0 \\
Annealing duration & Remaining training steps \\
\bottomrule
\end{tabular}
\caption{Method-specific hyperparameters for the selected SAF configuration.}
\label{tab:saf-training-settings}
\end{table}

\paragraph{Evaluation settings.} For every model, we use a temperature of 1.0, top-$p$ of 1.0, and a maximum generation length of 16,384 tokens, except that HumanEval+ and MBPP+ use greedy decoding. We sample 32 solutions per problem on each mathematics benchmark (AIME-24, AIME-25, HMMT25-Feb, and HMMT25-Nov) and report the average accuracy across samples, whereas on LiveCodeBench v6 (February--May 2025) we sample four solutions per problem and report pass@1, i.e., a problem counts as solved if at least one sampled solution passes. Mathematical answers are checked with Math-Verify, while code submissions are scored by the benchmark-provided unit tests. These decoding and scoring settings are held fixed across Base, all baselines, and SAF.

\paragraph{Computing infrastructure.} All experiments are run on a single node equipped with 8$\times$NVIDIA H200 GPUs (140GB HBM3e memory per GPU, Hopper architecture), 96 physical CPU cores (184 logical cores), and 1.5TB of system memory.

\section{Training Dynamics: Supplementary Parameter-Space Analysis}
\label{app:training-dynamics}

Section~\ref{sec:motivation-and-dynamics} reports the optimization trajectories of Qwen3-4B trained on mathematical reasoning tasks under GRPO, OPD, fixed-coefficient GRPO+OPD, and SAF (Figure~\ref{fig:training-dynamics}), based on entropy, student--teacher KL, critic score, response length, and validation accuracy. This appendix complements that analysis with a more exploratory, parameter-space view of the same training regimes.

\begin{figure*}[!t]
    \centering
    \begin{minipage}[t]{0.32\textwidth}
        \centering
        \includegraphics[width=\linewidth]{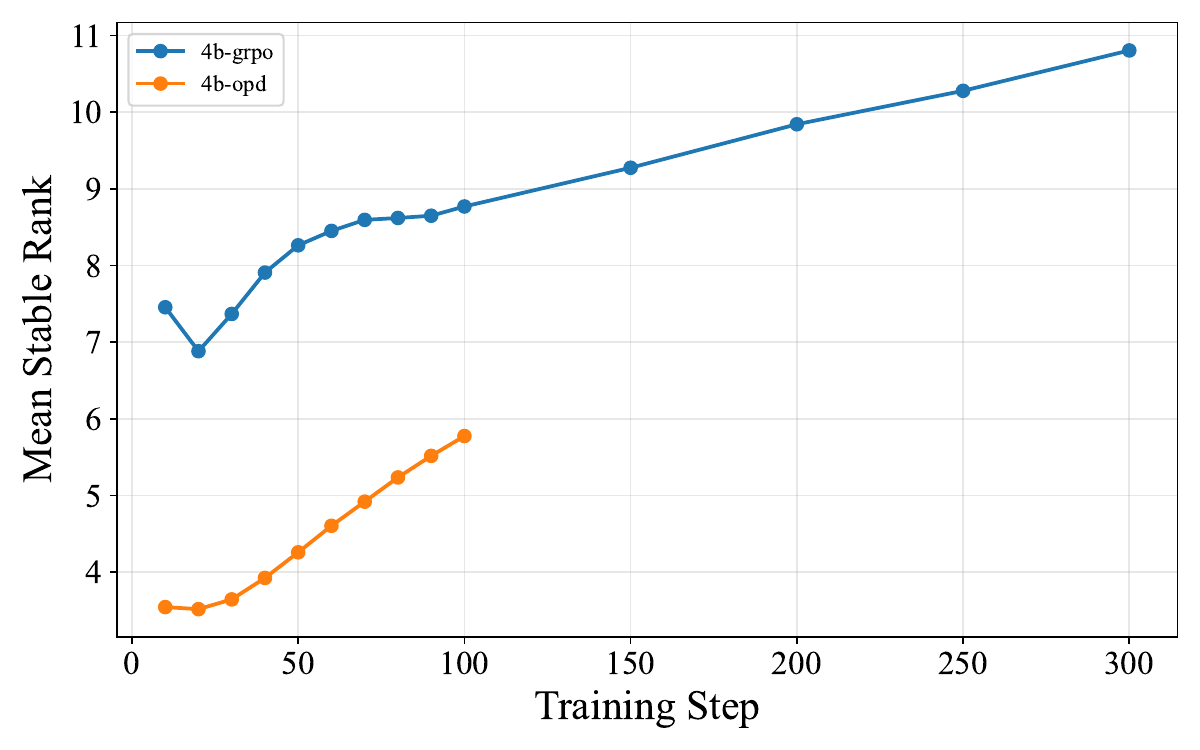}\\
        \small (a) Stable rank
    \end{minipage}\hfill
    \begin{minipage}[t]{0.32\textwidth}
        \centering
        \includegraphics[width=\linewidth]{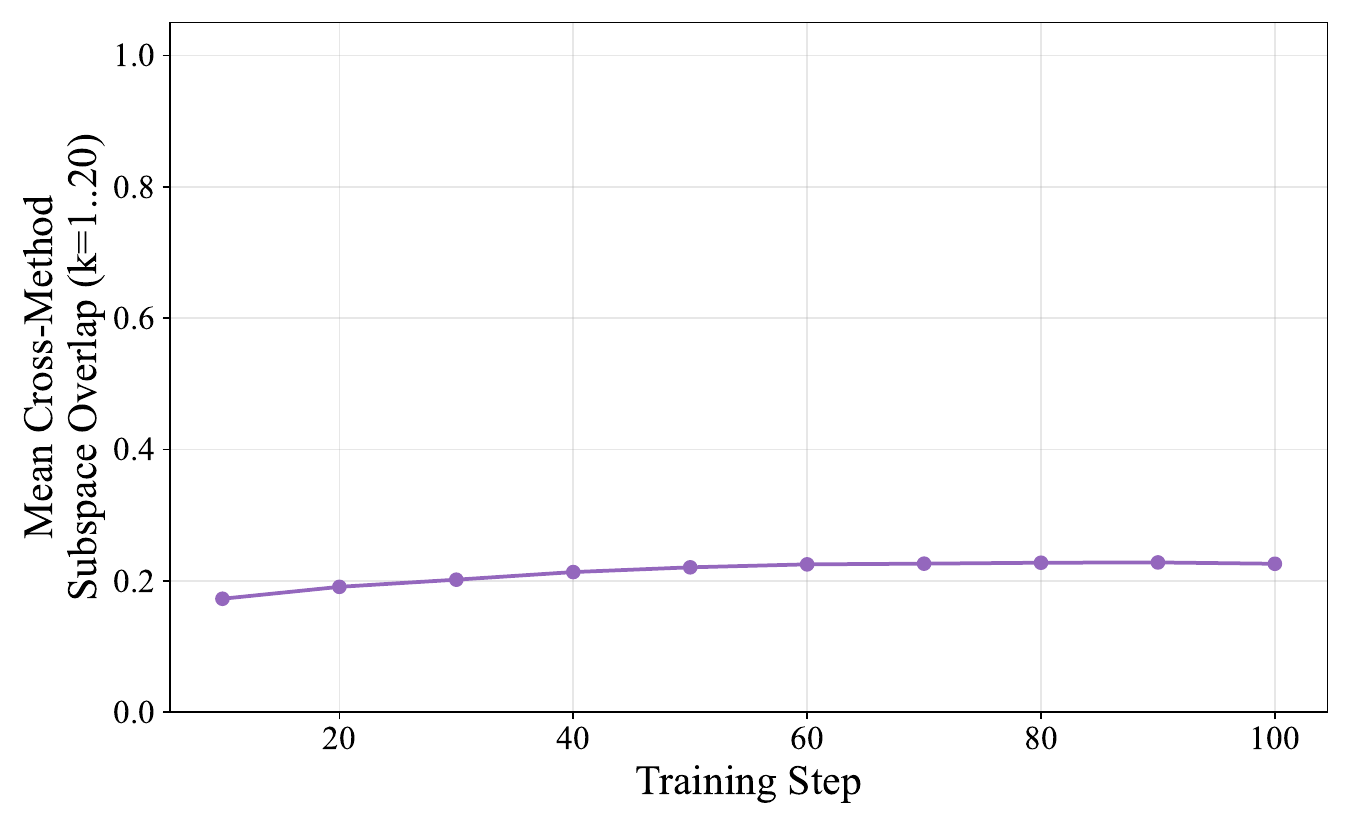}\\
        \small (b) Output-subspace overlap
    \end{minipage}\hfill
    \begin{minipage}[t]{0.32\textwidth}
        \centering
        \includegraphics[width=\linewidth]{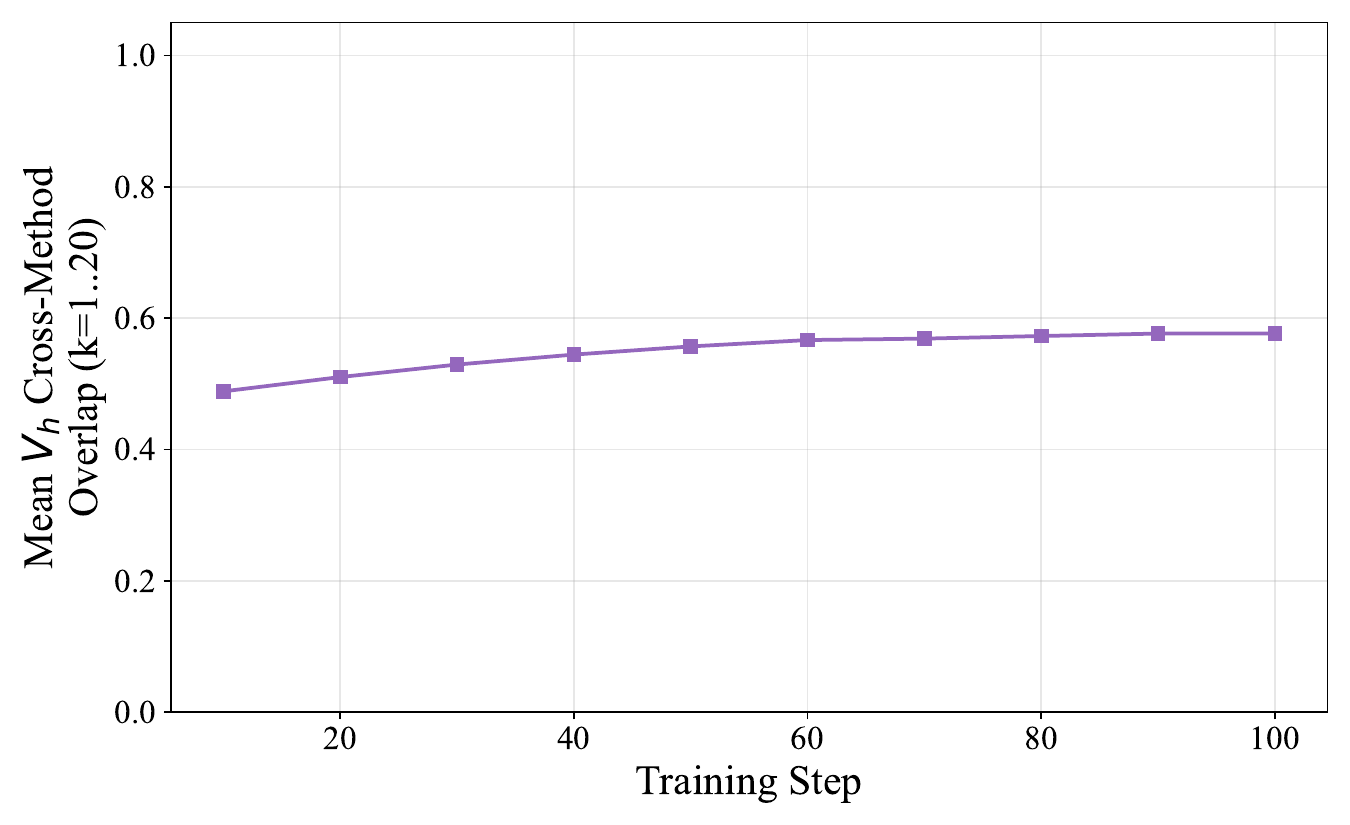}\\
        \small (c) Input-subspace overlap
    \end{minipage}
    \caption{Supplementary parameter-space update geometry for the GRPO-only and OPD-only checkpoints of Qwen3-4B on mathematical reasoning: (a) mean stable rank of the update matrices, (b) output (left-singular) subspace overlap, and (c) input (right-singular) subspace overlap between the two methods.}
    \label{fig:update-geometry}
\end{figure*}

\paragraph{Parameter-space update geometry (supplementary).} The diagnostics in Section~\ref{sec:motivation-and-dynamics} are all computed on the advantage signal or on training curves. As an additional, more exploratory piece of evidence, we also inspected the parameter-space update matrices $\Delta W^{(t)} = W_{\text{trained}}^{(t)} - W_{\text{base}}$ produced by the same GRPO-only and OPD-only checkpoints reported in Tables~\ref{tab:grpo-training-settings} and~\ref{tab:opd-only-training-settings}, restricted to the first 300 optimization steps and analyzed with layer-wise SVD. Figure~\ref{fig:update-geometry} summarizes three quantities computed over the first 100 shared optimization steps (the horizon at which OPD-only, which trains for 100 steps on mathematics, was checkpointed): (a) the mean stable rank $\|\Delta W\|_F^2/\|\Delta W\|_{\mathrm{op}}^2$ of each method's update matrices, (b) the cross-method overlap of their left singular subspaces (output/``write'' directions), and (c) the cross-method overlap of their right singular subspaces (input/``read'' directions). Consistent with the qualitative roles assigned to the two signals in Section~\ref{sec:motivation-and-dynamics}, OPD's updates exhibit a substantially lower stable rank than GRPO's throughout this horizon, indicating that its update energy is more concentrated along a small number of directions, while GRPO's updates remain comparatively higher-rank and more diffuse. The two methods' updates also read from similar input subspaces (higher right-singular overlap) while writing to largely different output subspaces (markedly lower left-singular overlap), suggesting that GRPO and OPD attend to overlapping input features but push the policy in different directions. We report this as a supplementary, exploratory observation at the level of parameter geometry, rather than as evidence directly supporting SAF's advantage-level design; we did not build any part of SAF's magnitude or temporal control on this analysis, and a rigorous treatment connecting update-matrix geometry to advantage-fusion design is left to future work~\cite{cai2026learningforeseeunveilingunlocking}.

\paragraph{Metric definitions.} We define the four quantities used throughout this appendix (stable rank, left/right singular-subspace overlap, and the two weight-drift statistics) precisely below, all computed layer-by-layer from the per-step update matrix $\Delta W^{(t)}_\ell = W^{(t)}_{\text{trained},\ell} - W^0_\ell$, where $\ell$ indexes a $2$D weight matrix (\texttt{q/k/v/o\_proj} or \texttt{gate/up/down\_proj}) at a given transformer layer and $W^0_\ell$ is the corresponding base-model weight. Every $\Delta W^{(t)}_\ell$ is factorized with the (economy) singular value decomposition $\Delta W^{(t)}_\ell = U^{(t)}_\ell \Sigma^{(t)}_\ell (V^{(t)}_\ell)^{\top}$, with singular values $\sigma_1 \geq \sigma_2 \geq \cdots \geq \sigma_r \geq 0$ ($r = \min(m,n)$); $\|\Delta W\|_F = \|\boldsymbol{\sigma}\|_2 = \bigl(\sum_i \sigma_i^2\bigr)^{1/2}$ denotes the Frobenius norm and $\|\Delta W\|_{\mathrm{op}} = \sigma_1$ the operator (spectral) norm. Columns of $U^{(t)}_\ell$ span the \emph{output} (``write'') subspace and rows of $(V^{(t)}_\ell)^{\top}$ span the \emph{input} (``read'') subspace of the update.

\emph{Stable rank.} For a single matrix, the stable rank is the energy-weighted effective dimensionality
\begin{equation}
\mathrm{StableRank}\bigl(\Delta W^{(t)}_\ell\bigr) = \frac{\|\Delta W^{(t)}_\ell\|_F^2}{\|\Delta W^{(t)}_\ell\|_{\mathrm{op}}^2} = \frac{\sum_{i=1}^{r}\sigma_i^2}{\sigma_1^2} \in [1, r],
\end{equation}
which equals $1$ when all update energy is concentrated in the top singular direction and equals $r$ when energy is spread uniformly across all directions; the curves in Figure~\ref{fig:update-geometry}(a) average this quantity over all $2$D weight matrices at a given step for each method.

\emph{Cross-method subspace overlap.} Let $U_{X,k}^{(t)} \in \mathbb{R}^{m \times k}$ (resp.\ $V_{X,k}^{(t)} \in \mathbb{R}^{n \times k}$) collect the top-$k$ left (resp.\ right) singular vectors of method $X \in \{\text{GRPO}, \text{OPD}\}$'s update matrix for the same layer at step $t$, with $k{=}20$ throughout. The Grassmann-style overlap between the two methods' output subspaces is the mean singular value of their cross-projection,
\begin{equation}
\mathrm{Overlap}_{U}(t) = \frac{1}{k}\sum_{j=1}^{k} s_j\!\left(\bigl(U_{\mathrm{GRPO},k}^{(t)}\bigr)^{\top} U_{\mathrm{OPD},k}^{(t)}\right) \in [0,1],
\end{equation}
where $s_j(\cdot)$ denotes the $j$-th singular value; $\mathrm{Overlap}_{U}(t) \approx 1$ indicates the two methods write to nearly identical output directions, while $\mathrm{Overlap}_{U}(t) \approx 0$ indicates (near-)orthogonal output subspaces. The input-subspace overlap $\mathrm{Overlap}_{V}(t)$ is defined identically with $V_{X,k}^{(t)}$ in place of $U_{X,k}^{(t)}$. Both quantities are averaged over layers and over the 2D weight matrices within each layer before plotting Figure~\ref{fig:update-geometry}(b)--(c).

\begin{figure*}[!t]
    \centering
    \begin{minipage}[t]{0.24\textwidth}
        \centering
        \includegraphics[width=\linewidth]{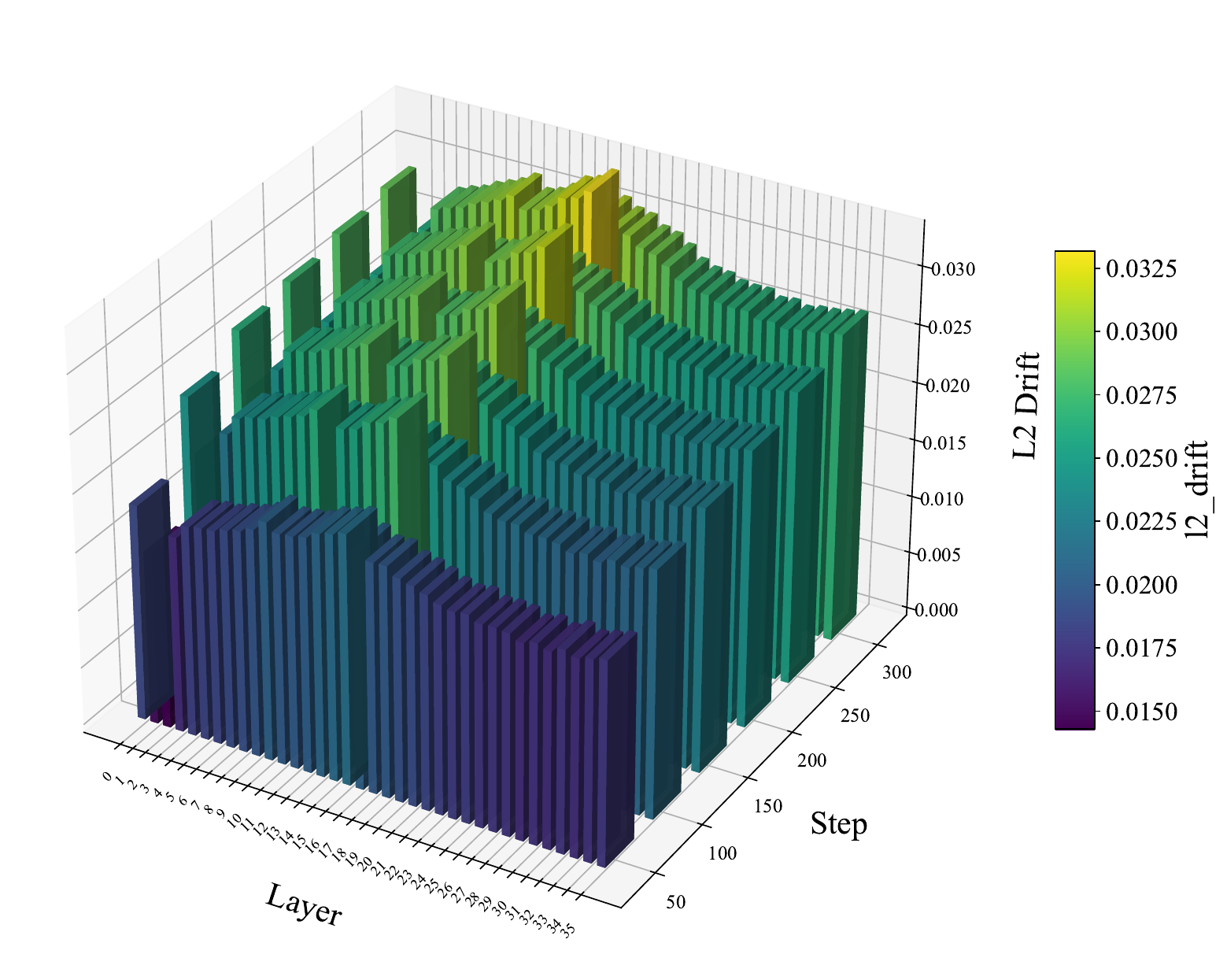}\\
        \small (a) GRPO-only
    \end{minipage}\hfill
    \begin{minipage}[t]{0.24\textwidth}
        \centering
        \includegraphics[width=\linewidth]{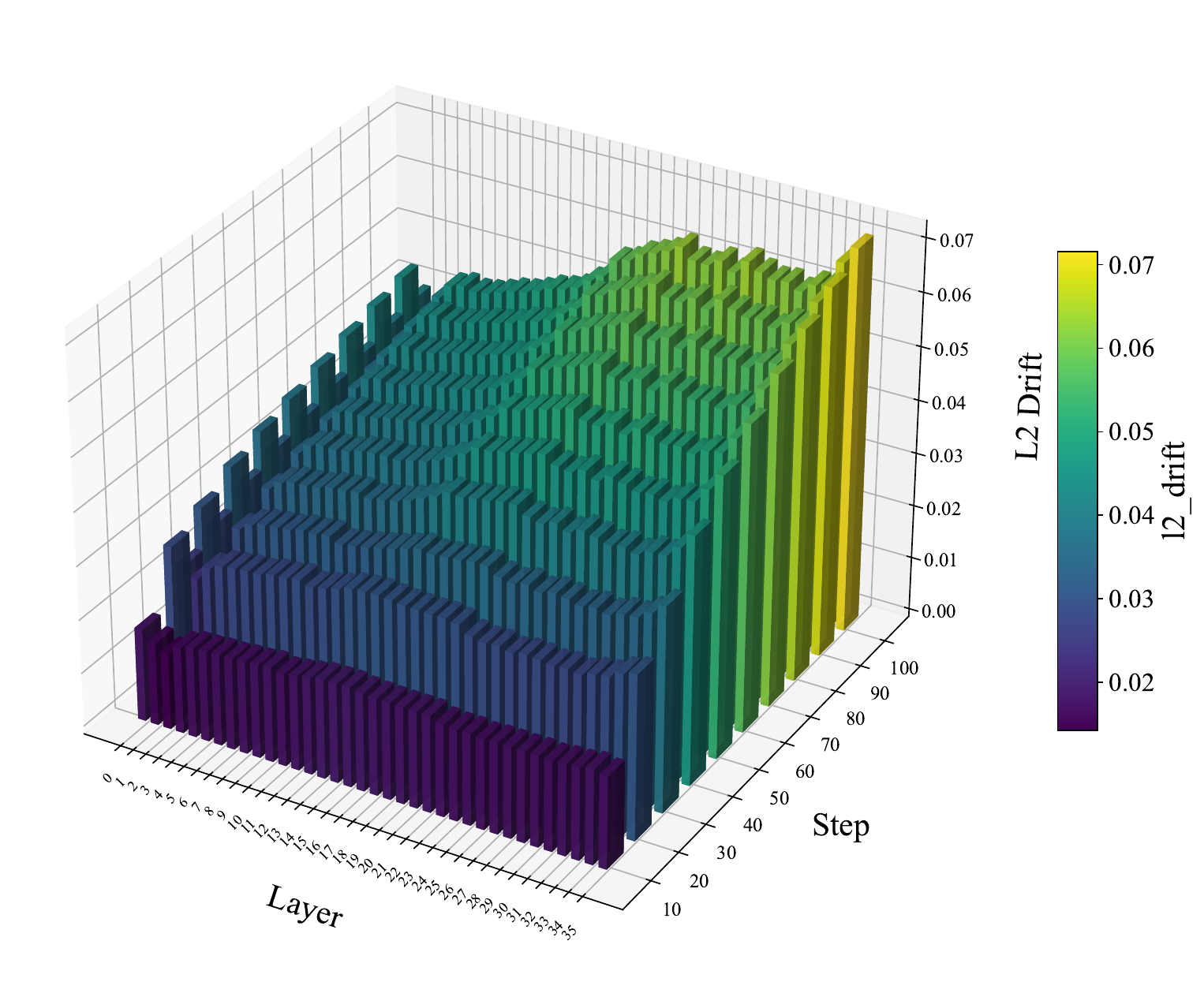}\\
        \small (b) OPD-only
    \end{minipage}\hfill
    \begin{minipage}[t]{0.24\textwidth}
        \centering
        \includegraphics[width=\linewidth]{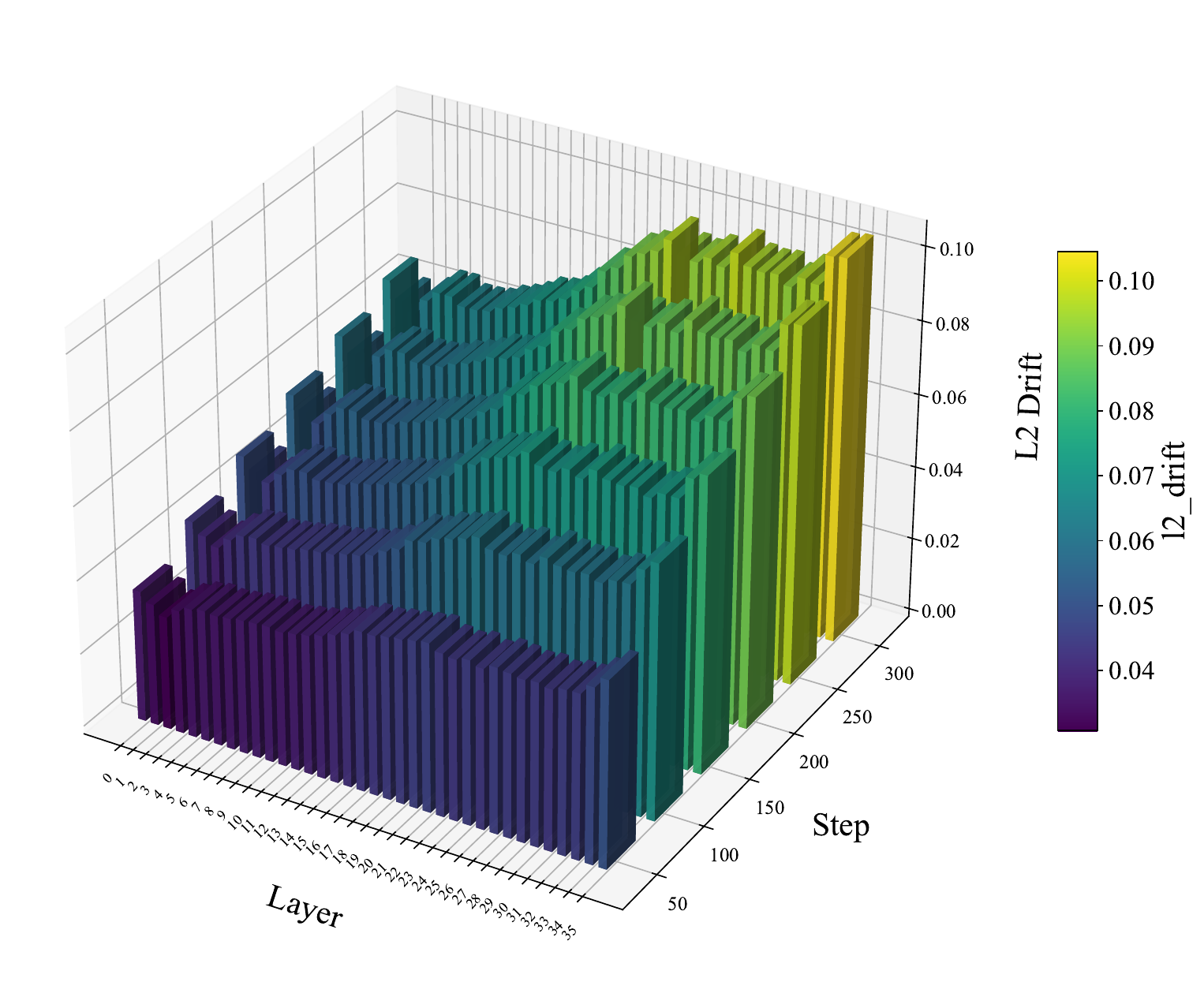}\\
        \small (c) Fixed GRPO+OPD
    \end{minipage}\hfill
    \begin{minipage}[t]{0.24\textwidth}
        \centering
        \includegraphics[width=\linewidth]{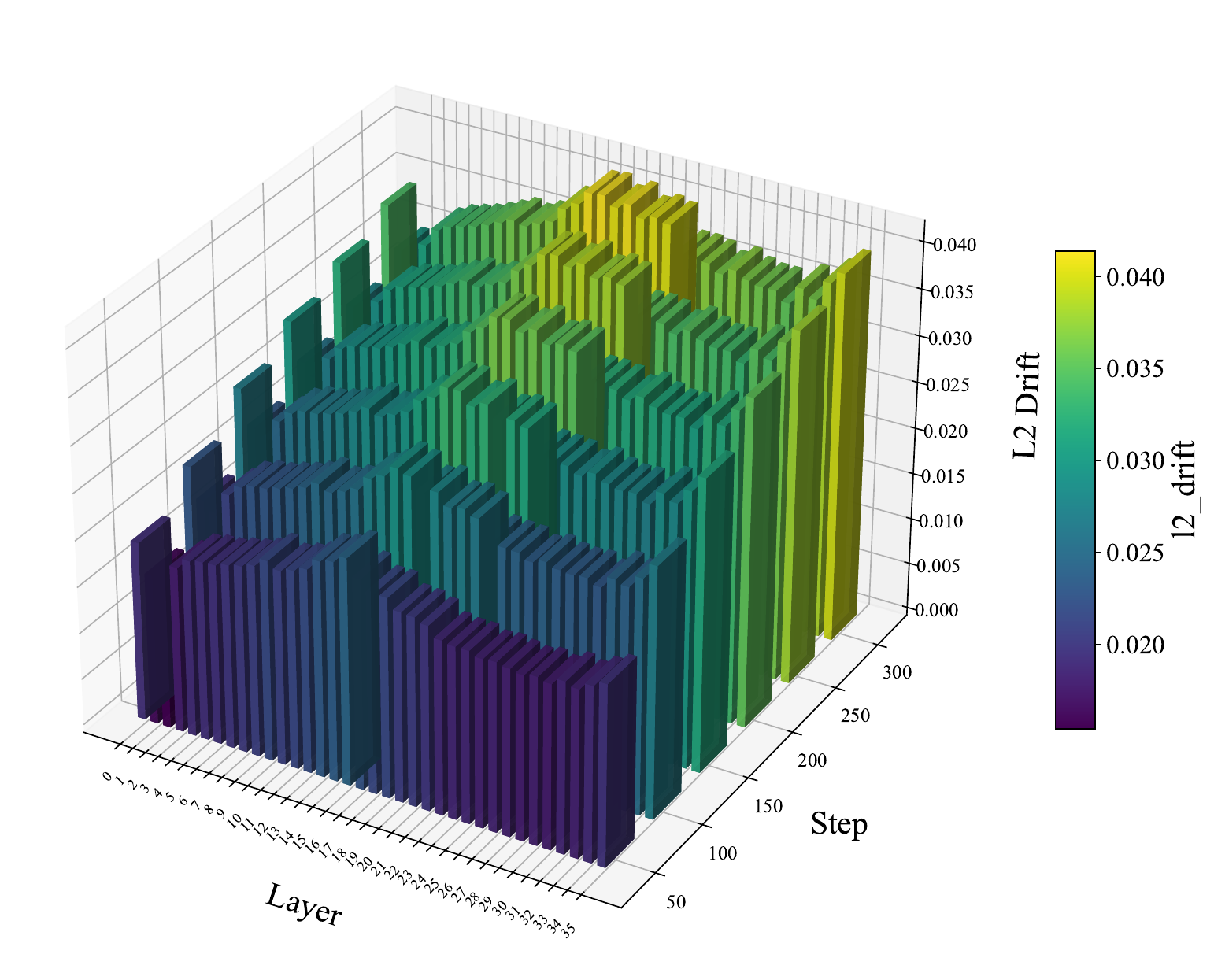}\\
        \small (d) SAF (ours)
    \end{minipage}
    \caption{Layer-wise L2 weight drift $\|\Delta W^{(t)}_\ell\|_2$ over transformer layers (x-axis) and optimization steps (y-axis) for the four training regimes of Qwen3-4B on mathematical reasoning: (a) GRPO-only, (b) OPD-only, (c) fixed-coefficient GRPO+OPD, and (d) SAF.}
    \label{fig:weight-drift-3d}
\end{figure*}

\begin{figure*}[!t]
    \centering
    \begin{minipage}[t]{0.24\textwidth}
        \centering
        \includegraphics[width=\linewidth]{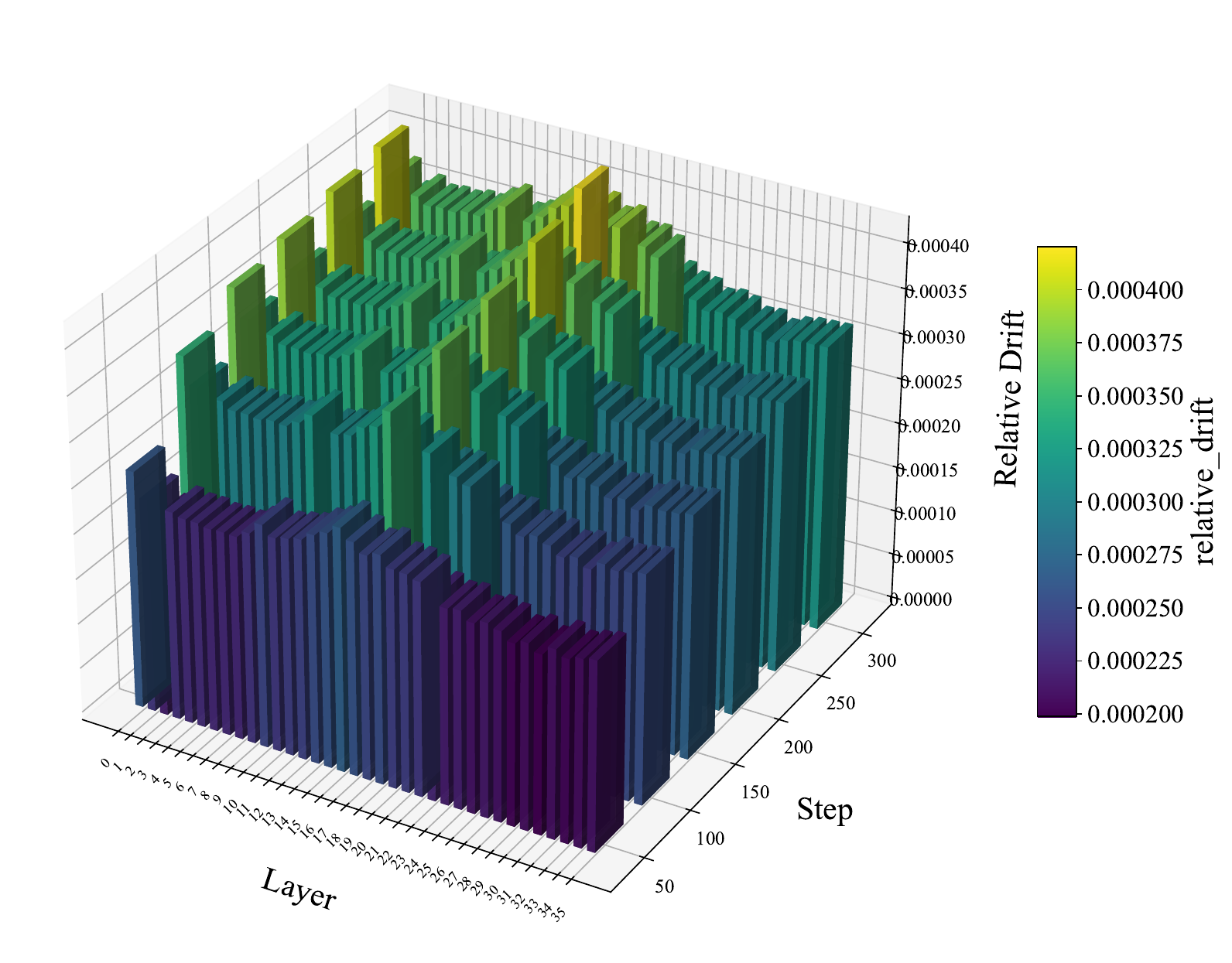}\\
        \small (a) GRPO-only
    \end{minipage}\hfill
    \begin{minipage}[t]{0.24\textwidth}
        \centering
        \includegraphics[width=\linewidth]{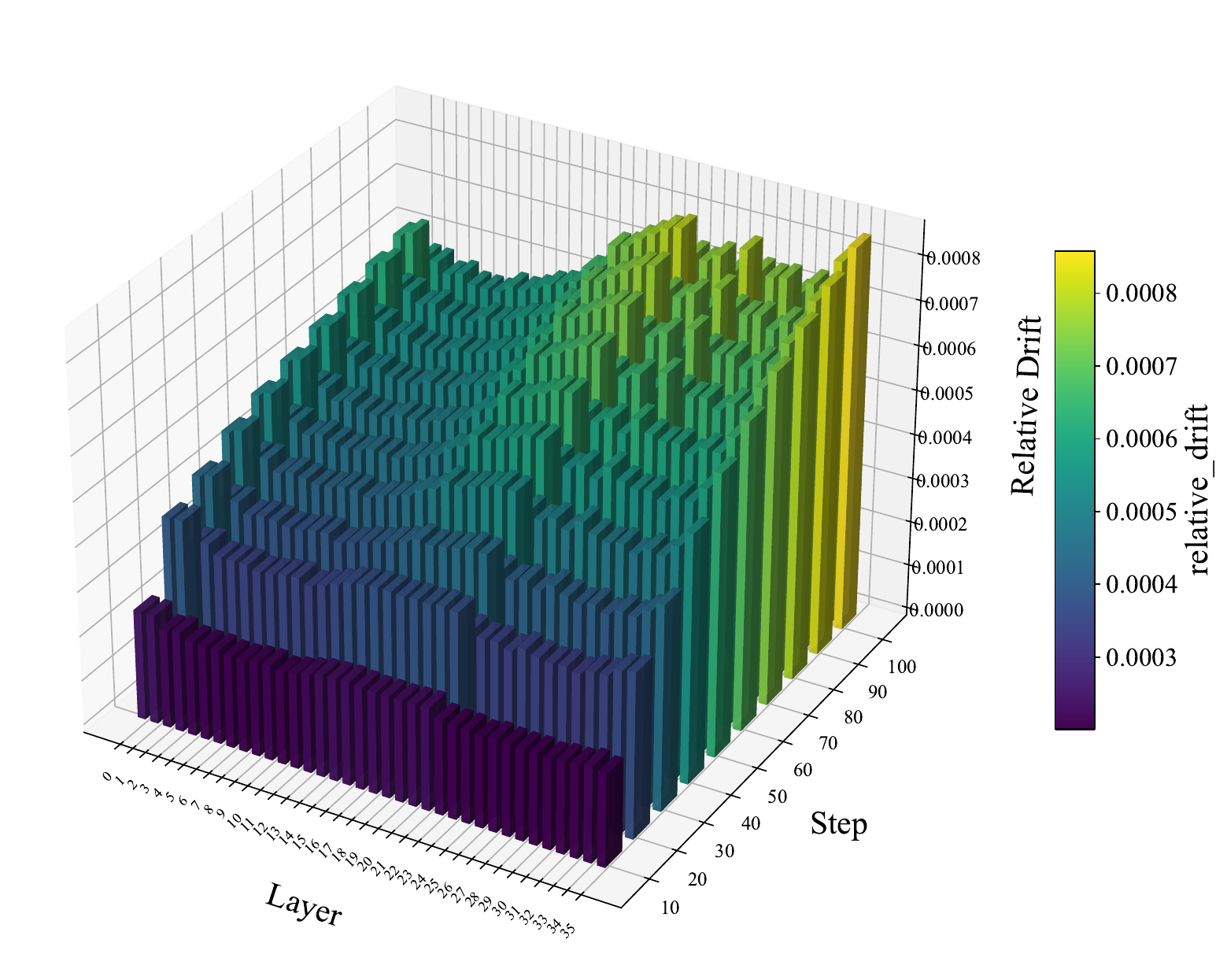}\\
        \small (b) OPD-only
    \end{minipage}\hfill
    \begin{minipage}[t]{0.24\textwidth}
        \centering
        \includegraphics[width=\linewidth]{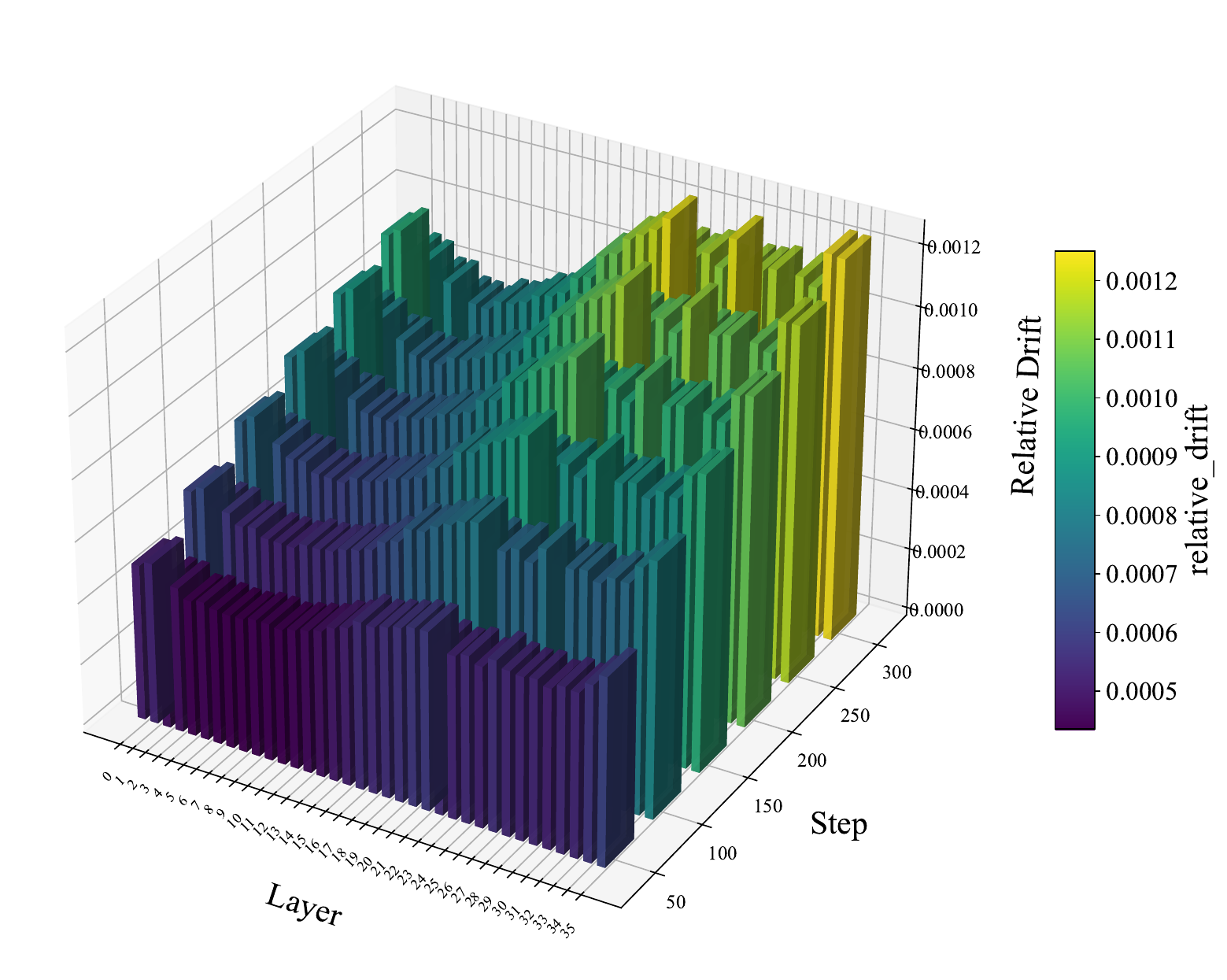}\\
        \small (c) Fixed GRPO+OPD
    \end{minipage}\hfill
    \begin{minipage}[t]{0.24\textwidth}
        \centering
        \includegraphics[width=\linewidth]{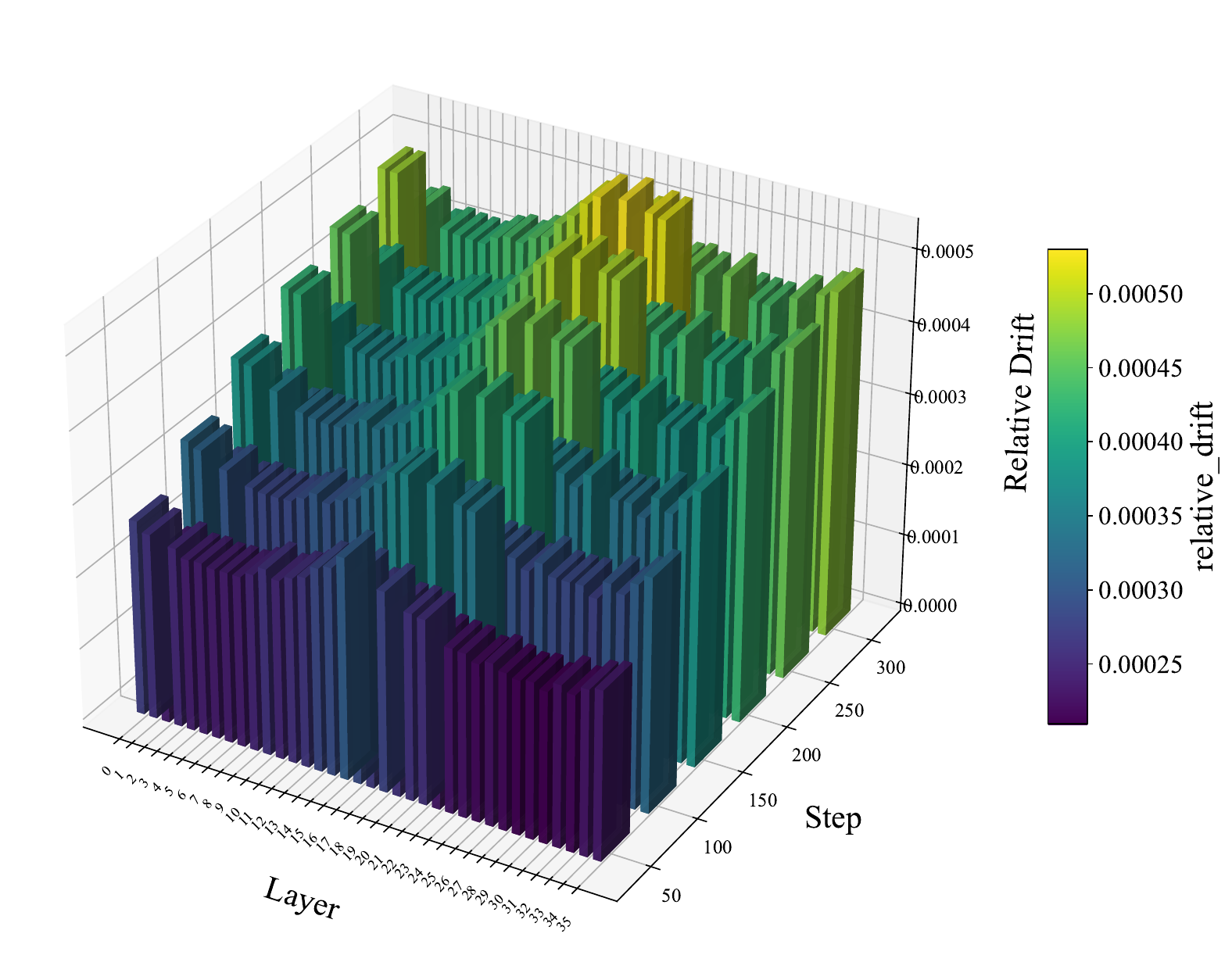}\\
        \small (d) SAF (ours)
    \end{minipage}
    \caption{Layer-wise relative weight drift $\|\Delta W^{(t)}_\ell\|_2 / (\|W^{0}_\ell\|_2+\epsilon)$ over transformer layers (x-axis) and optimization steps (y-axis) for the same four training regimes and checkpoints as Figure~\ref{fig:weight-drift-3d}: (a) GRPO-only, (b) OPD-only, (c) fixed-coefficient GRPO+OPD, and (d) SAF.}
    \label{fig:weight-drift-relative-3d}
\end{figure*}

\emph{Weight drift.} Writing $\theta^{(t)}_\ell$ for the flattened parameter vector of layer $\ell$ at step $t$ and $\theta^{0}_\ell$ for its base-model counterpart, the absolute L2 drift reported in Figure~\ref{fig:weight-drift-3d} is
\begin{equation}
\|\Delta W^{(t)}_\ell\|_2 = \bigl\| \theta^{(t)}_\ell - \theta^{0}_\ell \bigr\|_2,
\end{equation}
and the scale-normalized relative drift reported in Figure~\ref{fig:weight-drift-relative-3d} is
\begin{equation}
\frac{\|\Delta W^{(t)}_\ell\|_2}{\|W^{0}_\ell\|_2 + \epsilon} = \frac{\bigl\| \theta^{(t)}_\ell - \theta^{0}_\ell \bigr\|_2}{\bigl\| \theta^{0}_\ell \bigr\|_2 + \epsilon},
\end{equation}
with $\epsilon = 10^{-8}$ preventing division by zero. When a layer contains multiple named parameter tensors (e.g., separate \texttt{q/k/v/o\_proj} matrices), the layer-level value is the arithmetic mean of the metric over that layer's parameter tensors.

\paragraph{Layer-wise weight drift across all four training regimes (supplementary).} To further contextualize the update-matrix geometry above, Figure~\ref{fig:weight-drift-3d} visualizes the per-layer, per-step L2 weight drift $\|\Delta W^{(t)}_\ell\|_2$ for all four training regimes compared in the main text: GRPO-only, OPD-only, fixed-coefficient GRPO+OPD, and SAF, using the same Qwen3-4B mathematical-reasoning checkpoints as Figure~\ref{fig:training-dynamics}. Each surface plots drift magnitude against transformer layer index and optimization step; note that the z-axis and step-axis ranges differ across panels, since GRPO-only, fixed fusion, and SAF are shown over their full 300-step budget while OPD-only is shown over its own full 100-step training horizon. GRPO-only accumulates drift gradually and reaches its largest values only around step 300, with a magnitude (peak $\approx 0.03$) well below the other regimes. OPD-only, in contrast, drifts much faster per step, reaching a comparable or larger magnitude (peak $\approx 0.07$) within only 100 steps, consistent with the denser, token-level supervision it receives at every update. Fixed-coefficient GRPO+OPD compounds both effects: because it applies full-strength OPD guidance on top of the GRPO update at every step, its drift grows the fastest and reaches the largest magnitude of all four regimes (peak $\approx 0.10$) by step 300, concentrated in the later layers. SAF, which applies the same magnitude control and temporal annealing described in Section~\ref{sec:method}, keeps the drift substantially smaller (peak $\approx 0.04$) than fixed fusion despite training for the same number of steps, while still exceeding GRPO-only, indicating that SAF's controllers curb, without eliminating, the excess parameter movement induced by an unregulated distillation signal. This pattern is consistent with, and provides an additional parameter-space view of, the entropy and student--teacher KL trajectories already discussed in Figure~\ref{fig:training-dynamics}: the regime with the least-controlled OPD signal (fixed fusion) also drifts the most in parameter space, while SAF's controllers rein in this drift without fully suppressing it. As with Figure~\ref{fig:update-geometry}, we report this as a supplementary, exploratory observation rather than as a mechanism SAF's design was derived from.

\paragraph{Layer-wise relative weight drift across all four training regimes (supplementary).} The absolute L2 drift in Figure~\ref{fig:weight-drift-3d} is influenced by the parameter count of each layer, which makes cross-layer comparisons within a single regime less direct. As a complementary, scale-normalized view, Figure~\ref{fig:weight-drift-relative-3d} reports the relative weight drift $\|\Delta W^{(t)}_\ell\|_2 / (\|W^{0}_\ell\|_2+\epsilon)$ for the same four regimes and checkpoints. Normalizing by the base-parameter norm removes the confound of layer size and yields a picture consistent with, but sharper than, Figure~\ref{fig:weight-drift-3d}: GRPO-only stays the most conservative throughout training, with its layer-averaged relative drift rising from about $2.3\times10^{-4}$ at step 50 to about $3.6\times10^{-4}$ at step 300 (a $1.6\times$ increase). OPD-only again shows the fastest per-step growth, moving from about $2.2\times10^{-4}$ at step 10 to about $6.8\times10^{-4}$ at step 100 (a $3.2\times$ increase in one third of the steps), with its largest relative drift concentrated in the deepest layers (e.g., layer 35). Fixed-coefficient GRPO+OPD attains both the fastest growth rate and the largest final magnitude, reaching a layer-averaged relative drift of about $1.0\times10^{-3}$ by step 300 (a $4.8\times$ increase from step 10), again concentrated in the later layers. SAF closely tracks the same qualitative shape but is uniformly compressed relative to fixed fusion, reaching only about $4.6\times10^{-4}$ by step 300 despite a comparable $4.1\times$ relative increase from its own step-10 starting point, confirming on a scale-normalized basis that SAF's controllers curb the magnitude of parameter movement rather than merely its layer distribution.

\section{Student--Teacher KL Computation}
\label{app:student-teacher-kl}

The temporal controller uses a sampled-token estimate of the reverse KL divergence from the student policy $\pi_{\theta}$ to the teacher policy $\pi_T$. The estimate is computed on the same student-generated responses used for policy optimization, avoiding a full-vocabulary KL evaluation at every response position. Let $s_{i,t}=(\boldsymbol{x}_i,\boldsymbol{y}_{i,<t})$ denote the context at token $t$ of response $i$. For the sampled token $y_{i,t}$, we first compute the clipped teacher-to-student log-probability ratio
\begin{equation}
\begin{aligned}
d_{i,t}=\operatorname{clip}\!\bigl(&
\log \pi_T(y_{i,t}\mid s_{i,t})
-\log \pi_{\theta}(y_{i,t}\mid s_{i,t}),\\
&-20,20\bigr).
\end{aligned}
\end{equation}
Following the low-variance $k_3$ estimator, we then define the token-level contribution as
\begin{equation}
 \widehat{D}_{i,t}
 = \operatorname{clip}\!\left(
 \exp(d_{i,t})-d_{i,t}-1,
 -10,10
 \right).
 \label{eq:actor-kl-token}
\end{equation}
Without clipping, Eq.~\ref{eq:actor-kl-token} reduces to the non-negative estimator $r-\log r-1$, where $r=\pi_T(y_{i,t}\mid s_{i,t})/\pi_{\theta}(y_{i,t}\mid s_{i,t})$. Its expectation under $y_{i,t}\sim\pi_{\theta}(\cdot\mid s_{i,t})$ equals $D_{\mathrm{KL}}\!\left(\pi_{\theta}(\cdot\mid s_{i,t})\,\|\,\pi_T(\cdot\mid s_{i,t})\right)$. The clipping operations are implementation safeguards against numerical overflow and extreme finite-sample contributions.

Let $m_{i,t}\in\{0,1\}$ be the response mask, which excludes prompt and padding positions. We aggregate valid token contributions using a token mean:
\begin{equation}
 KL_s
 = \frac{\sum_{i,t}m_{i,t}\widehat{D}_{i,t}}
        {\sum_{i,t}m_{i,t}}.
 \label{eq:actor-kl-loss}
\end{equation}
This scalar is logged as \texttt{actor/kl\_loss}. In our configuration, \texttt{use\_kl\_loss=True} enables its computation, whereas \texttt{kl\_loss\_coef=0} prevents it from contributing an additional KL regularization term to the actor objective. It therefore serves only as an observable student--teacher alignment statistic. The warm-up controller records its initial value as $KL_0$ and applies the relative decrease $(KL_0-KL_s)/KL_0$ in the early-stop criterion of Section~\ref{sec:temporal-control}.

%

\section{token-level Raw Signal Analysis}
\label{app:raw-signal}

This appendix inspects the two unprocessed training advantages over the first five consecutive training steps. We use the notation from Section~\ref{sec:preliminary}: $A_{i,t}^{\mathrm{OPD}}=\log\pi_T(y_{i,t}\mid s_{i,t})-\log\pi_{\theta}(y_{i,t}\mid s_{i,t})$ denotes the token-level OPD advantage, and $A_i^{\mathrm{GRPO}}$ denotes the response-level GRPO advantage shared by all tokens in response $i$. These quantities correspond to the implementation fields \texttt{raw\_opd} and \texttt{raw\_grpo}, respectively.

For each training step, we select three representative responses: \textbf{(a)} the response with the most negative $A_i^{\mathrm{GRPO}}$ (incorrect), \textbf{(b)} a response with near-zero $A_i^{\mathrm{GRPO}}$ (neutral/mixed), and \textbf{(c)} the response with the most positive $A_i^{\mathrm{GRPO}}$ (correct). For each response, we report the 12 positions with the largest $|A_{i,t}^{\mathrm{OPD}}|$, the first 15 response tokens for local context, and summary statistics over all response tokens. In each response block, the two token-level tables place token positions along the horizontal axis, followed by a compact summary table.

\bigskip

\subsection{Training Step~1}

\textbf{(a) Response 189 ($A_i^{\mathrm{GRPO}} = -2.4749$, incorrect, 3148 tokens)}

\smallskip\noindent
\noindent\begin{minipage}[t]{\linewidth}
\centering
\textbf{Top 12 by $|A_{i,t}^{\mathrm{OPD}}|$ (ordered by position):}

\vspace{2pt}
{\scriptsize
\setlength{\tabcolsep}{2pt}
\resizebox{\linewidth}{!}{%
\begin{tabular}{l*{12}{c}}
\toprule
Position & 189 & 346 & 540 & 641 & 847 & 946 & 994 & 1383 & 1847 & 3091 & 3092 & 3134 \\
Token & \texttt{ several} & \texttt{2} & \texttt{ will} & \texttt{ term} & \texttt{ E} & \texttt{7} & \texttt{ cancel} & \texttt{\}} & \texttt{\textbackslash{}} & \texttt{Now} & \texttt{ take} & \texttt{---\nl\nl} \\
$A_{i,t}^{\mathrm{OPD}}$ & -13.7876 & -17.4881 & -10.5098 & -13.9257 & -14.4314 & -9.0760 & -8.5954 & -10.7748 & -9.3498 & -10.9243 & -10.2098 & -8.9048 \\
\bottomrule
\end{tabular}}
}
\end{minipage}

\smallskip
\noindent\begin{minipage}[t]{\linewidth}
\centering
\textbf{First 15 response tokens (with context):}

\vspace{2pt}
{\scriptsize
\setlength{\tabcolsep}{2pt}
\resizebox{\linewidth}{!}{%
\begin{tabular}{l*{15}{c}}
\toprule
Position & 0 & 1 & 2 & 3 & 4 & 5 & 6 & 7 & 8 & 9 & 10 & 11 & 12 & 13 & 14 \\
Token & \texttt{We} & \texttt{ are} & \texttt{ given} & \texttt{ the} & \texttt{ infinite} & \texttt{ product} & \texttt{:\nl\nl} & \texttt{\$\$} & \texttt{\nl} & \texttt{\textbackslash{}} & \texttt{prod} & \texttt{\_\{} & \texttt{n} & \texttt{=} & \texttt{2} \\
$A_{i,t}^{\mathrm{OPD}}$ & +0.0000 & +0.0004 & -4.6245 & +0.0068 & +0.0007 & +0.0000 & +0.0005 & +0.0000 & +0.0000 & +0.0019 & +0.0000 & +0.0000 & +0.0000 & +0.0000 & +0.0000 \\
\bottomrule
\end{tabular}}
}
\end{minipage}

\smallskip
\noindent\begin{minipage}[t]{\linewidth}
\centering
\textbf{Summary (3148 tokens):}

\vspace{2pt}
{\scriptsize
\setlength{\tabcolsep}{3pt}
\begin{tabular}{lrr}
\toprule
Stat. & $A_{i,t}^{\mathrm{OPD}}$ & $A_i^{\mathrm{GRPO}}$ \\
\midrule
mean & -0.2594 & -2.4749 \\
std  & +1.2329 &  0.0000 \\
min  & -17.4881 & -2.4749 \\
max  & +5.6796 & -2.4749 \\
\bottomrule
\end{tabular}}
\end{minipage}

\bigskip

\textbf{(b) Response 0 ($A_i^{\mathrm{GRPO}} = +0.3536$, neutral, 363 tokens)}

\smallskip\noindent
\noindent\begin{minipage}[t]{\linewidth}
\centering
\textbf{Top 12 by $|A_{i,t}^{\mathrm{OPD}}|$ (ordered by position):}

\vspace{2pt}
{\scriptsize
\setlength{\tabcolsep}{2pt}
\resizebox{\linewidth}{!}{%
\begin{tabular}{l*{12}{c}}
\toprule
Position & 66 & 92 & 94 & 107 & 110 & 120 & 132 & 136 & 177 & 258 & 347 & 362 \\
Token & \texttt{ find} & \texttt{This} & \texttt{ an} & \texttt{ exponential} & \texttt{ **} & \texttt{ \$} & \texttt{ over} & \texttt{ real} & \texttt{ variables} & \texttt{ our} & \texttt{ Final} & \texttt{\textless{}\textbar{}im\_end\textbar{}\textgreater{}} \\
$A_{i,t}^{\mathrm{OPD}}$ & -8.4885 & -6.3400 & -6.1097 & -5.9964 & -8.8939 & -8.1489 & -6.5749 & -10.1239 & -7.3847 & -5.4155 & -9.6873 & -5.2518 \\
\bottomrule
\end{tabular}}
}
\end{minipage}

\smallskip
\noindent\begin{minipage}[t]{\linewidth}
\centering
\textbf{First 15 response tokens (with context):}

\vspace{2pt}
{\scriptsize
\setlength{\tabcolsep}{2pt}
\resizebox{\linewidth}{!}{%
\begin{tabular}{l*{15}{c}}
\toprule
Position & 0 & 1 & 2 & 3 & 4 & 5 & 6 & 7 & 8 & 9 & 10 & 11 & 12 & 13 & 14 \\
Token & \texttt{We} & \texttt{ are} & \texttt{ given} & \texttt{ a} & \texttt{ joint} & \texttt{ density} & \texttt{ function} & \texttt{:\nl\nl} & \texttt{\$\$} & \texttt{\nl} & \texttt{f} & \texttt{(x} & \texttt{,} & \texttt{ y} & \texttt{,} \\
$A_{i,t}^{\mathrm{OPD}}$ & +0.0000 & +0.0001 & +0.0139 & -0.0141 & +0.1487 & -1.6046 & +0.0000 & -2.2116 & +0.0000 & +0.0000 & +0.0000 & +0.0001 & -1.1151 & +0.0000 & +0.0000 \\
\bottomrule
\end{tabular}}
}
\end{minipage}

\smallskip
\noindent\begin{minipage}[t]{\linewidth}
\centering
\textbf{Summary (363 tokens):}

\vspace{2pt}
{\scriptsize
\setlength{\tabcolsep}{3pt}
\begin{tabular}{lrr}
\toprule
Stat. & $A_{i,t}^{\mathrm{OPD}}$ & $A_i^{\mathrm{GRPO}}$ \\
\midrule
mean & -0.5425 & +0.3536 \\
std  & +1.6162 &  0.0000 \\
min  & -10.1239 & +0.3536 \\
max  & +1.8588 & +0.3536 \\
\bottomrule
\end{tabular}}
\end{minipage}

\bigskip

\textbf{(c) Response 135 ($A_i^{\mathrm{GRPO}} = +2.4749$, correct, 695 tokens)}

\smallskip\noindent
\noindent\begin{minipage}[t]{\linewidth}
\centering
\textbf{Top 12 by $|A_{i,t}^{\mathrm{OPD}}|$ (ordered by position):}

\vspace{2pt}
{\scriptsize
\setlength{\tabcolsep}{2pt}
\resizebox{\linewidth}{!}{%
\begin{tabular}{l*{12}{c}}
\toprule
Position & 23 & 89 & 183 & 219 & 224 & 247 & 248 & 259 & 274 & 327 & 356 & 636 \\
Token & \texttt{ **} & \texttt{ function} & \texttt{ equal} & \texttt{ \}} & \texttt{ since} & \texttt{ **} & \texttt{zero} & \texttt{(g} & \texttt{ permutation} & \texttt{ Half} & \texttt{number} & \texttt{ Final} \\
$A_{i,t}^{\mathrm{OPD}}$ & -7.6053 & -8.7120 & -10.3569 & -11.7299 & -6.7478 & -8.6477 & -11.7095 & -8.4914 & -8.5523 & -7.0024 & -14.7974 & -9.8838 \\
\bottomrule
\end{tabular}}
}
\end{minipage}

\smallskip
\noindent\begin{minipage}[t]{\linewidth}
\centering
\textbf{First 15 response tokens (with context):}

\vspace{2pt}
{\scriptsize
\setlength{\tabcolsep}{2pt}
\resizebox{\linewidth}{!}{%
\begin{tabular}{l*{15}{c}}
\toprule
Position & 0 & 1 & 2 & 3 & 4 & 5 & 6 & 7 & 8 & 9 & 10 & 11 & 12 & 13 & 14 \\
Token & \texttt{We} & \texttt{ are} & \texttt{ asked} & \texttt{ to} & \texttt{ determine} & \texttt{ whether} & \texttt{ the} & \texttt{ following} & \texttt{ statement} & \texttt{ is} & \texttt{ **} & \texttt{true} & \texttt{ for} & \texttt{ all} & \texttt{ \$} \\
$A_{i,t}^{\mathrm{OPD}}$ & +0.0001 & +0.0002 & +0.4287 & +0.0065 & +0.0016 & +0.0023 & +0.0028 & +0.3882 & +0.0249 & +0.0000 & -1.9667 & +0.0000 & +0.0067 & +0.0000 & -0.6986 \\
\bottomrule
\end{tabular}}
}
\end{minipage}

\smallskip
\noindent\begin{minipage}[t]{\linewidth}
\centering
\textbf{Summary (695 tokens):}

\vspace{2pt}
{\scriptsize
\setlength{\tabcolsep}{3pt}
\begin{tabular}{lrr}
\toprule
Stat. & $A_{i,t}^{\mathrm{OPD}}$ & $A_i^{\mathrm{GRPO}}$ \\
\midrule
mean & -0.3897 & +2.4749 \\
std  & +1.5458 &  0.0000 \\
min  & -14.7974 & +2.4749 \\
max  & +1.9325 & +2.4749 \\
\bottomrule
\end{tabular}}
\end{minipage}


\subsection{Training Step~2}

\textbf{(a) Response 8 ($A_i^{\mathrm{GRPO}} = -2.4749$, incorrect, 733 tokens)}

\smallskip\noindent
\noindent\begin{minipage}[t]{\linewidth}
\centering
\textbf{Top 12 by $|A_{i,t}^{\mathrm{OPD}}|$ (ordered by position):}

\vspace{2pt}
{\scriptsize
\setlength{\tabcolsep}{2pt}
\resizebox{\linewidth}{!}{%
\begin{tabular}{l*{12}{c}}
\toprule
Position & 68 & 118 & 137 & 160 & 249 & 287 & 289 & 498 & 554 & 571 & 719 & 721 \\
Token & \texttt{ term} & \texttt{(} & \texttt{\}\nl} & \texttt{ \textbackslash{}} & \texttt{1} & \texttt{ } & \texttt{ +} & \texttt{ } & \texttt{ Therefore} & \texttt{Then} & \texttt{---\nl\nl} & \texttt{ Final} \\
$A_{i,t}^{\mathrm{OPD}}$ & -7.2954 & -9.7388 & -7.1736 & -11.2268 & -17.9447 & -16.4380 & -10.4996 & -15.2381 & -8.4721 & -8.6563 & -9.9507 & -10.2704 \\
\bottomrule
\end{tabular}}
}
\end{minipage}

\smallskip
\noindent\begin{minipage}[t]{\linewidth}
\centering
\textbf{First 15 response tokens (with context):}

\vspace{2pt}
{\scriptsize
\setlength{\tabcolsep}{2pt}
\resizebox{\linewidth}{!}{%
\begin{tabular}{l*{15}{c}}
\toprule
Position & 0 & 1 & 2 & 3 & 4 & 5 & 6 & 7 & 8 & 9 & 10 & 11 & 12 & 13 & 14 \\
Token & \texttt{We} & \texttt{ are} & \texttt{ asked} & \texttt{ to} & \texttt{ evaluate} & \texttt{ the} & \texttt{ limit} & \texttt{:\nl\nl} & \texttt{\$\$} & \texttt{\nl} & \texttt{\textbackslash{}} & \texttt{lim} & \texttt{\_\{} & \texttt{n} & \texttt{ \textbackslash{}} \\
$A_{i,t}^{\mathrm{OPD}}$ & +0.0000 & +0.0008 & +0.1321 & +0.0000 & +0.0027 & +0.0053 & +0.0085 & +0.0053 & +0.0000 & +0.0000 & +0.0000 & +0.0000 & +0.0000 & +0.0000 & -3.3742 \\
\bottomrule
\end{tabular}}
}
\end{minipage}

\smallskip
\noindent\begin{minipage}[t]{\linewidth}
\centering
\textbf{Summary (733 tokens):}

\vspace{2pt}
{\scriptsize
\setlength{\tabcolsep}{3pt}
\begin{tabular}{lrr}
\toprule
Stat. & $A_{i,t}^{\mathrm{OPD}}$ & $A_i^{\mathrm{GRPO}}$ \\
\midrule
mean & -0.4116 & -2.4749 \\
std  & +1.7312 &  0.0000 \\
min  & -17.9447 & -2.4749 \\
max  & +2.5161 & -2.4749 \\
\bottomrule
\end{tabular}}
\end{minipage}

\bigskip

\textbf{(b) Response 4 ($A_i^{\mathrm{GRPO}} = +0.3536$, neutral, 602 tokens)}

\smallskip\noindent
\noindent\begin{minipage}[t]{\linewidth}
\centering
\textbf{Top 12 by $|A_{i,t}^{\mathrm{OPD}}|$ (ordered by position):}

\vspace{2pt}
{\scriptsize
\setlength{\tabcolsep}{2pt}
\resizebox{\linewidth}{!}{%
\begin{tabular}{l*{12}{c}}
\toprule
Position & 318 & 334 & 359 & 379 & 432 & 447 & 477 & 497 & 571 & 574 & 587 & 590 \\
Token & \texttt{ \$\nl\nl} & \texttt{1} & \texttt{ looks} & \texttt{ +} & \texttt{approx} & \texttt{Thus} & \texttt{Therefore} & \texttt{ \$\nl\nl} & \texttt{Therefore} & \texttt{ final} & \texttt{---} & \texttt{ Final} \\
$A_{i,t}^{\mathrm{OPD}}$ & -10.7052 & -6.9898 & -7.0174 & -9.4997 & -7.7371 & -12.4224 & -11.3634 & -7.8916 & -10.2927 & -9.2755 & -7.8828 & -8.2090 \\
\bottomrule
\end{tabular}}
}
\end{minipage}

\smallskip
\noindent\begin{minipage}[t]{\linewidth}
\centering
\textbf{First 15 response tokens (with context):}

\vspace{2pt}
{\scriptsize
\setlength{\tabcolsep}{2pt}
\resizebox{\linewidth}{!}{%
\begin{tabular}{l*{15}{c}}
\toprule
Position & 0 & 1 & 2 & 3 & 4 & 5 & 6 & 7 & 8 & 9 & 10 & 11 & 12 & 13 & 14 \\
Token & \texttt{We} & \texttt{ are} & \texttt{ asked} & \texttt{ to} & \texttt{ evaluate} & \texttt{ the} & \texttt{ limit} & \texttt{:\nl\nl} & \texttt{\$\$} & \texttt{\nl} & \texttt{\textbackslash{}} & \texttt{lim} & \texttt{\_\{} & \texttt{n} & \texttt{ \textbackslash{}} \\
$A_{i,t}^{\mathrm{OPD}}$ & +0.0000 & +0.0008 & +0.1333 & +0.0000 & +0.0027 & +0.0052 & +0.0150 & +0.0053 & +0.0000 & +0.0000 & +0.0000 & +0.0000 & +0.0000 & +0.0000 & -4.1184 \\
\bottomrule
\end{tabular}}
}
\end{minipage}

\smallskip
\noindent\begin{minipage}[t]{\linewidth}
\centering
\textbf{Summary (602 tokens):}

\vspace{2pt}
{\scriptsize
\setlength{\tabcolsep}{3pt}
\begin{tabular}{lrr}
\toprule
Stat. & $A_{i,t}^{\mathrm{OPD}}$ & $A_i^{\mathrm{GRPO}}$ \\
\midrule
mean & -0.4214 & +0.3536 \\
std  & +1.6275 &  0.0000 \\
min  & -12.4224 & +0.3536 \\
max  & +2.0158 & +0.3536 \\
\bottomrule
\end{tabular}}
\end{minipage}

\bigskip

\textbf{(c) Response 37 ($A_i^{\mathrm{GRPO}} = +2.4749$, correct, 1383 tokens)}

\smallskip\noindent
\noindent\begin{minipage}[t]{\linewidth}
\centering
\textbf{Top 12 by $|A_{i,t}^{\mathrm{OPD}}|$ (ordered by position):}

\vspace{2pt}
{\scriptsize
\setlength{\tabcolsep}{2pt}
\resizebox{\linewidth}{!}{%
\begin{tabular}{l*{12}{c}}
\toprule
Position & 148 & 154 & 205 & 329 & 336 & 528 & 530 & 650 & 713 & 749 & 814 & 991 \\
Token & \texttt{ space} & \texttt{ can} & \texttt{F} & \texttt{ measurable} & \texttt{ also} & \texttt{ can} & \texttt{ associated} & \texttt{ However} & \texttt{This} & \texttt{ power} & \texttt{ can} & \texttt{Since} \\
$A_{i,t}^{\mathrm{OPD}}$ & -8.0621 & -7.8963 & -10.5909 & -20.3585 & -13.4059 & -12.1148 & -8.4644 & -9.0888 & -7.9808 & -10.7333 & -11.8686 & -12.1417 \\
\bottomrule
\end{tabular}}
}
\end{minipage}

\smallskip
\noindent\begin{minipage}[t]{\linewidth}
\centering
\textbf{First 15 response tokens (with context):}

\vspace{2pt}
{\scriptsize
\setlength{\tabcolsep}{2pt}
\resizebox{\linewidth}{!}{%
\begin{tabular}{l*{15}{c}}
\toprule
Position & 0 & 1 & 2 & 3 & 4 & 5 & 6 & 7 & 8 & 9 & 10 & 11 & 12 & 13 & 14 \\
Token & \texttt{We} & \texttt{ are} & \texttt{ asked} & \texttt{ to} & \texttt{ find} & \texttt{ the} & \texttt{ **} & \texttt{card} & \texttt{inality} & \texttt{**} & \texttt{ of} & \texttt{ the} & \texttt{ set} & \texttt{ of} & \texttt{ **} \\
$A_{i,t}^{\mathrm{OPD}}$ & +0.0015 & +0.0082 & +0.0010 & +0.0000 & +0.0779 & +0.0001 & +0.0004 & -0.0000 & +0.0000 & +0.0007 & -0.0001 & +0.0000 & +0.0019 & +0.0000 & -0.0332 \\
\bottomrule
\end{tabular}}
}
\end{minipage}

\smallskip
\noindent\begin{minipage}[t]{\linewidth}
\centering
\textbf{Summary (1383 tokens):}

\vspace{2pt}
{\scriptsize
\setlength{\tabcolsep}{3pt}
\begin{tabular}{lrr}
\toprule
Stat. & $A_{i,t}^{\mathrm{OPD}}$ & $A_i^{\mathrm{GRPO}}$ \\
\midrule
mean & -0.4614 & +2.4749 \\
std  & +1.5323 &  0.0000 \\
min  & -20.3585 & +2.4749 \\
max  & +2.5852 & +2.4749 \\
\bottomrule
\end{tabular}}
\end{minipage}


\subsection{Training Step~3}

\textbf{(a) Response 152 ($A_i^{\mathrm{GRPO}} = -2.4749$, incorrect, 1033 tokens)}

\smallskip\noindent
\noindent\begin{minipage}[t]{\linewidth}
\centering
\textbf{Top 12 by $|A_{i,t}^{\mathrm{OPD}}|$ (ordered by position):}

\vspace{2pt}
{\scriptsize
\setlength{\tabcolsep}{2pt}
\resizebox{\linewidth}{!}{%
\begin{tabular}{l*{12}{c}}
\toprule
Position & 42 & 231 & 318 & 416 & 450 & 457 & 546 & 559 & 881 & 940 & 946 & 985 \\
Token & \texttt{(} & \texttt{+} & \texttt{ \textbackslash{}} & \texttt{ very} & \texttt{approx} & \texttt{So} & \texttt{left} & \texttt{\nl} & \texttt{ Take} & \texttt{ \textless{}} & \texttt{So} & \texttt{ something} \\
$A_{i,t}^{\mathrm{OPD}}$ & -12.3997 & -6.8740 & -7.9706 & -8.4695 & -10.7473 & -8.2718 & -7.1454 & -7.7500 & -7.3970 & -7.7376 & -12.9474 & -9.0241 \\
\bottomrule
\end{tabular}}
}
\end{minipage}

\smallskip
\noindent\begin{minipage}[t]{\linewidth}
\centering
\textbf{First 15 response tokens (with context):}

\vspace{2pt}
{\scriptsize
\setlength{\tabcolsep}{2pt}
\resizebox{\linewidth}{!}{%
\begin{tabular}{l*{15}{c}}
\toprule
Position & 0 & 1 & 2 & 3 & 4 & 5 & 6 & 7 & 8 & 9 & 10 & 11 & 12 & 13 & 14 \\
Token & \texttt{We} & \texttt{ are} & \texttt{ given} & \texttt{ a} & \texttt{ recursive} & \texttt{ sequence} & \texttt{ defined} & \texttt{ by} & \texttt{:\nl\nl} & \texttt{\$\$} & \texttt{\nl} & \texttt{b} & \texttt{\_} & \texttt{1} & \texttt{ =} \\
$A_{i,t}^{\mathrm{OPD}}$ & +0.0000 & +0.0002 & +0.0028 & +0.0797 & +0.2692 & +0.0007 & -2.8324 & +0.2013 & +0.0022 & +0.0273 & +0.0000 & +0.0000 & +0.0000 & +0.0000 & +0.0000 \\
\bottomrule
\end{tabular}}
}
\end{minipage}

\smallskip
\noindent\begin{minipage}[t]{\linewidth}
\centering
\textbf{Summary (1033 tokens):}

\vspace{2pt}
{\scriptsize
\setlength{\tabcolsep}{3pt}
\begin{tabular}{lrr}
\toprule
Stat. & $A_{i,t}^{\mathrm{OPD}}$ & $A_i^{\mathrm{GRPO}}$ \\
\midrule
mean & -0.2942 & -2.4749 \\
std  & +1.2957 &  0.0000 \\
min  & -12.9474 & -2.4749 \\
max  & +2.9957 & -2.4749 \\
\bottomrule
\end{tabular}}
\end{minipage}

\bigskip

\textbf{(b) Response 2 ($A_i^{\mathrm{GRPO}} = -0.3536$, neutral, 552 tokens)}

\smallskip\noindent
\noindent\begin{minipage}[t]{\linewidth}
\centering
\textbf{Top 12 by $|A_{i,t}^{\mathrm{OPD}}|$ (ordered by position):}

\vspace{2pt}
{\scriptsize
\setlength{\tabcolsep}{2pt}
\resizebox{\linewidth}{!}{%
\begin{tabular}{l*{12}{c}}
\toprule
Position & 6 & 54 & 87 & 206 & 227 & 261 & 280 & 311 & 362 & 422 & 519 & 551 \\
Token & \texttt{ (} & \texttt{ to} & \texttt{ important} & \texttt{ is} & \texttt{subset} & \texttt{ is} & \texttt{subset} & \texttt{ The} & \texttt{-} & \texttt{ ring} & \texttt{ Conclusion} & \texttt{\textless{}\textbar{}im\_end\textbar{}\textgreater{}} \\
$A_{i,t}^{\mathrm{OPD}}$ & -7.2268 & -7.1247 & -6.3155 & -9.5342 & -7.9704 & -7.0191 & -16.1068 & -9.0377 & -8.4210 & -6.8585 & -9.7430 & -9.3995 \\
\bottomrule
\end{tabular}}
}
\end{minipage}

\smallskip
\noindent\begin{minipage}[t]{\linewidth}
\centering
\textbf{First 15 response tokens (with context):}

\vspace{2pt}
{\scriptsize
\setlength{\tabcolsep}{2pt}
\resizebox{\linewidth}{!}{%
\begin{tabular}{l*{15}{c}}
\toprule
Position & 0 & 1 & 2 & 3 & 4 & 5 & 6 & 7 & 8 & 9 & 10 & 11 & 12 & 13 & 14 \\
Token & \texttt{We} & \texttt{ are} & \texttt{ given} & \texttt{ a} & \texttt{ **} & \texttt{local} & \texttt{ (} & \texttt{not} & \texttt{ No} & \texttt{ether} & \texttt{ian} & \texttt{)} & \texttt{ domain} & \texttt{**} & \texttt{ \$} \\
$A_{i,t}^{\mathrm{OPD}}$ & +0.0001 & +0.0000 & +0.0256 & +0.1121 & +0.3423 & +0.0000 & -7.2268 & +0.0141 & -4.0181 & +0.0000 & +0.0000 & -1.3005 & +0.0001 & -0.0074 & -0.0000 \\
\bottomrule
\end{tabular}}
}
\end{minipage}

\smallskip
\noindent\begin{minipage}[t]{\linewidth}
\centering
\textbf{Summary (552 tokens):}

\vspace{2pt}
{\scriptsize
\setlength{\tabcolsep}{3pt}
\begin{tabular}{lrr}
\toprule
Stat. & $A_{i,t}^{\mathrm{OPD}}$ & $A_i^{\mathrm{GRPO}}$ \\
\midrule
mean & -0.4796 & -0.3536 \\
std  & +1.6198 &  0.0000 \\
min  & -16.1068 & -0.3536 \\
max  & +2.2114 & -0.3536 \\
\bottomrule
\end{tabular}}
\end{minipage}

\bigskip

\textbf{(c) Response 16 ($A_i^{\mathrm{GRPO}} = +2.4749$, correct, 921 tokens)}

\smallskip\noindent
\noindent\begin{minipage}[t]{\linewidth}
\centering
\textbf{Top 12 by $|A_{i,t}^{\mathrm{OPD}}|$ (ordered by position):}

\vspace{2pt}
{\scriptsize
\setlength{\tabcolsep}{2pt}
\resizebox{\linewidth}{!}{%
\begin{tabular}{l*{12}{c}}
\toprule
Position & 118 & 155 & 173 & 259 & 283 & 577 & 594 & 708 & 712 & 765 & 874 & 901 \\
Token & \texttt{ such} & \texttt{ solutions} & \texttt{ pairs} & \texttt{ unity} & \texttt{ only} & \texttt{Now} & \texttt{ the} & \texttt{ the} & \texttt{:\nl\nl} & \texttt{This} & \texttt{So} & \texttt{---\nl\nl} \\
$A_{i,t}^{\mathrm{OPD}}$ & -6.7144 & -6.9551 & -15.1645 & -7.5274 & -6.8762 & -6.7637 & -5.8708 & -7.6130 & -11.5820 & -7.2075 & -8.5703 & -8.0519 \\
\bottomrule
\end{tabular}}
}
\end{minipage}

\smallskip
\noindent\begin{minipage}[t]{\linewidth}
\centering
\textbf{First 15 response tokens (with context):}

\vspace{2pt}
{\scriptsize
\setlength{\tabcolsep}{2pt}
\resizebox{\linewidth}{!}{%
\begin{tabular}{l*{15}{c}}
\toprule
Position & 0 & 1 & 2 & 3 & 4 & 5 & 6 & 7 & 8 & 9 & 10 & 11 & 12 & 13 & 14 \\
Token & \texttt{We} & \texttt{ are} & \texttt{ given} & \texttt{ a} & \texttt{ **} & \texttt{con} & \texttt{gr} & \texttt{u} & \texttt{ence} & \texttt{ equation} & \texttt{**} & \texttt{:\nl\nl} & \texttt{\$\$} & \texttt{\nl} & \texttt{x} \\
$A_{i,t}^{\mathrm{OPD}}$ & +0.0000 & -0.0002 & -3.5894 & +0.2090 & -1.3109 & -0.0987 & +0.0000 & +0.0000 & +0.0000 & -0.0027 & -0.0141 & -0.0475 & -0.0001 & +0.0000 & +0.0000 \\
\bottomrule
\end{tabular}}
}
\end{minipage}

\smallskip
\noindent\begin{minipage}[t]{\linewidth}
\centering
\textbf{Summary (921 tokens):}

\vspace{2pt}
{\scriptsize
\setlength{\tabcolsep}{3pt}
\begin{tabular}{lrr}
\toprule
Stat. & $A_{i,t}^{\mathrm{OPD}}$ & $A_i^{\mathrm{GRPO}}$ \\
\midrule
mean & -0.3583 & +2.4749 \\
std  & +1.3354 &  0.0000 \\
min  & -15.1645 & +2.4749 \\
max  & +3.4178 & +2.4749 \\
\bottomrule
\end{tabular}}
\end{minipage}


\subsection{Training Step~4}

\textbf{(a) Response 34 ($A_i^{\mathrm{GRPO}} = -2.4749$, incorrect, 1397 tokens)}

\smallskip\noindent
\noindent\begin{minipage}[t]{\linewidth}
\centering
\textbf{Top 12 by $|A_{i,t}^{\mathrm{OPD}}|$ (ordered by position):}

\vspace{2pt}
{\scriptsize
\setlength{\tabcolsep}{2pt}
\resizebox{\linewidth}{!}{%
\begin{tabular}{l*{12}{c}}
\toprule
Position & 26 & 112 & 129 & 212 & 217 & 319 & 580 & 1041 & 1283 & 1311 & 1329 & 1396 \\
Token & \texttt{ \$} & \texttt{b} & \texttt{for} & \texttt{:\nl\nl} & \texttt{ +} & \texttt{ expression} & \texttt{ =} & \texttt{\}\nl} & \texttt{ /} & \texttt{/} & \texttt{Thus} & \texttt{\textless{}\textbar{}im\_end\textbar{}\textgreater{}} \\
$A_{i,t}^{\mathrm{OPD}}$ & -9.9835 & -12.5009 & -15.3752 & -12.1700 & -12.9890 & -11.4401 & -11.1801 & -13.6017 & -9.9589 & -12.8283 & -11.2880 & -10.1182 \\
\bottomrule
\end{tabular}}
}
\end{minipage}

\smallskip
\noindent\begin{minipage}[t]{\linewidth}
\centering
\textbf{First 15 response tokens (with context):}

\vspace{2pt}
{\scriptsize
\setlength{\tabcolsep}{2pt}
\resizebox{\linewidth}{!}{%
\begin{tabular}{l*{15}{c}}
\toprule
Position & 0 & 1 & 2 & 3 & 4 & 5 & 6 & 7 & 8 & 9 & 10 & 11 & 12 & 13 & 14 \\
Token & \texttt{We} & \texttt{ are} & \texttt{ given} & \texttt{ a} & \texttt{ sequence} & \texttt{ \$(} & \texttt{a} & \texttt{\_n} & \texttt{)} & \texttt{\_\{} & \texttt{n} & \texttt{ \textbackslash{}} & \texttt{ge} & \texttt{ } & \texttt{1} \\
$A_{i,t}^{\mathrm{OPD}}$ & +0.0000 & +0.0001 & +0.0000 & +0.0983 & +0.0002 & +0.0564 & +0.0000 & +0.0000 & +0.0052 & +0.0000 & +0.0000 & +0.0005 & +0.0000 & -0.1782 & +0.0000 \\
\bottomrule
\end{tabular}}
}
\end{minipage}

\smallskip
\noindent\begin{minipage}[t]{\linewidth}
\centering
\textbf{Summary (1397 tokens):}

\vspace{2pt}
{\scriptsize
\setlength{\tabcolsep}{3pt}
\begin{tabular}{lrr}
\toprule
Stat. & $A_{i,t}^{\mathrm{OPD}}$ & $A_i^{\mathrm{GRPO}}$ \\
\midrule
mean & -0.3822 & -2.4749 \\
std  & +1.5979 &  0.0000 \\
min  & -15.3752 & -2.4749 \\
max  & +4.9387 & -2.4749 \\
\bottomrule
\end{tabular}}
\end{minipage}

\bigskip

\textbf{(b) Response 21 ($A_i^{\mathrm{GRPO}} = +0.3536$, neutral, 1023 tokens)}

\smallskip\noindent
\noindent\begin{minipage}[t]{\linewidth}
\centering
\textbf{Top 12 by $|A_{i,t}^{\mathrm{OPD}}|$ (ordered by position):}

\vspace{2pt}
{\scriptsize
\setlength{\tabcolsep}{2pt}
\resizebox{\linewidth}{!}{%
\begin{tabular}{l*{12}{c}}
\toprule
Position & 232 & 345 & 391 & 401 & 441 & 513 & 547 & 589 & 889 & 917 & 947 & 951 \\
Token & \texttt{ f} & \texttt{ Ideal} & \texttt{ann} & \texttt{dimension} & \texttt{nil} & \texttt{ Coordinate} & \texttt{ f} & \texttt{ elements} & \texttt{So} & \texttt{These} & \texttt{ Step} & \texttt{ Conclusion} \\
$A_{i,t}^{\mathrm{OPD}}$ & -8.2226 & -11.8967 & -9.0663 & -8.3018 & -8.8799 & -8.2837 & -10.0756 & -9.4204 & -9.2330 & -11.6286 & -8.6771 & -12.8161 \\
\bottomrule
\end{tabular}}
}
\end{minipage}

\smallskip
\noindent\begin{minipage}[t]{\linewidth}
\centering
\textbf{First 15 response tokens (with context):}

\vspace{2pt}
{\scriptsize
\setlength{\tabcolsep}{2pt}
\resizebox{\linewidth}{!}{%
\begin{tabular}{l*{15}{c}}
\toprule
Position & 0 & 1 & 2 & 3 & 4 & 5 & 6 & 7 & 8 & 9 & 10 & 11 & 12 & 13 & 14 \\
Token & \texttt{We} & \texttt{ are} & \texttt{ asked} & \texttt{ to} & \texttt{ determine} & \texttt{ the} & \texttt{ **} & \texttt{dimension} & \texttt{**} & \texttt{ of} & \texttt{ the} & \texttt{ quotient} & \texttt{ ring} & \texttt{  \nl} & \texttt{\$\$} \\
$A_{i,t}^{\mathrm{OPD}}$ & +0.0002 & +0.0001 & +0.6647 & +0.0000 & +0.0742 & +0.0000 & +0.1900 & +0.0000 & +0.3095 & +0.0122 & +0.0000 & +0.0892 & +0.0000 & +0.5187 & -0.0000 \\
\bottomrule
\end{tabular}}
}
\end{minipage}

\smallskip
\noindent\begin{minipage}[t]{\linewidth}
\centering
\textbf{Summary (1023 tokens):}

\vspace{2pt}
{\scriptsize
\setlength{\tabcolsep}{3pt}
\begin{tabular}{lrr}
\toprule
Stat. & $A_{i,t}^{\mathrm{OPD}}$ & $A_i^{\mathrm{GRPO}}$ \\
\midrule
mean & -0.5035 & +0.3536 \\
std  & +1.5621 &  0.0000 \\
min  & -12.8161 & +0.3536 \\
max  & +2.1031 & +0.3536 \\
\bottomrule
\end{tabular}}
\end{minipage}

\bigskip

\textbf{(c) Response 52 ($A_i^{\mathrm{GRPO}} = +2.4749$, correct, 2310 tokens)}

\smallskip\noindent
\noindent\begin{minipage}[t]{\linewidth}
\centering
\textbf{Top 12 by $|A_{i,t}^{\mathrm{OPD}}|$ (ordered by position):}

\vspace{2pt}
{\scriptsize
\setlength{\tabcolsep}{2pt}
\resizebox{\linewidth}{!}{%
\begin{tabular}{l*{12}{c}}
\toprule
Position & 64 & 365 & 449 & 859 & 863 & 935 & 1095 & 1466 & 1707 & 1771 & 1819 & 2214 \\
Token & \texttt{.\nl} & \texttt{T} & \texttt{T} & \texttt{ \textbackslash{}} & \texttt{-} & \texttt{tr} & \texttt{ first} & \texttt{\^{}n} & \texttt{maximum} & \texttt{---\nl\nl} & \texttt{ triangle} & \texttt{ Final} \\
$A_{i,t}^{\mathrm{OPD}}$ & -14.9193 & -8.8550 & -10.1490 & -14.4367 & -10.2472 & -10.0332 & -8.7231 & -10.6075 & -9.5449 & -8.4645 & -10.9428 & -10.9494 \\
\bottomrule
\end{tabular}}
}
\end{minipage}

\smallskip
\noindent\begin{minipage}[t]{\linewidth}
\centering
\textbf{First 15 response tokens (with context):}

\vspace{2pt}
{\scriptsize
\setlength{\tabcolsep}{2pt}
\resizebox{\linewidth}{!}{%
\begin{tabular}{l*{15}{c}}
\toprule
Position & 0 & 1 & 2 & 3 & 4 & 5 & 6 & 7 & 8 & 9 & 10 & 11 & 12 & 13 & 14 \\
Token & \texttt{We} & \texttt{ are} & \texttt{ given} & \texttt{ a} & \texttt{ linear} & \texttt{ operator} & \texttt{ \$} & \texttt{ T} & \texttt{:} & \texttt{ \textbackslash{}} & \texttt{ell} & \texttt{\^{}} & \texttt{2} & \texttt{ \textbackslash{}} & \texttt{rightarrow} \\
$A_{i,t}^{\mathrm{OPD}}$ & +0.0002 & +0.0000 & +0.0229 & +0.0025 & +0.0516 & +0.0000 & +0.0006 & +0.0000 & +0.1267 & +0.0000 & +0.0000 & +0.0000 & +0.0000 & +0.0007 & -3.0091 \\
\bottomrule
\end{tabular}}
}
\end{minipage}

\smallskip
\noindent\begin{minipage}[t]{\linewidth}
\centering
\textbf{Summary (2310 tokens):}

\vspace{2pt}
{\scriptsize
\setlength{\tabcolsep}{3pt}
\begin{tabular}{lrr}
\toprule
Stat. & $A_{i,t}^{\mathrm{OPD}}$ & $A_i^{\mathrm{GRPO}}$ \\
\midrule
mean & -0.3849 & +2.4749 \\
std  & +1.3496 &  0.0000 \\
min  & -14.9193 & +2.4749 \\
max  & +4.1275 & +2.4749 \\
\bottomrule
\end{tabular}}
\end{minipage}


\subsection{Training Step~5}

\textbf{(a) Response 93 ($A_i^{\mathrm{GRPO}} = -2.4749$, incorrect, 1616 tokens)}

\smallskip\noindent
\noindent\begin{minipage}[t]{\linewidth}
\centering
\textbf{Top 12 by $|A_{i,t}^{\mathrm{OPD}}|$ (ordered by position):}

\vspace{2pt}
{\scriptsize
\setlength{\tabcolsep}{2pt}
\resizebox{\linewidth}{!}{%
\begin{tabular}{l*{12}{c}}
\toprule
Position & 111 & 118 & 284 & 287 & 293 & 319 & 461 & 522 & 578 & 587 & 1342 & 1448 \\
Token & \texttt{ (} & \texttt{ a} & \texttt{stretch} & \texttt{ plane} & \texttt{ each} & \texttt{.\nl\nl} & \texttt{no} & \texttt{ f} & \texttt{This} & \texttt{ function} & \texttt{If} & \texttt{ conclusion} \\
$A_{i,t}^{\mathrm{OPD}}$ & -10.7766 & -10.9075 & -10.9272 & -8.3615 & -9.5392 & -8.6375 & -8.6105 & -8.3328 & -7.7441 & -8.6470 & -10.8534 & -8.9139 \\
\bottomrule
\end{tabular}}
}
\end{minipage}

\smallskip
\noindent\begin{minipage}[t]{\linewidth}
\centering
\textbf{First 15 response tokens (with context):}

\vspace{2pt}
{\scriptsize
\setlength{\tabcolsep}{2pt}
\resizebox{\linewidth}{!}{%
\begin{tabular}{l*{15}{c}}
\toprule
Position & 0 & 1 & 2 & 3 & 4 & 5 & 6 & 7 & 8 & 9 & 10 & 11 & 12 & 13 & 14 \\
Token & \texttt{We} & \texttt{ are} & \texttt{ given} & \texttt{:\nl\nl} & \texttt{\textgreater{}} & \texttt{ If} & \texttt{ \$} & \texttt{ f} & \texttt{(x} & \texttt{,} & \texttt{ y} & \texttt{)} & \texttt{ =} & \texttt{ } & \texttt{0} \\
$A_{i,t}^{\mathrm{OPD}}$ & +0.0000 & +0.0000 & +0.0658 & -4.9537 & +0.0067 & +0.1240 & -0.0001 & +0.0000 & +0.0000 & +0.0000 & +0.0000 & +0.0000 & +0.0000 & +0.0000 & +0.0000 \\
\bottomrule
\end{tabular}}
}
\end{minipage}

\smallskip
\noindent\begin{minipage}[t]{\linewidth}
\centering
\textbf{Summary (1616 tokens):}

\vspace{2pt}
{\scriptsize
\setlength{\tabcolsep}{3pt}
\begin{tabular}{lrr}
\toprule
Stat. & $A_{i,t}^{\mathrm{OPD}}$ & $A_i^{\mathrm{GRPO}}$ \\
\midrule
mean & -0.4781 & -2.4749 \\
std  & +1.4702 &  0.0000 \\
min  & -10.9272 & -2.4749 \\
max  & +5.1961 & -2.4749 \\
\bottomrule
\end{tabular}}
\end{minipage}

\bigskip

\textbf{(b) Response 1 ($A_i^{\mathrm{GRPO}} = +0.3536$, neutral, 546 tokens)}

\smallskip\noindent
\noindent\begin{minipage}[t]{\linewidth}
\centering
\textbf{Top 12 by $|A_{i,t}^{\mathrm{OPD}}|$ (ordered by position):}

\vspace{2pt}
{\scriptsize
\setlength{\tabcolsep}{2pt}
\resizebox{\linewidth}{!}{%
\begin{tabular}{l*{12}{c}}
\toprule
Position & 21 & 115 & 117 & 153 & 177 & 220 & 244 & 279 & 395 & 444 & 458 & 489 \\
Token & \texttt{ asked} & \texttt{ Consider} & \texttt{ Equation} & \texttt{ A} & \texttt{So} & \texttt{ \textbackslash{}} & \texttt{ that} & \texttt{ that} & \texttt{ \$} & \texttt{ \$} & \texttt{ Conclusion} & \texttt{:\nl\nl} \\
$A_{i,t}^{\mathrm{OPD}}$ & -6.6305 & -8.5071 & -5.7975 & -10.1694 & -5.5674 & -4.6801 & -6.9888 & -5.6696 & -4.6985 & -7.5715 & -6.0161 & -7.9329 \\
\bottomrule
\end{tabular}}
}
\end{minipage}

\smallskip
\noindent\begin{minipage}[t]{\linewidth}
\centering
\textbf{First 15 response tokens (with context):}

\vspace{2pt}
{\scriptsize
\setlength{\tabcolsep}{2pt}
\resizebox{\linewidth}{!}{%
\begin{tabular}{l*{15}{c}}
\toprule
Position & 0 & 1 & 2 & 3 & 4 & 5 & 6 & 7 & 8 & 9 & 10 & 11 & 12 & 13 & 14 \\
Token & \texttt{We} & \texttt{ are} & \texttt{ given} & \texttt{ the} & \texttt{ matrix} & \texttt{ equation} & \texttt{:\nl\nl} & \texttt{\$\$} & \texttt{\nl} & \texttt{A} & \texttt{\^{}} & \texttt{2} & \texttt{ +} & \texttt{ A} & \texttt{ =} \\
$A_{i,t}^{\mathrm{OPD}}$ & +0.0000 & +0.0001 & -0.2871 & -0.1381 & +0.2385 & +0.0001 & -0.0001 & +0.0000 & +0.0000 & +0.0000 & +0.0000 & +0.0000 & +0.0000 & +0.0000 & +0.0000 \\
\bottomrule
\end{tabular}}
}
\end{minipage}

\smallskip
\noindent\begin{minipage}[t]{\linewidth}
\centering
\textbf{Summary (546 tokens):}

\vspace{2pt}
{\scriptsize
\setlength{\tabcolsep}{3pt}
\begin{tabular}{lrr}
\toprule
Stat. & $A_{i,t}^{\mathrm{OPD}}$ & $A_i^{\mathrm{GRPO}}$ \\
\midrule
mean & -0.2358 & +0.3536 \\
std  & +1.1496 &  0.0000 \\
min  & -10.1694 & +0.3536 \\
max  & +2.7361 & +0.3536 \\
\bottomrule
\end{tabular}}
\end{minipage}

\bigskip

\textbf{(c) Response 55 ($A_i^{\mathrm{GRPO}} = +2.4749$, correct, 2845 tokens)}

\smallskip\noindent
\noindent\begin{minipage}[t]{\linewidth}
\centering
\textbf{Top 12 by $|A_{i,t}^{\mathrm{OPD}}|$ (ordered by position):}

\vspace{2pt}
{\scriptsize
\setlength{\tabcolsep}{2pt}
\resizebox{\linewidth}{!}{%
\begin{tabular}{l*{12}{c}}
\toprule
Position & 590 & 826 & 969 & 1012 & 1452 & 1625 & 1709 & 2074 & 2440 & 2738 & 2837 & 2844 \\
Token & \texttt{ y} & \texttt{0} & \texttt{ =} & \texttt{Int} & \texttt{ which} & \texttt{]} & \texttt{ =} & \texttt{ \$} & \texttt{ -} & \texttt{Each} & \texttt{The} & \texttt{\textless{}\textbar{}im\_end\textbar{}\textgreater{}} \\
$A_{i,t}^{\mathrm{OPD}}$ & -7.7754 & -8.1879 & -8.0133 & -14.2993 & -7.8828 & -7.1504 & -7.6268 & -9.1528 & -6.9237 & -9.4345 & -10.6689 & -8.4856 \\
\bottomrule
\end{tabular}}
}
\end{minipage}

\smallskip
\noindent\begin{minipage}[t]{\linewidth}
\centering
\textbf{First 15 response tokens (with context):}

\vspace{2pt}
{\scriptsize
\setlength{\tabcolsep}{2pt}
\resizebox{\linewidth}{!}{%
\begin{tabular}{l*{15}{c}}
\toprule
Position & 0 & 1 & 2 & 3 & 4 & 5 & 6 & 7 & 8 & 9 & 10 & 11 & 12 & 13 & 14 \\
Token & \texttt{We} & \texttt{ are} & \texttt{ asked} & \texttt{ to} & \texttt{ evaluate} & \texttt{ the} & \texttt{ double} & \texttt{ integral} & \texttt{:\nl\nl} & \texttt{\$\$} & \texttt{\nl} & \texttt{\textbackslash{}} & \texttt{i} & \texttt{int} & \texttt{\_E} \\
$A_{i,t}^{\mathrm{OPD}}$ & +0.0000 & +0.0001 & +0.3269 & +0.0000 & +0.0115 & +0.0001 & +0.1603 & +0.0000 & -0.7045 & -0.0000 & +0.0000 & +0.0004 & +0.0000 & +0.0000 & +0.0022 \\
\bottomrule
\end{tabular}}
}
\end{minipage}

\smallskip
\noindent\begin{minipage}[t]{\linewidth}
\centering
\textbf{Summary (2845 tokens):}

\vspace{2pt}
{\scriptsize
\setlength{\tabcolsep}{3pt}
\begin{tabular}{lrr}
\toprule
Stat. & $A_{i,t}^{\mathrm{OPD}}$ & $A_i^{\mathrm{GRPO}}$ \\
\midrule
mean & -0.2342 & +2.4749 \\
std  & +1.0454 &  0.0000 \\
min  & -14.2993 & +2.4749 \\
max  & +3.8790 & +2.4749 \\
\bottomrule
\end{tabular}}
\end{minipage}

\paragraph{Cross-response observation.}
Across the 15 representative responses above (three responses from each of five consecutive steps), every one of the 180 reported top-magnitude token positions satisfies $|A_{i,t}^{\mathrm{OPD}}|>|A_i^{\mathrm{GRPO}}|$. For example, the reported OPD extremes reach $20.3585$, whereas the largest absolute GRPO advantage among these responses is $2.4749$. Thus, whenever a salient OPD token is active in the fixed 1:1 fusion, its contribution can exceed the verifier-derived response-level contribution by a wide margin and locally dominate the fused update. This selected-token analysis establishes the existence and persistence of the \textbf{magnitude mismatch} across response types and consecutive steps; it does not claim that most tokens in the full batch are large, since Figure~\ref{fig:opd_bin_dist} conversely shows that most raw OPD values are concentrated near zero. Together, the near-zero mass and the few dominant active tokens motivate sparsifying first and then bounding the retained values.

\end{document}